\documentclass[twoside,11pt]{article}

\usepackage{jmlr2e}
\usepackage{color}
\usepackage{graphics}

\usepackage{amsmath,amssymb,amsfonts}

\usepackage[noend]{algpseudocode}
 \usepackage[linesnumbered, ruled, vlined]{algorithm2e}
\usepackage{multirow}
\usepackage{xcolor}

\begin{document}

\title{The Tensor Brain: A Unified Theory of Perception, Memory and Semantic Decoding}

\author{\name Volker Tresp \email volker.tresp@lmu.de \\
 \addr LMU Munich and Siemens Munich, Germany
\AND
 \name Sahand Sharifzadeh$^+$\email$^*$ sahand.sharifzadeh@gmail.com \\
 \addr LMU Munich, Germany
\AND
 \name Hang Li$^+$ \email hang.li@campus.lmu.de \\
 \addr LMU Munich and Siemens Munich, Germany
\AND
 \name Dario Konopatzki \email dk@dkonopatzki.de \\
 \addr LMU Munich, Germany 
\AND
 \name Yunpu Ma \email cognitive.yunpu@gmail.com \\
 \addr LMU Munich and Siemens Munich, Germany 
}
 
\editor{$^+$equal contributions.}

 \maketitle
\begin{center}
{\center Extended Version}\footnote{Sahand Sharifzadeh is now with DeepMind London. Dario Konopatzki is now with ETH Zurich. 
Extended version of the paper that appeared in \textit{Neural Computation}, 2023. The naming of the models has been changed. 
 The discussion on the \textit{post-observation model} and Dirichlet fusion in Section~\ref{sec:dirichlet} has been added. 
Section~\ref{sec:autoen} describes the BTN as an autoencoder. 
Section~\ref{sec:forgetting}, which discusses forgetting and modularity, has been added.}
\end{center}

\begin{abstract}

We present a unified computational theory of an agent's perception and memory. In our model, both perception and memory are realized by different operational modes of the oscillating interactions between a symbolic index layer and a subsymbolic representation layer.
The two layers form a bilayer tensor network (BTN). 
The index layer encodes indices for concepts, predicates, and episodic instances. The representation layer broadcasts information and reflects the cognitive brain state; it is our model of what authors have called the ``mental canvas'' or the ``global workspace.''
As a bridge between perceptual input and the index layer, the representation layer enables the grounding of indices by their subsymbolic embeddings, which are implemented as connection weights linking both layers. The brain is a sampling engine: Only activated indices are communicated to the remaining parts of the brain. Although memory appears to be about the past, its main purpose is to support the agent in the present and the future. Recent episodic memory provides the agent with a sense of the here and now. Remote episodic memory retrieves relevant past experiences to provide information about possible future scenarios. This aids the agent in decision-making. ``Future'' episodic memory, based on expected future events, guides planning and action. Semantic memory retrieves specific information, which is not delivered by current perception, and defines priors for future observations. Our approach explains the great similarity between episodic and semantic memory: Semantic memory models the episodic memory of a future instance. 
We analyse episodic memory and semantic memory in the context of meta-learning. We argue that it is important for the agent to encode individual entities, not just classes and attributes.  Perception is learning: Episodic memories are constantly being formed, and we demonstrate that a form of self-supervised learning can acquire new concepts and refine existing ones. We test our model on a standard benchmark data set, which we expanded to contain richer representations for attributes, classes, and individuals. Our key hypothesis is that obtaining a better understanding of perception and memory is a crucial prerequisite to comprehending human-level intelligence. 

\end{abstract}

\section{Introduction}

With an increase in higher animals' abilities to move and act came a growing demand for high-performing perceptual systems, beyond simple labeling of entities with attributes and classes \citep{hommel2001theory}. This might have been a driving force to develop episodic and semantic memory. Episodic memory engrams permit the recall of recent and remote memories and provide guidance for acting right. Semantic memory engrams provide concept grounding and complement perceived information with background knowledge about concepts.  The agent does not just acquire new skills but gains and memorizes knowledge: It learns about and remembers things in the world.  Memories support the agent in the present and the future: Without memory systems, the brain is literally memoryless.

Episodic memories retrieve previous experiences in an agent's life. 
\textit{Recent episodic memory} permits the agent to remember the immediate past since some state information cannot be directly derived from perceptual input.  A recall is triggered by nearness in time and relevance.  It has evolved such that the agent 
can remember where it has been before, why it is where it is, and what the general context is. Recent episodic memory is an episodic memory that is almost treated as a current observation.
For instance, the agent needs to remember that, even though perception does not give a clue, it is still in the hideout because the bear had been chasing it and might still be lurking outside.
 \textit{Remote episodic memory} can remind an agent about past situations ---similar to the current--- and imminent danger and favorable actions associated with that situation.
 It provides estimates about possible future scenarios and aids the agent in decision-making. Continuing the previous example, the agent might remember previous personal bear encounters and subsequent dangerous situations.
Recall in remote episodic memory is triggered by closeness between episodic representation and scene representation.
Finally, we define \textit{future episodic memories} as events that are expected to be a memory in the future. 
Information on future events influences planning and action.
 
Semantic memory models the time-invariant statistics of statement probabilities and acts as their prior.
It is a dictionary view of the agent's life.\footnote{Some authors distinguish between a formal dictionary and a grounded cognitive encyclopedia \citep{evans2012cognitive}. We will not make this distinction.} 
It aggregates entity information, which might have been acquired at different episodes, and this information becomes local in the graph formed by semantic memory.
Semantic memory enables multimodal integration. 
For example, if Sparky is discovered in a scene, semantic memory provides background information, for example, that Sparky is a young dog and is owned by Jack; and although dogs, in general, might be aggressive, Sparky is a friendly dog.
Semantic memory support can be essential for survival. An agent simply knows that bears are dangerous, even when a bear looks cozy and sleepy and even if the agent did not yet have an unpleasant encounter with a bear.
Semantic memory is the default semantic state estimator and is complemented by perception and recent episodic memories, when and where available. 
Our approach explains the great similarity between episodic and semantic memory: 
 Semantic memory is the expected episodic memory of a future instance. 
An elementary symbolic statement says that an entity has a particular attribute, belongs to a specific class, or that an entity has a particular relationship with another entity. 
This motivates our assumption that basic facts are expressed as triple statements of the form \textit{(subject, predicate, object)}.\footnote{
In a relational Bayesian network or a Markov logic network, a triple statement is represented as a node; in a graph neural network, it would be a typical output node.}
In most languages, basic facts are expressed in this or a similar format. Thus, triple sentences are arguably of fundamental relevance for encoding and communicating perception, episodic memory, and semantic memory, all of which humans can easily describe by language. 
Communication is important to make the invisible visible: for example, an agent could inform peers that a bear is lurking outside the hideout, even if it cannot be seen. 
Triple statements can represent relationships between entities, which enables, for example, rich scene descriptions and reasoning in social and spatial networks. 
An agent can understand not only that a bear and a deer are in a scene but that, luckily, the bear is not chasing the agent itself but the deer. 
Symbolic representations and reasoning, classical System-2 properties, depend on the representations of entity-to-entity relationships \citep{halford2014categorizing}. 

 The starting point of our work is a mathematical model, that is, the bilayer tensor network (BTN). It implements mathematical models for perception and memory that obey some constraints imposed by biology. We then discuss how particular features of our model relate to current discussions in cognition and neuroscience and make predictions about implications of our model to both.  The BTN implies a very simple architecture that contains two basic layers: the \textit{index layer} and the \textit{representation layer}.
Context is provided by a third layer, the \textit{dynamic context layer}, which interacts with the representation layer, provides context, and stores state information when the brain's attention moves from one entity to another.

The symbolic index layer contains indices for concepts, predicates, and episodic instances known to the agent.
The index layer labels the activation pattern in the representation layer 
and then feeds back the embedding of that label to the representation layer.
The embedding vectors are implemented as connection weights linking both layers.
An index is a focal point of activity and competes with other indices, but, since it constantly interacts with the representation layer, it is never active in isolation. Embeddings have an integrative character: the embedding vector for a concept index integrates all that is known about that concept, and the embedding vector for an episodic index represents the world state at that instance. 
 The subsymbolic representation layer is the main communication platform.
In cognitive neuroscience, it would correspond to, what authors call, the ``mental canvas'' or the ``global workspace'' and reflects the cognitive brain state. 
In bottom-up mode, scene inputs activate the representation layer, which then activates the index layer. In top-down mode, an index activates the representation layer, which might subsequently activate even earlier processing layers. This last process is called the embodiment of a concept. 

Perception and memories first produce subsymbolic representations. They represent the agent's current and past cognitive state and are subsequently decoded semantically to produce 
sequences of activated indices that form symbolic triple statements.
These sequences then give feedback to the representation layer and earlier processing layers and thus inform the brain as a whole about what has been decoded. The brain is a sampling engine: Only activated indices are communicated to the remaining parts of the brain.
 
In our approach, perception and memory produce triple statements by stochastic sampling,
which is a central bottleneck, as also discussed by \cite{dehaene2014consciousness}.
We introduce an attention approximation, which avoids intermediate sampling decisions on visual entities.

One can consider a graph where concepts are represented as nodes, and triples become labeled directed links pointing from subject node to object node. In computer science, this would be called a knowledge graph (KG). Whereas symbolic reasoning would purely act on the graph, embedded reasoning, as realized by the BTN, involves concept embeddings and performs information propagation via the representation layer. 

This article is organized as follows.
In the next section, we cover related work.
Section~\ref{sec:triples-tensors} introduces concepts, triple statements, and probabilistic memory models.
In Section~\ref{sec:ourmodel}, we present our proposed bilayer tensor network (BTN) and demonstrate how it realizes perception and the different memory functions. 
 We relate it to hierarchical Bayesian modelling and meta-learning. We show how a posterior model for an instance can be defined and how this posterior model, but also episodic memory and semantic memory, can be viewed as different variants of Dirichlet updates. 
Section~\ref{sec:algimp} presents and discusses the BTN's algorithmic implementation, sampling, and attention approximations. 
The following sections present experimental results and discuss potential relationships to cognition and neuroscience.
Section~\ref{sec:expdisc} introduces the data set and discusses perception, the representation layer, and the dynamic context layer. 
Section~\ref{sec:conceptorg} discusses semantic memory engrams and their semantic decoding. Using social network data, we demonstrate multimodal integration in semantic memory.
Section~\ref{sec:reas} compares embedded reasoning with symbolic reasoning and proposes that \textit{embedded symbolic reasoning} performs symbolic reasoning using embedding vectors. 
We introduce the one-brain hypothesis, which emphasizes, first, that the brain uses only one representation layer and, second, that perception, episodic memory, semantic memory, and embedded reasoning all rely on the same BTN architecture. We discuss relationships between our approach to cognitive linguistics and consciousness research. We argue that the internal triple-oriented fast speech is transformed into an external, sophisticated slow speech.
Section~\ref{sec:cogmem-em} covers episodic memory and the way the agent estimates current and future world states. 
In Section~\ref{sec:coglearn}, we discuss that perception involves the storing of new episodic memories 
and show how a form of self-supervised learning can learn new concepts and refine existing ones.
We consider that new perceptual experiences might require representations for new entities; that is, we do not assume domain closure. Section~\ref{sec:concl} contains our conclusions.

\section{Related Work}

\subsection{Tensor Networks for Modeling Knowledge Graphs}

The bilayer tensor network (BTN) is an example of a tensor network. 
RESCAL was the first tensor-based embedding model for triple prediction in relational data sets and knowledge graphs \citep{nickel2011three,nickelfactorizing2012}.
Embedding learning for knowledge graphs evolved into a sprawling research area \citep{bordestranslating2013,socherreasoning2013,yang2014embedding,nickel2015holographic,trouillon2016complex,dettmers2018convolutional}. \citep{nickel2015} provides an overview and PyKEEN \citep{ali2021pykeen} a comprehensive software library. In contrast to previous approaches, the BTN can be implemented as an interaction between an index layer and a representation layer, and thus it is more suitable for brainware implementations.

\subsection{Cognitive Tensor Networks and Related Models}

Tensors have been used previously as memory models where the main focus was on simple associations \citep{hintzman1984minerva,kanerva1988sparse,humphreys1989different,osth2015sources} and compositional
structures \citep{smolensky_tensor_1990,pollack1990recursive,platecommon1997,halfordprocessing1998,ma2018holistic}. 
In the tensor product approach \citep{smolensky_tensor_1990}, encoding or binding is realized by a tensor product (generalized outer product) and compositionality by tensor addition. 
In the structured tensor analogical reasoning (STAR) model \citep{halfordprocessing1998}, a working memory is constructed by a superposition of tensor products. 
Either standard basis (one-hot) or random vectors are used as representations for concepts and predicates. 
Some basic symbolic reasoning operations are implemented, such as analogical and transitive inference. 
The main difference to our work is that we approximate the data tensor by a tensor network, that is, the BTN. 
This permits generalization and realizes embedded reasoning, as well as the seamless integration of sensory input. 

The application of embedding-based tensor models to neurocognitive models started with \citep{tresp2015learning}, which introduced tensor networks with index embeddings for perception, as well as semantic and episodic memory. It did not explicitly consider scene bounding boxes and did not contain experimental results. 
In further work, the connection between temporal and semantic tensor networks was analysed \citep{tresp2017tensor, tresp2017ccn,tresp2017embedding, ma2018embedding}.
Semantic memory was derived from episodic memory by an integration step performed in latent space.
In this article, we avoid explicit integration and extend the approach to include perception. 
 
The neurocognitive tensor network theory evolved in the 1980s and followed the idea of geometrization of biology. It is a theory of brain function, particularly that of the cerebellum.
Metric tensors transform sensory space-time coordinates into motor coordinates \citep{pellionisz1980tensorial}. 
There, tensor networks are used very differently from the work presented here.

\subsection{Visual Relationship Detection and Scene Graphs}

In 2016, the Stanford Visual Relationship data set was published, which contained images annotated with triple sentences \citep{lu2016visual} and \citep{krishna2017visual}.
The two articles made their annotated data available, which spawned an explosion of research activity in visual relationship detection (VRD). 
The background information in \citep{lu2016visual} was extracted from a text corpus. 
Recent work in this direction is \citep{luo2019context}. 

VRD models for knowledge graphs were proposed by \citep{baier2017improving,zhang2017visual,baier2018improving}. 
\citep{baier2017improving} showed how a prior distribution derived from triple occurrences could significantly improve on pure vision-based approaches and on approaches that used prior distributions derived from language models.
 \citep{sharifzadeh2019improving} showed further improvements by including 3D image information.
 \citep{trespmodel2019,tresp2020tensor} describe more recent publications in this tradition. 
The presented work introduces more clearly the different operational modes and provides more extensive experimental results. 

Triple statements generated from an image form a scene graph \citep{johnson2015image}. Work on scene graphs attempts to find a unique, globally optimal interpretation of an image. 
State-of-the-art scene graph models are described in \citep{yang2018graph,zellers2018neural,hudson2019learning}. 
\citep{sharifzadeh2020classification} captures the interplay between perception and semantic knowledge by introducing schema representations and implementing the classification as an attention layer between image-based representations and the schema.

\subsection{Related Modern Technical Models for Memory}

\citep{hochreiter1997long} convincingly demonstrated the importance of memory systems in recurrent neural networks. Important later extensions are the neural Turing machine (NTMs) \citep{graves2014neural} and memory networks \citep{weston2014memory,sukhbaatar2015end}.
In those articles, episodic memory acts as an instance buffer. Both use a recurrent neural network in combination with attention mechanisms. Large language models, like OpenAI’s GPT-3, are more recent developments \citep{brown2020language}.

In our BTN, a dynamic context layer is part of a (nonstandard) recurrent neural network. 
The attention mechanisms in our model, episodic attention and semantic attention, are quite different from the attention mechanisms in those articles. 
Also, our goal is to derive triple statements, whereas, in those models, the tasks are query answering and improved token embeddings.

\subsection{Dual Process Theory and Complementary Learning Systems (CLS)}

In psychology, dual process theory concerns the interplay in the mental processing of an \textit{implicit}, automatic, unconscious process (shared with animals) and an \textit{explicit}, controlled, conscious process (uniquely human). 
See \citep{evans2003two} for a review.
In our model, the implicit side would be on the level of embeddings and representations, whereas the explicit side is on the level of the concept indices and the extracted triple sentences. 

One instance of a dual process theory is Kahneman's System-1 / System-2 dichotomy \citep{kahneman2011thinking}. System-1 is fast, instinctive, and emotional and does not require mental effort. System-2 is slower, more deliberate, and more logical and requires mental effort.

CLARION is a dual-process model of implicit and explicit learning \citep{sun1996learning}.
It is based on one-shot explicit rule learning (i.e., explicit learning) and gradual implicit tuning (i.e., implicit learning).

A different but related dichotomy can be found in the complementary learning systems (CLS) theory \citep{mcclelland1995there,kumaran2016learning}, where the formation of time indices and their embeddings would be part of a nonparametric learning system centered on the hippocampus, which allows rapid learning of the specifics of individual items and experiences \citep{kumaran2016learning}.
Slow training would be part of a parametric learning system, which serves as the basis for the gradual acquisition of structured knowledge about the environment to neocortex \citep{kumaran2016learning}.
In this article, we introduce self-supervised learning for rapid learning and discuss a consolidation process of learned knowledge. 
\cite{mcclelland2020placing} analysed the connection between memory, perception, and language, which is also a focus of this article.

\subsection{The Bayesian Brain}

Our approach can be related to the tradition of Bayesian approaches to brain modeling \citep{dayan1995helmholtz,rao1999predictive,knill2004bayesian,kording2004bayesian,tenenbaum2006theory,griffiths2008bayesian,friston2010free}.
In our approach, all past experiences train the perceptual system and also contribute to perception via memory. 
In \citep{baier2017improving} an explicit semantic prior distribution was used, describing \textit{a priori} probabilities for triple sentences. For inference, Bayes' formula is used.
The remarkable improvement in performance after integrating the prior information indicates that triple representations might be a powerful abstraction level for formulating prior knowledge, in general.
\citep{sharifzadeh2020classification} showed that the probabilistic knowledge graph acts as an inductive bias in perception. It discusses the role of a prior as an integrator of multimodal information and its role in filling in nonperceptual background information. 

In the work presented here, we separately model and perform completion on the observed statements, 
statement priors, and statement posteriors. This is in the context of Bayesian probabilistic inference but not in the sense of Bayesian statistics. 

In a Bayesian brain approach, top-down connections are often associated with predictive models; they provide predictions about future state probabilities. Here, we emphasize that, in addition, top-down connections inform the representation layer and earlier processing layers--- what the brain has detected.

\section{Concepts, Triple Statements and Probabilistic Memory Models}
\label{sec:triples-tensors}

\vspace{0.25cm}
{\raggedleft\textit{You only see what you know.} ---Johann Wolfgang von Goethe, an Friedrich von Müller, April 24, 1819}
\vspace{0.25cm}

\subsection{Triple Statements}
\label{sec:ts}

% concepts
To understand what is perceived, an agent needs to have an understanding of the things in the world and their relationships.
We assume that the agent's mind is aware of $N_C$ concepts $\mathcal{C} = \{c_1,\dots, c_{N_C}\}$. 
A concept can, for example, represent an entity $e \in \mathcal{E} \subset \mathcal{C}$, or a class $k \in \mathcal{K} \subset \mathcal{C}$, which stands for a collection of entities, or an attribute $b \in \mathcal{B} \subset \mathcal{C}$.

% predicates
In addition to concepts, we also consider a set of predicates $p \in {\mathcal{P} = \{p_1, \dots, p_{N_P}\}}$, where $N_P$ is the number of predicates the agent is aware of.
A triple statement has the form $(s, p, o)$, where $s \in \mathcal{C}$ assumes the role of the subject, $o \in \mathcal{C}$ assumes the role of the object, and 
predicate $p \in \mathcal{P}$.
Examples of triple statements are: \textit{(Munich, partOf, Bavaria)}, \textit{(Sparky, looksAt, Jack)}, \textit{(AkiraKurosawa, directorOf, SevenSamurai)}, and
 \textit{(Jack, knows, Mary)}.

Following the notion of the semiotic triangle of \citep{ogden1923meaning}, we can look at triples and their semantics from three different perspectives:
\begin{itemize}
\item The agent-independent objective world (world semantics). Some triples have an interpretation in the real world, and they stand for propositions,
like \textit{(Munich, partOf, Bavaria)} and 
\textit{(Sparky, looksAt, Jack)}. This view is mostly taken in a formal analysis of cognition and linguistics \citep{montague1970universal,fodor1975language} and is sometimes referred to as truth-conditional semantics. 
An agent is only aware of some of the concepts in the world, sometimes called the ``projected world'' or the ``construal world.''

\item An agent's mind and brain (an agent's neurocognitive semantics): 
A triple is a statement based on concepts in an agent's mind. 
These might have an immediate meaning in the objective world but could also be untestable statements with unclear world semantics such as \textit{(Love, stongerThan, Hate)}. 
A hypothesis of this article is that triple statements correspond to processes in the brain, that is, a sequential firing of index representations. They form the inner fast speech.

\item {An agent's utterance (an agent's linguistic semantics). A triple is a simple language clause involving symbols. Within the approach in this article, we assume that any triple statement in the mind can also be expressed linguistically. But what is spoken, in general, represents external slow speech and is obviously more complex, nuanced, and sophisticated than the inner fast speech, and is modulated, for example, by intent, social context, and cultural background.}
In general, the relationship between statements in the mind (with concepts) and in language (with symbols) is a matter of an open debate~\citep{evans2012cognitive}.
 
\end{itemize}
We are aware that the relationship of the three perspectives touches on fundamental issues in cognition, linguistics, neuroscience, philosophy, and many other scientific fields and academic disciplines. 
A more detailed discussion would be beyond the scope of this article. This article assumes the position of the agent. We are concerned only about how the agent views the world. We focus on personal Bayesian probabilities, modeling the agent's neurocognitive semantics, where we restrict ourselves to statements that,
 in the mind of the agent, can be true or false, for example, \textit{(Sparky, ownedBy, Jack)}, and facts that can be related to observations, for example, \textit{(Sparky, hasColor, Black)}. 
This is more of a practical than a principled constraint. On some dimensions thought and reality agree simply to act right and to guarantee survival; on others, there is likely no agreement between agents, no clear relationship to an objective reality, and in determining the truth values on statements, personal emotions and judgments might play a significant role.

\subsection{Knowledge Graphs}

In a knowledge graph (KG), a triple is represented as a directed labeled link from concept $s \in \mathcal{C}$ to concept $o \in \mathcal{C}$, where the link is labeled by $p$.
Thus, a labeled link represents a triple statement of the form $(s, p, o)$ where $s$ is called \textit{head} or \textit{source node} and $o$ is called \textit{tail} or \textit{target node}. Knowledge graphs currently have a great impact in applications and, as our approach, are entity oriented.

\subsection{Unary and Binary Statements}
\label{Sec:semstat}

When $s$ and $o$ are entities, $(s, p, o)$ stands for the ground atom $p(s, o)$. 
\textit{(Munich, partOf, Bavaria)} would stand for \textit{partOf(Munich, Bavaria)}. 

We assume a strong default predicate \textit{hasAttribute} (abbreviated as \textit{hA}), which, depending on the subject type and the object type, can stand for certain other predicates from the set $\mathcal{P}^B \subseteq \mathcal P$.
We write \textit{(s, hA, c)}, $c \in \mathcal{C}$.
 \begin{itemize}
 \item If both subject and object are entities, the implicit predicate is \textit{sameAs}; thus, \textit{(Jack, hA, John)} stands for \textit{(Jack, sameAs, John)}. 

 \item If the subject is an entity and the object is an attribute, the implicit predicate is attribute specific but should be obvious; thus \textit{(Jack, hA, Tall)} stands for \textit{(Jack, height, Tall)}. 

 \item If the subject is an entity and the object is a class, the implicit predicate is \textit{type}, as in \textit{(Sparky, type, Dog)}.

 \item If both subject and object are classes, the implicit predicate is \textit{subClass}, as in \textit{(Dog, subClass, Mammal)}; note that the transitive \textit{subClass} predicate permits the modeling of deep ontologies.

 \end{itemize}

Triple sentences involving the \textit{hasAttribute} predicate (and its substitutes) we call \textit{unary statements}.
The remaining $N_B$ predicates from $\mathcal{P}_B \subseteq \mathcal P$ form \textit{binary statements}. 
We will refer to the object in a unary statement also simply as the unary label of the subject, and to the predicate in a binary statement as the binary label of a concept pair.
Binary statements are required when the default unary interpretation is not applicable, as in \textit{(Jane, motherOf, Jack)} (the default would be \textit{(Jane, sameAs, Jack)}).

Higher-order relations can be reduced to a set of binary statements, for example, by using additional concepts \citep{noy2006defining}. 
For example, \textit{match(Player1, Player2, Location)} becomes \textit{(matchID, hasPlayer, Player1)}, \textit{(matchID, hasPlayer, Player2)}, and \textit{(matchID, hasLocation, Location)}, where \textit{(matchID)} is the additional concept.

\subsection{Probabilistic Models with Boolean Variables}

In the assumed lifetime and the mind of an agent, some statements are always true, as \textit{(Munich, partOf, Bavaria)}, but the truth values of other statements can change in time, as \textit{(Munich, weather, Sunny)}.
Thus, with each triple sentence, at each episodic instance $t$, the agent associates a \textit{Boolean semantic state variable} $Y_{s, p, o, t}$.
If the agent is certain that $(s, p, o)$ is true at episodic instance $t$, then $Y_{s, p, o, t} = 1$
and if $(s, p, o)$ is observed to be false at episodic instance $t$, then $Y_{s, p, o, t} = 0$. 
The agent's \textit{semantic world state} at episodic instance $t$ is defined as the states of all semantic state variables.
Here, we assume that the agent is concerned with $N_T$ past episodic instance $\mathcal{T} = \{t_1,\dots, t_{N_T}\}$.

We consider that data, that is, information on the states on triple statements, arrives in chunks, which we call episodes. \cite{mcclelland2020placing} call them situations and in cognition they are called event frames. 
In knowledge graphs, they could form a namespace.
We use the terms events and episodes almost interchangeably, although, in a narrower sense, we reserve the term episode for a sequence of events. (See also the discussion in Section~\ref{sec:cogmem-em}.)
Each episode provides information about the truth values of a subset of all statements.
In this article, we assume that episodic data is either provided by a human annotator (supervised learning) or generated in self-supervised learning (see Section~\ref{sec:coglearn}).
We can form a data tensor of observed triples as a four-mode triple-observation tensor of the form
\begin{equation*} \label{eq:dtensor}
{\underline {\underline T}} = 
\sum_{s, p, o, t} 
y_{s, p, o, t} \; \mathbf{e}^s \otimes \mathbf{e}^p \otimes \mathbf{e}^o \otimes \mathbf{e}^t .
\end{equation*}
Here, $\mathbf{e}^i$ is our notation for a standard basis vector, that is, a one-hot vector, with the one at position $i$. Such data tensors have also been the starting point for the tensor memories of \cite{halfordprocessing1998} and 
\citep{ma2018holistic}. A tensor entry $y_{s, p, o, t} $ is simply equal to one if the statement $(s, p, o)$ was observed by the agent to be true at $t$ and equal to zero if the corresponding statement was observed or concluded by the agent to be false at $t$; otherwise, it is unknown.
The temporal KG (tKG) is the graph representation of ${\underline {\underline T}}$, where the labeled links between concepts are time-dependent .
We assume that data is acquired from different modalities, like vision or language. 
We invoke a local closed-world assumption (LCWA): We assume that 
for entities for which data is acquired in a modality, the truth values of all modality-specific unary statements are available. 
For entity pairs in a modality, we assume that the truth values of all modality-specific binary statements are available.
The issue of which entries in ${\underline {\underline T}}$
are observed might involve some selection or attention process. For example, state changes or singular events might be selected with some priority. We discuss this issue in Section~\ref{sec:stem}.

In the next section, we introduce the bilayer tensor network (BTN), which performs data completion.   For all
$y_{s, p, o, t}$ that are either observed to be true or concluded by the LCWA to be false, the BTN \underline{Boolean observation model} is
\begin{equation} \label{eq:Eobservation-model}
 \mathbb{E}(Y_{s, p, o, t} | \; {\underline {\underline T}})
 \equiv 
 \mathbb{E}(Y_{s, p, o, t})
 \approx 
 y_{s, p, o, t} .
\end{equation} 
The observation model has the characteristics of an autoencoder, where ${\underline {\underline T}}$ represents noisy observations. This relationship is indicated by the approximation symbol ($\approx$).
As indicated, we simplify notations here and in the following by not explicitly conditioning on ${\underline {\underline { T}}}$.

We also define the quantity $i_{s, p, o, t}$, where 
$i_{s, p, o, t} = 1$, if $y_{s, p, o, t}=1$, and $i_{s, p, o, t}=0$, otherwise. Thus, unknowns in the triple-observation tensor are assigned the value $0$.
Consider a future instance $t'$ for which no data is yet available. We define the BTN's 
 \underline{expected-state model}
as 
\begin{equation} \label{eq:priorm}
\mathbb{E}(Y_{s, p, o, \bar t} | \; {\underline {\underline { T}}})
\equiv \mathbb{E}(Y_{s, p, o, \bar t})
\approx 
\frac{1}{{N_{s, p, o, \mathrm{known}}}} \sum_{t} i_{s, p, o, t}
.
\end{equation}
Here, $\bar t$ is a new instance not represented in ${\underline {\underline { T}}}$.
$N_{s, p, o, \mathrm{known}}>0$ counts the times that the truth values of $(s, p, o)$ are known, either because they were observed to be true or as concluded by the LCWA to be false. The expected-state model simply predicts the mean truth value of a future triple statement.
The average in the last equation might only consider a limited time horizon to be able to account for state changes (see also Section~\ref{sec:coglearn}).

If we consider that the observations at an instance
 describe a knowledge graph, then the expected-state model performs integration by modelling the average truth values of known triples' truth values.

\subsection{Observation Model (Episodic Memory)}
\label{sec:retrm}

In this context $s, p, o, t$ are the states of categorical variables.
For subject and object, they have $N_C$ states, for the predicate $N_P$ states, and for instances $N_T$ states.
The BTN's \underline{observation model} is
\begin{equation} \label{eq:OM*}
 \mathbb{P}(s, p, o| t, {\underline {\underline { T}}})
 \equiv
 \mathbb{P}(s, p, o| t)
 \approx 
 \frac{1}{N_{t}}
 i_{s, p, o, t} .
\end{equation}
It models the distribution of the indices. 
$N_{t}=\sum_{s, p, o} i_{s, p, o, t}$ is the total number of triples observed to be true at $t$. 

As we will discuss further down, 
the observation model mathematically describes the symbolic aspect of episodic memory. 
Episodic memory is about what was actually observed. It implicitly assumes a closed-world assumption (CWA), in the sense that what was not observed should not be sampled.

From the definitions, it follows that for observed triples and where the LCWA applies 
\[
\mathbb{P}(s, p, o |t) \approx \mathbb{E}(Y_{s, p, o, t}) / {N_{\textit{t}}} .
\]
 The agent can thus assume that for a given $t$, a sample generated from the observation model is true. 
 If ${\underline {\underline T}}$ is a database, $\mathbb{E}(Y_{s, p, o, t})$ describes a probabilistic model of that database, and $\mathbb{P}(s, p, o |t)$ is a database retrieval model.

\subsection{Pre-observation model (Semantic Memory)}
\label{sec:semmem}

The BTN's \underline{pre-observation model} is
\begin{equation} \label{eq:multpr}
\mathbb{P}(s, p, o| \bar t, {\underline {\underline T}}) 
 \equiv
 \mathbb{P}(s, p, o| \bar t)
\approx 
 \frac{1 }{N_{\mathrm{total}}} \sum_t i_{s, p, o, t} .
\end{equation} 
$N_{\mathrm{total}} = \sum_t N_{t}$ is the total number of observed true triples. The pre-observation model
provides background information on the subject.  It represents the predictions of a model with an informative learned prior for some future instance $\bar t$.
As we will discuss in the course of the paper, 
the pre-observation model mathematically describes the symbolic aspect of semantic memory. 
Samples generated by the pre-observation model correspond to triples often observed to be true. See a more detailed discussion in Appendix~\ref{sec:bm}.

If we consider again that the observations at an instance
 describe a knowledge graph, then the pre-observation model performs integration by summing the truth values of a triple statement at different instances, followed by normalization.
Whereas the expected-state model predicts what is true or false, the 
pre-observation model predicts what should be observed next at a new instance.

\subsection{Post-observation Model and Meta-learning}
\label{sec:dirichlet}

The pre-observation model estimates the semantic world state prior to observations for a new instance $t'$. 
It predicts what should be observed next.
The post-observation model combines episodic and semantic memory to estimate the state for $t'$ after taking into account observations at $t'$. 
Motivated by a hierarchical Bayesian analysis, we consider the pre-observation model 
to be the base distribution of a Dirichlet distribution with concentration parameter $\gamma > 0$.
We obtain the \underline{post-observation model} by a Dirichlet update of the form
\begin{equation} \label{eq:dirfusdef}
\frac{1}{\gamma + \tilde N_{t'}} \left(\gamma \mathbb{P}(s, p, o| \bar t)
+ \tilde N_{t'} \mathbb{P}(s, p, o| t') 
\right) .
\end{equation}
The value of $\gamma$ indicates the trust in the pre-observation model and $\tilde N_t = N_t$. 
We can sample from this distribution by sampling from the pre-observation model (semantic memory)
with a frequency proportional to $\gamma$ and from
the observation model (episodic memory) 
with a frequency proportional to $N_t$. 
The latter recalls actual observations. The former  adds triples describing the  state, prior to the observation. As discussed in Section~\ref{sec:stem}, the observations might include samples from recent episodic memories, which are treated as current observations. 
The post-observation model should not be confused with the updates of the pre-observation model. For the update of the pre-observation model, we would have $\gamma = N_{\textit{total}}$. 
In essence, in the post-observation model, the current observations obtain a larger weight. In terms of meta-learning, an instance is a domain, and an instance-specific episodic memory a domain-specific model, but with shared embedding vectors. Semantic memory would be an instance-independent model trained by pooling all data. The post-observation model is the target-domain prediction. 
Table~\ref{tab:dirs} provides a summary of the different models. 
Appendix~\ref{sec:visual} provides a visualization.

\begin{table}[htp]
 \centering
 \begin{tabular}{|l|c|l|c|c|}
 \hline 
 Model & Function & $\gamma$ & $\tilde N_t$ & Embedding \\
 \hline 
 \hline 
 
Pre-observation model ($<t'$) & Semantic memory & $N_{\textit{total}}$ & $0$ & $\mathbf{\bar a}$
 \\
 Observation model ($t'$) & Episodic memory & 0 & $N_{t'}$ & ${\mathbf{a}_t}$
 \\
 Post-observation model ($t'$) & Posterior state at $t'$ & hyperp. & $N_{t'}$ & ${\approx \mathbf{a}_t} + (\gamma/N_{t'}) \mathbf{\bar a}$
 \\
 \hline 
Pre-observation model ($>t'$) & Semantic memory & $N_{\textit{total}}$ & $N_{t'}$
& $\mathbf{\bar a}$ (updated)
\\

 \hline 
 \hline 

 \end{tabular}
 \caption{
Special cases of the Dirichlet update in Equation~\ref{eq:dirfusdef}.
The pre-observation model
can generate samples from the state prior to observations at $t'$ ($<t'$). 
The observation model
can generate samples from the observations at $t'$. 
The post-observation
can generate samples from the state posterior to observations at $t'$, for the state at $t'$. 
The pre-observation model ($>t'$)
integrates observations at $t'$ to predict future states. Embedding vectors will be introduced in Section~\ref{sec:ourmodel}.
}
 \label{tab:dirs}
\end{table}

\subsection{Modelling Dependencies: Generalized Statements}
\label{sec:gensta}

Can classes or attributes be the subject in a triple statement? Whereas the semantics of \textit{(Sparky, hasColor, Black)} is clear, the semantics of \textit{(Dog, hasColor, Black)} is less well defined: For example, it could mean that there is a least one dog that is black, or that some or most dogs are black, or that all dogs are black. In our work, we consider frequencies. 

We define a triple $(c_1, \textit{hA}, c_2)$ with $c_1, c_2 \in \mathcal{C}$ where 
\begin{equation} \label{eq:genm}
\mathbb{E}(Y_{c_1, \textit{hA}, c_2, \bar t}) \equiv 
\mathbb{E}(Y_{\bar s, \textit{hA}, c_2, \bar t} | Y_{\bar s, \textit{hA}, c_1, \bar t} ) 
\end{equation} 
\[
\approx 
\frac{1}{{\sum_s N_{s, \textit{hA}, c_1, \mathrm{known}}}} 
{\sum_{t, s} i_{s, \textit{hA}, c_1, t}} \; i_{s, \textit{hA}, c_2, t}
.
\]
This is the probability that a new entity $\bar s$ at a new time instance $\bar t$ that has unary label $c_1$ will also have unary label $c_2$. Generalized statements are useful for predictions at a new instance $\bar t$ where only partial measurements are available. With a predicted unary label \textit{Dog} and using the generalized statement \textit{(Dog, hA, Mammal)}, the unary label \textit{Mammal} can be predicted.

A generalized statement corresponds to a probabilistic version of the rule $\forall s: \textit{(s, hA, Mammal)} \leftarrow \textit{(s, hA, Dog)}$ where $s$ is a variable that stands for an entity. 
 Generalized statements  are  the basis for embedded symbolic reasoning described in  Section~\ref{sec:embsymreas}.

Distinguishing between entities, classes, and attributes is important for some of the discussions. But generalized statements permit us to treat all concepts almost identically in the algorithmic implementation in Section~\ref{sec:algimp}.

\section{A Bilayer Tensor Network (BTN)}
\label{sec:ourmodel}

\subsection{The BTN Observation Model}
\label{sec:autoen}

In this section, we introduce the embedding-based bilayer tensor network (BTN).
It assigns an $r$-dimensional latent embedding vector to each concept $c$, predicate $p$, and instance $t$.
The embedding vectors
$\{\mathbf{a}_c\}_{c \in \mathcal{C}}$, 
$\{\mathbf{a}_p\}_{p \in \mathcal{P}_B}$, 
and
$\{\mathbf{a}_t\}_{t \in \mathcal{T}}$
 form the columns of the embedding matrix $A$.

In the BTN, all observed triples are independent, given the embedding matrix $A$. With some of the children observed, represented as the triple-observation
tensor ${\underline {\underline T}}$, we get for a triple $(s, p, o)$ at $t$,
\begin{equation} \label{eq:inflatent}
\mathbb{P}(s, p, o |t, {\underline {\underline T}} ) = \int \mathbb{P}(s, p, o |t, \mathbf{a}_s, \mathbf{a}_p, \mathbf{a}_o, \mathbf{a}_{t}) \; \mathbb{P} (A | {\underline {\underline T}}) \; d A .
\end{equation}
 The last equation has the characteristics of an autoencoder, where ${\underline {\underline T}}$ represents noisy observations.
We use a point estimate approximation as 
\begin{equation} \label{Eq:delta}
 \mathbb{P} (A | {\underline {\underline T}}) \approx \delta(A - \hat A)
\end{equation}
where $\delta(\cdot)$ is the Dirac delta distribution.
The observation model then becomes
\begin{equation} \label{eq:inflatentpoint}
\mathbb{P}(s, p, o |t, {\underline {\underline T}} )
\approx
\mathbb{P}(s, p, o |\mathbf{\hat a}_s, \mathbf{\hat a}_p, \mathbf{\hat a}_o, \mathbf{\hat a}_{t}) .
\end{equation}
We ignore in this notation that due to the softmax transfer function, the conditional probability implicitly depends on the complete embedding matrix. 

The point estimates $\mathbf{\hat a}_s, \mathbf{\hat a}_p, \mathbf{\hat a}_o, \mathbf{\hat a}_t$ can be sensitive to changes in ${\underline {\underline T}}$, implying that there are global dependencies between all triple statements. For example, if a new triple becomes known for the current instance $t$, this information will become part of ${\underline {\underline T}}$ and potentially results in major changes to $\mathbf{\hat a}_{t}$, affecting all triple statements at $t$.

In the following notation
we do not use the hat notation; for example, we write $\mathbf{a}_t$ instead of $\mathbf{\hat a}_t$.

In the sampling mode, we decompose into conditional probabilities and generate samples for binary statements from
\begin{equation} \label{eq:OM*xxx}
 \mathbb{P}(s, p, o | t) = \mathbb{P}(s | t)
 \mathbb{P}(o | s, t)
 \mathbb{P}(p|s, o, t)
 \end{equation}
where $p \in \mathcal{P}_B$ is a binary label.
 For unary statements where we use $p = \textit{hA}$ (\textit{hasAttribute}) and we generate samples from
 $\mathbb{P}(s | t)
 \mathbb{P}(c|s, t)$, where $c$ is the unary label.

\subsection{The BTN Pre-observation and Post-observation Model}

We obtain the pre-observation model (semantic memory) by using the approximation in Equation~\ref{eq:inflatentpoint} but by replacing $\mathbf{\hat a_{t}} \leftarrow \mathbf{\bar a}$. 
 Technically, $\mathbf{\bar a}$ is a hyperparameter vector that was fit on all past data (data pooling). 

The post-observation model  is achieved simply by applying the Dirichlet update Equation~\ref{eq:dirfusdef} to the learned BTN models. Actually, we would propose that the brain generates samples alternately from the observation model and semantic memory, enabling the mind to keep track of what was observed and what was added by semantic memory, representing background knowledge.

\subsection{BTN Architecture}

\begin{figure}[htp]
\begin{center}
\includegraphics[width=1\linewidth]{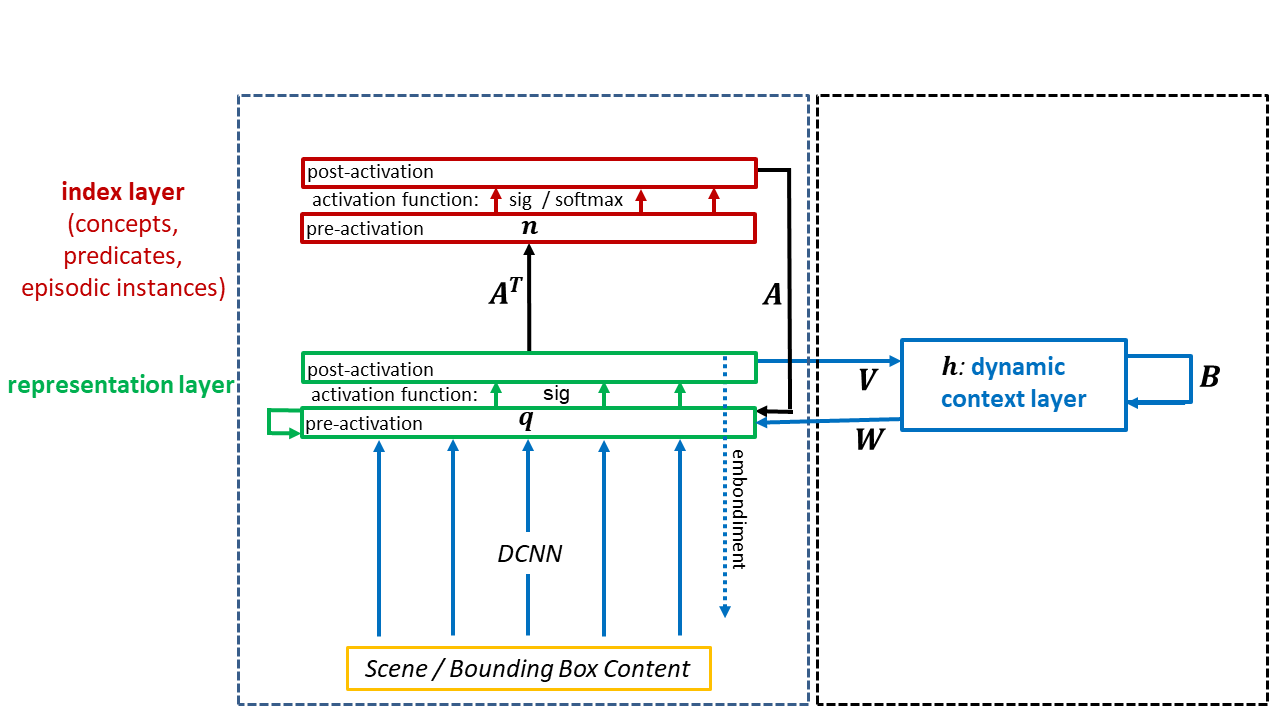}
\end{center}
\vspace{-0.5cm}
\caption{
Perception and memory: Our model architecture consists of two main layers, the \textit{representation layer} ${q}$ and the \textit{index layer} ${n}$. $\mathbf{q}$ and $\mathbf{n}$ are the vectors of pre-activations. 
In perception, the representation layer obtains inputs from the scene. 
Whereas the computational operations in these layers are practically instantaneous, the \textit{dynamic context layer} ${h}$ represents internal brain activities and stores information when attention moves from one entity to another.
The dotted line represents embodiment, the activation of earlier processing layers via the representation layer. 
}
\label{fig-basis-arch}
\end{figure}

\begin{figure}[htp]
\vspace{0cm}
\begin{center}
\includegraphics[width=\linewidth]{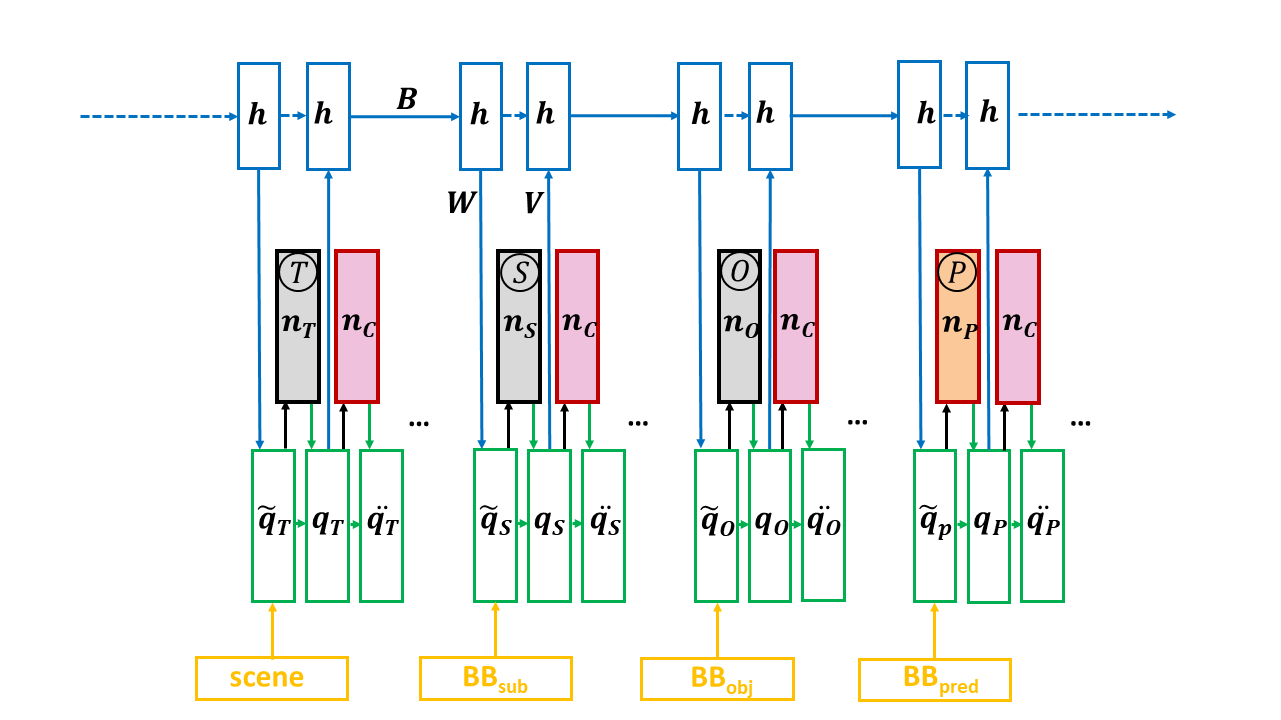}
\end{center}
\vspace{-0.5cm}
\caption{
{Unfolded representation of episodic and semantic memory. }
In perception, the structures in yellow color at the bottom are added; they stand for inputs from the visual scene. 
}
\label{fig-justsem}
\end{figure}

The BTN is implemented by an interaction of two layers, which, as we argue later, might be related to functional brain operations. 
One layer is the representation layer $q$ with pre-activation vector $\mathbf{q}\in \mathbb{R}^r$; here, $r \in \mathbb{N}$ is the embedding dimension, that is, the rank of the BTN approximation. As we propose later, 
$\mathbf{q}$ reflects the cognitive state of the brain (see Figure~\ref{fig-basis-arch}). 
The other one is the index layer ${n}$ with pre-activation vector $\mathbf{n}\in (0, 1)^d$, with $d = N_C + N_P + N_T$. The index layer contains one dimension or unit for each concept, predicate, and episodic instance.
The connection matrix $A$ links both layers, and its columns are the embeddings of the indices.
We also introduce a dynamic context layer $h$ with pre-activation vector $\mathbf{h}$. 
In the following, we describe the implementations of layers and operations using the unfolded view in Figure~\ref{fig-justsem}.

\subsection{BTN Algorithm for Memory Recall}
\label{sec:predmo}

In this section, the embedding vectors contain free parameters optimized for good performance. 
In the following sections, we will ground them in perception, in the spirit of an embodied approach. 

 We start with an instance $t$ to activate episodic memory. 
Thus the activation of the index layer is $\mathbf{n}_T = \mathbf{e}^t$. 
This activates 
$\mathbf{q}_T = \mathbf{a}_t$. $\mathbb{P}(c_{\mathrm{instance}} | t) =
\mathrm{softmax}^{\beta}_{c} (\mathbf{n}_C)$ 
is the probability for a unary label of the instance, for example, a particular vacation day, or a particular location, for example, Paris. 
Here, $\mathbf{n}_C = \mathbf{a}^{\top}_c \mathrm{sig}(\mathbf{q}_T)$.
In our experiments, we do not use instance labels.
We get for the subject
\begin{equation}\label{eq:sample-subject}
\mathbb{P}(s|t) = \mathrm{softmax}^{\beta}_s
(\mathbf{n}_S) 
\end{equation}
with pre-activations $\mathbf{n}_S = \mathbf{a}_s^\top \mathrm{sig} (\mathbf{g}(\mathbf{q}_T))$.% and $\mathbf{q}_T = \mathbf{a}_t$.

For unary entity labels, we obtain
 \begin{equation}\label{eq:tkg-attr-opm}
 \mathbb{P}(c | s, t) 
 =
\mathrm{softmax}^{\beta}_{c} (\mathbf{n}_C)
\end{equation} 
with $\mathbf{n}_C = \mathbf{a}^{\top}_c \mathrm{sig}(\mathbf{q}_S)$.
For the object, we obtain
\begin{equation}\label{eq:sample-object}
\mathbb{P}(o|s, t) = \mathrm{softmax}^{\beta}_o
(\mathbf{n}_O)
 \end{equation}
where $\mathbf{n}_O = \mathbf{a}_o^\top \mathrm{sig} (\mathbf{g}(\mathbf{q}_S))$ and $\mathbf{q}_S = \mathbf{a}_{s} + \mathbf{g}(\mathbf{q}_T)$.
For binary labels, we get
\begin{equation}\label{eq:tkg-rel-opm}
\mathbb{P}(p | s, o, t) 
=
\mathrm{softmax}^{\beta}_p
 (\mathbf{n}_P)
\end{equation}
with $\mathbf{n}_P = \mathbf{a}_p^\top \mathrm{sig} (\mathbf{g}(\mathbf{q}_O))$ and $\mathbf{q}_O = \mathbf{a}_{o} + \mathbf{g}(\mathbf{q}_S)$.

For Boolean variables and unary statement, we obtain 
(cf. Equation~\ref{eq:Eobservation-model})
\begin{equation} %\label{eq:tkg-attr}
 \mathbb{E}(Y_{s, \textit{hA}, c, t}) 
 =
\mathrm{sig} (\mathbf{n}_C)
\end{equation}
and for binary statements, 
\begin{equation} 
\mathbb{E}(Y_{s, p, o, t}) 
=
 \mathrm{sig} (\mathbf{n}_P) . 
\end{equation}
Essentially, we replace the softmax activation function with the sig activation function. This renormalization is discussed in Appendix~\ref{sec:bm}.

For the semantic memory model (i.e., the pre-observation model), we use
$\mathbf{a}_{t} \leftarrow \mathbf{\bar a}$. 
Also, we input $s$.

Here, $\mathbf{g}(\cdot)$ is a nonlinear function. In our approach, we use
\[
\mathbf{g}(\mathbf{q} )
= W \textrm{sig} ( B \textrm{sig} ( V \textrm{sig}(\mathbf{q})))
\]
which performs computations in the hidden layer ${h}$ of the recurrent network. $W, B, V$ are learned matrices. 
Also, $\mathrm{sig}\left(x\right) = 1/(1 + \exp (-x))$ is the logistic function and
\begin{equation*} \label{eq:softmax}
 \mathrm{softmax}_i^{\beta} \left(\mathbf{x}\right)
=
\frac{\exp \beta x_i}
{\sum_{i'} \exp \beta x_{i'}} .
\end{equation*}
Then, the post-activation vector $\mathrm{softmax}^{\beta} \left(\mathbf{x}\right)$ (without the subscript) is the column vector of all $\{\mathrm{softmax}^{\beta}_i \left(\mathbf{x}\right) \}_i$.
Here, $\beta \ge 0$ is an inverse-temperature parameter and can be used for making the response more or less focused. $\mathbf{x}$ is the vector of pre-activations.

\subsection{Discussion on the BTN}
\label{sec:ddis}

An important property of the BTN is that in the algorithmic implementation and at any iteration step, the representation vectors have the dimension of the embedding vector and, 
for example, cannot represent the concatenation of two embeddings. 
We can relate this to brain function. Since the representation layer might occupy a significant portion of the brain, leaving no space for a second concurrent representation, we call this property the ``one-brain hypothesis'' (discussed in more detail in Section~\ref{sec:obhyp}). 
Also, due to the nonlinearity of the function $\mathbf{g}(\cdot)$, one breaks the symmetry between subject and object embedding. It is clear which entity is the subject and which one is the object, and the agent can distinguish between \textit{(Sparky, ownedBy, Jack)} and \textit{(Jack, ownedBy, Sparky)}.
Thus, proper role labeling is performed. \cite{halford2014categorizing} call this ``structural alignment.'' 
We did not select more standard models for several reasons.
First, most tensor factorizations \citep{hackbusch2012tensor}, such as CPD, the Tucker decomposition, the tensor train, and RESCAL \citep{nickel2011three}, require an excessive multiplication of factors; multiplication is an operation that is not easily implemented in biological hardware.
Second, standard tensor networks are functions of several embedding vectors that would need to be presented concurrently; this would violate our one-brain hypothesis.
By representing functions of functions, our model is compositional (in the sense of \cite{poggio2020theoretical}), which is a property that is used to explain the superior performance of deep architectures. 
Other forms of compositionality are discussed in Section~\ref{sec:reas}.

\subsection{BTN Algorithm for Perception}
\label{sec:ppp}

We consider the following setting: At a new episodic instance $t'$, the agent encounters a new $\textit{scene}_{t'}$.
Then a bounding box $\textit{BB}_{\textit{sub}}$ is segmented, whose content describes a visual entity $s'$. The visual entity $s'$ might be a known entity, or it might be a novel entity, not yet known to the agent. 
The agent might also detect a second visual object $o'$ with bounding box $\textit{BB}_{\textit{obj}}$ in the scene and might be interested in its relationship to $s'$.
The content of a third bounding box $\textit{BB}_{\textit{pred}}$, typically encompassing the subject and object bounding boxes, describes the predicate. 
The goal is now to produce statements that are likely true, considering the context of the scene.

We get for the episodic instance
 \begin{equation}
 \label{eq:sstime}
 \mathbb{P}(t'= t| \textit{scene}_{t'}) =
\mathrm{softmax}^{\beta}_t (\mathbf{n}_T )
 \end{equation}
where $\mathbf{n}_T = 
 \mathbf{a}_t^\top
\textrm{sig} ( \mathbf{f}(\textit{scene}_{t'}))$.

We obtain for the subject (cf. Equation~\ref{eq:sample-subject})
 \begin{equation*}
 \label{eq:ss}
 \mathbb{P}(s'= s | {t'=t}, \textit{BB}_{sub}, \textit{scene}_{t'}) 
 =
\mathrm{softmax}^{\beta}_s (\mathbf{n}_S)
 \end{equation*}
 with $
 \mathbf{n}_S = 
 \mathbf{a}_s^\top
\textrm{sig} \left( \mathbf{f}(\textit{BB}_{sub}) + 
\mathbf{g}(
\mathbf{q}_T
)\right)$
 and
 $\mathbf{q}_T = \mathbf{a}_t +\mathbf{f}(\textit{scene}_{t'})$.

For unary entity labels, we obtain
 \begin{equation*}\label{eq:tkg2-attr-opm}
 \mathbb{P}(c'=c | s'=s, {t'=t}, 
 \textit{BB}_{\textit{sub}}, \textit{scene}_{t'})
 =
\mathrm{softmax}^{\beta}_{c} (\mathbf{n}_C)
\end{equation*}
with $\mathbf{n}_C = \mathbf{{a}}_{c}^\top \textrm{sig} (\mathbf{q}_S)$.

We obtain for the object (cf. Equation~\ref{eq:sample-object})
 \begin{equation*} \label{eq:so}
\mathbb{P}(o' = o| s'=s, t'=t, \textit{BB}_{sub}, \textit{BB}_{obj}, \textit{scene}_{t'}) =
\mathrm{softmax}^{\beta}_o (\mathbf{q}_O)
 \end{equation*}
 with $
 \mathbf{q}_O =
 \mathbf{a}_o^\top
 \textrm{sig} (
 \mathbf{f}(\textit{BB}_{obj}) + \mathbf{g}
 (
\mathbf{q}_S
 ) 
 ) $ and
 $\mathbf{q}_S = \mathbf{a}_{s} + \mathbf{f}(\textit{BB}_{\textit{sub}}) + \mathbf{g}(\mathbf{q}_T)$. 
 
 For binary labels, we get 
 \begin{equation*} \label{eq_aipp2}
 \mathbb{P}(p'=p |
o'=o, s'=s, t'=t,
 \textit{BB}_{\textit{sub}}, \textit{BB}_{\textit{obj}}, \textit{BB}_{\textit{pred}}, \textit{scene}_{t'}) = 
\mathrm{softmax}^{\beta}_p (\mathbf{n}_P)
\end{equation*}
with $
\mathbf{n}_P =
 \mathbf{a}_p^\top
 \textrm{sig} 
(\mathbf{f}\left(\textit{BB}_{\textit{pred}}\right) +
\mathbf{g}(\mathbf{q}_O))
$ 
and $\mathbf{q}_O = \mathbf{a}_{o} + \mathbf{f}(\textit{BB}_{\textit{obj}}) + \mathbf{g}(\mathbf{q}_S)$.
 For Boolean variables and for unary statements, we obtain
\begin{equation} 
 \mathbb{E}(Y_{s', \textit{hA}, c, t'} |
s'=s, {t'=t}, 
 \textit{BB}_{\textit{sub}}, \textit{scene}_{t'}) 
=
 \mathrm{sig} (\mathbf{n}_C ) 
\label{eq_aip}
\end{equation}
and for 
binary statements 
 \begin{equation} \label{eq_aipp}
 \mathbb{E}(Y_{s', p, o',t'} |
o'=o, s'=s, t'=t,
 \textit{BB}_{\textit{sub}}, \textit{BB}_{\textit{obj}}, \textit{BB}_{\textit{pred}}, \textit{scene}_{t'}) = 
\mathrm{sig} (\mathbf{n}_P) .
\end{equation}
Here, $\mathbf{f}(\cdot)$ is a representation vector derived from visual inputs realized by a deep
convolutional neural network (DCNN, see Section~\ref{sec:expdisc}).

\section{Algorithmic Implementation and the Attention Approximations}
\label{sec:algimp}

\begin{algorithm}[!htb]
	\SetKwInput{KwIn}{Input}
	\SetKwInput{KwOut}{Output}
 \SetAlgoLined
 \Switch{\textit{Perception}}
{
\KwIn{$\textit{scene}$, $\textit{BB}_{\textit{sub}}$, $\textit{BB}_{\textit{obj}}$, $\textit{BB}_{\textit{pred}}$; $u=1$}
}
 \Switch{\textit{Episodic Memory}}
{
\KwIn{$t_{\textit{in}}^*$; $u=0$}
}
\KwOut{$t^*$, $s^*$, $c^*$, $o^*$, $p^*$
}
$\mathbf{h} = \mathbf{0}$
\Comment{Alternatively, $\mathbf{h}$ is inherited from past decoding}\\

$\mathbf{\tilde q}_T \leftarrow u \mathbf{f}(\textit{scene})$ \Comment{Representation of overall scene}\\

\Switch{\textit{Perception}}
{

 $ \forall t \in \mathcal{T}: 
 n_{T(t)} \leftarrow 
 \mathbf{a}_{t}^\top \mathrm{sig} (\mathbf{\tilde q}_T) $ \\ 
 Sample $t^* \sim \mathrm{softmax}^{\beta} 
\left(\mathbf{n}_{T} \right)$ \label{line-unary-sams.ti} \Comment{Sample a past episodic} \\

}
\Switch{\textit{Episodic Memory}}
{
$t^* = t_{\textit{in}}^*$ \Comment{Input for episodic memory}\\
}

 $\mathbf{q}_T \leftarrow \mathbf{\tilde q}_T + \mathbf{{a}}_{t^*}
$ 
\label{algo:ttt}
\\
 $ \forall c \in \mathcal{C}: 
 n_{C(c)} \leftarrow 
 \mathbf{a}_{c}^\top \mathrm{sig} (\mathbf{q}_T) $ \\

 $\mathbf{h} \leftarrow 
 B \mathrm{sig} 
 [\mathrm{sig}(\mathbf{h}) + V \mathrm{sig} (\mathbf{q}_T) ]$ \\

 $\mathbf{\tilde q}_S \leftarrow
 u \mathbf{f}(\textit{BB}_{\textit{sub}}) +
 W \mathrm{sig} (\mathbf{h}) $ \Comment{$\mathbf{g}(\cdot) = W \mathrm{sig} (\mathbf{h})$}\\

$\forall s \in \mathcal{C}: {{n}}_{S}(s) \leftarrow 
\mathbf{a}_s^\top \mathrm{sig} (\mathbf{{\tilde q}}_{S})$
\label{line-perc-sub} \\ 
Sample $s^* \sim \mathrm{softmax}^{\beta} 
(\mathbf{{n}}_{S})$ 
\Comment{Sample a subject entity} \\
 $\mathbf{q}_S \leftarrow \mathbf{\tilde q}_S + \mathbf{{a}}_{s^*}$ \label{alg:saobx} 
\\

$\forall c \in \mathcal{C}:
n_{C}(c)
 \leftarrow \mathbf{a}_{c}^\top \mathrm{sig} (\mathbf{q}_S) $
 \label{line-perc-sub-attr}

Sample $c^* \sim \mathrm{softmax}^{\beta}
\left(\mathbf{{n}}_{C}\right)$ \Comment{Sample a unary label for the subject} \label{line-unary-sams} \\

 $\mathbf{h} \leftarrow 
 B \mathrm{sig} 
 [\mathrm{sig}(\mathbf{h}) + V \mathrm{sig} (\mathbf{q}_S) ]$ \\

 $\mathbf{\tilde q}_O \leftarrow
 u \mathbf{f}(\textit{BB}_{\textit{obj}}) +
 W \mathrm{sig} (\mathbf{h}) $ \\

$\forall o \in \mathcal{C}: {{n}}_{O}(o)
\leftarrow \mathbf{a}_o^\top \mathrm{sig} ( \mathbf{{\tilde q}}_{O} ) $ \\
Sample $o^* \sim \mathrm{softmax}^{\beta} 
\left( \mathbf{{n}}_{O}\right) $ 
\Comment{Sample an object entity} \\
$\mathbf{q}_O \leftarrow \mathbf{\tilde q}_O + \mathbf{a}_{o^{*}} $ 
\label{alg:sabbx}\\

 $\mathbf{h} \leftarrow 
 B \mathrm{sig} 
 [\mathrm{sig}(\mathbf{h}) + V \mathrm{sig} (\mathbf{q}_O) ]$ \\

 $\mathbf{\tilde q}_P \leftarrow
 u \mathbf{f}(\textit{BB}_{\textit{pred}}) +
 W \mathrm{sig} (\mathbf{h}) $ \\

$\mathbf{q}_P \leftarrow \mathbf{\tilde q}_P$ \\
$ \forall p \in \mathcal{P}^B: n_{P}(p) \leftarrow 
 \mathbf{a}_p^\top \mathrm{sig} ( \mathbf{q}_P ) $ 
 \label{line-perc-relpred} 
 \\
 Sample $p^* \sim \mathrm{softmax}^{\beta} (\mathbf{{n}}_{P})$ 
 \Comment{Sample a binary label} \label{line:ssbinp}\\

\KwRet{$t^*$, $s^*$, $c^*$, $o^*$, $p^*$
}
\caption{\textbf{The BTN for Perception and Episodic Memory.}}
\label{algo:total}
\end{algorithm}

\subsection{Algorithmic Implementation by Stochastic Sampling}
\label{Sec: AImp}

Figure~\ref{fig-justsem} illustrates the unfolded processing steps of the architecture shown in Figure~\ref{fig-basis-arch}.
Algorithm~\ref{algo:total} describes the processing steps implementing 
the Equations in the previous sections by generating samples from the conditional probabilities. 
The main difference between episodic memory and perception is, 
that, in the latter, 
 visual inputs from the contents of the bounding boxes are integrated.
The algorithm outputs $t^*$, $s^*$, $c^*$, $o^*$, $p^*$ 
from which we can form triples of the form 
% $(t', \textit{hA}, c^*_{\mathrm{instance}})$, 
$(s', \textit{hA}, c^*)$ and
$(s', p^*, o')$, but also 
$(s^*, \textit{hA}, c^*)$ and
$(s^*, p^*, o^*)$. 
Here, for example, $s'$ is an entity in the scene, and $s^*$ is an entity with a permanent representation in the index layer.
In perception, the episodic instance $t^*$ would be the index of a similar episode in the past.

In Figure~\ref{fig-justsem}, we indicate that we can also sample 
a unary label for the complete scene or episode
 and for the predicate, which we do not actually do in the experiment. 
We also indicate additional processing steps for calculating embeddings. For example, 
$\mathbf{\ddot{q}}_S \leftarrow \mathbf{q}_S + \mathbf{{a}}_{c^*}$ is the embedding of the sentence \textit{($s^*$, \textit{hA}, $c^*$)} in the context of the scene.
{Whereas concepts, predicates, and episodic instances have static embeddings realized as connection weights linking the index layer and the representation layer, the embedding of a sentence is dynamic.}

In Algorithm~\ref{algo:total}, we explicitly introduce the latent vector $\mathbf{h}$ modelling dynamic context, supporting the implementation of the nonlinear function $\mathbf{g}(\cdot)$.
Table~\ref{tab:firing} illustrates the generation of activated indices (``firing indices or neurons'') in different operational modes. 

The algorithm for semantic memory (i.e., the  pre-observation model) is identical to the one for episodic memory, only that 
$\mathbf{a}_{t^*} \leftarrow \mathbf{\bar a}$. One might also specify $s^* \leftarrow s$ and not sample it.

Sampling is a unique, maybe temporary, commitment to a concept.
In a biological interpretation, a single winning unit (neuron) fires. Even a single or a few spikes can be sufficient for communicating the winning concept to later processing. 
Since sampling commits unique indices $t^*$ and $s^*$, this allows the association of semantic and episodic memory experiences to the observation.

\subsection{Embedded Symbolic Reasoning by Chaining}
\label{sec:chaining}

As indicated by the four groups of dots in Figure~\ref{fig-justsem}, unary decoding can continue during the time of operation of the dynamic context layer. For example, 
if $\mathbf{\ddot{q}}_S$ represents the triples 
\textit{(Sparky, hA, Dog)} and the agent has learned the generalized 
statement \textit{(Dog, hA, Mammal)}, then it is likely that the index for \textit{Mammal} fires next. Thus the sequential index pattern \textit{Sparky, Dog, Mammal} is interpreted by the agent that Sparky is a dog and a mammal.
This form of embedded symbolic reasoning, as further discussed in Section~\ref{sec:embsymreas}, is based on generalized statements discussed in Section~\ref{sec:gensta}.

\subsection{Sampling and the Attention Approximation}
\label{sec:aas}

For episodic and semantic memory, we need to generate triples of the form \textit{($s^*$, hA, $c^*$)}, whereas to label entities in perception, we only need: \textit{($s'$, hA, $c^*$)}, that is, $s'$, the subject in the scene, does not need to be identified as a stored entity $s^*$. For example, if the agent is attacked by a bear, it does not care what the name of the bear might be; it might be a bear not yet known. Our approach is 
motivated by the computational attention approach used in deep learning \citep{vaswani2017attention}. It can be related to the concept of internal attention in cognition to distinguish it from external attention, for example, the attention to a particular perceptual entity.   With parallel hardware (e.g., brainware) the sampling process can be replaced by a computational attention approximation, which can be executed in parallel and does not commit to a particular $s^*$.

The \textit{semantic attention} (SA) approximation for the subject replaces line~\ref{alg:saobx} in 
Algorithm~\ref{algo:total}
with 
\begin{equation*}\label{eq:ea2}
\mathbf{q}_S 
\leftarrow
 \mathbf{\tilde q}_S
+
A \; \mathrm{softmax}^{\beta} (A^{\top} \textrm{sig} (\mathbf{\tilde q}_S) ) .
\end{equation*}
In matrix $A$ we only consider the columns relating to entities. In terms of standard attention, as defined by \citep{weston2014memory,sukhbaatar2015end,vaswani2017attention},
the argument of the softmax calculates the\textit{score vector},
the softmax output is the \textit{alignment vector}, 
the multiplication with matrix $A$ calculates the \textit{context vector}, the columns of $A$ are the \textit{key vectors} and the \textit{value vectors}, and
 $\textrm{sig} (\mathbf{\tilde q}_S)$ is the \textit{query}.
The main differences to the standard approaches to attention is that, in our approach, \textit{key vectors} and \textit{value vectors} are \textit{stored} semantic embeddings and the first term on the right is the pre-activation instead of the post-activation; 
processing several bounding boxes from the same image in parallel, as it is done in visual attention, for example, in~\citep{koner2020relation}, would violate our one-brain hypothesis from Section~\ref{sec:obhyp}. 
 In the sampling mode of operation, a symbolic unit in the index layer competes with other units in the same layer to be activated.
In contrast, in our attention approximation, the index layer
 acts as a standard neural network layer (without sampling) with softmax activation (i.e., the sum in the equation) and pre-activation skip connections ($\mathbf{\tilde q}_S$ in the equation). 
 
In training the model, and also in the attention approximation, we use $\beta = 1$.
{With $\beta = 1$, predictive samples reflect the predictive uncertainty in prediction. 
In the actual sampling experiments}, we set the inverse-temperature parameter $\beta \rightarrow \infty$, effectively taking the concept with the highest probability. We call this winner-take-all sampling.
If there were a dominating dimension or if we would set $\beta \rightarrow \infty$ in the attention approximation, the attention approximation would become identical to winner-take-all sampling.

Similarly, for the object we replace line~\ref{alg:sabbx} with 
\begin{equation*}\label{eq:eao}
\mathbf{q}_O 
\leftarrow
 \mathbf{\tilde q}_O
+
A \; \mathrm{softmax}^{\beta} (A^{\top} \textrm{sig} (\mathbf{\tilde q}_O)) .
\end{equation*}
In matrix $A$ we only consider the columns relating to entities.
Attention can also be applied to episodes in perception. The \textit{episodic
attention} (EA) for the episodic index replaces line~\ref{algo:ttt} 
with 
\begin{equation*}\label{eq:sa}
\mathbf{q}_T 
\leftarrow
 \mathbf{\tilde q}_T
+
A\; \mathrm{softmax}^{\beta} ( A^{\top} \textrm{sig} (\mathbf{\tilde q}_T)) .
\end{equation*}
In matrix $A$ we only consider the columns relating to episodic instances.
EA is the default in all experiments on perception.

 With no scene input, 
$\mathbf{q}_T 
\leftarrow
({1}/{N_T}) \sum_t a_t \approx \bar a$. Thus we can consider semantic memory also as a situation where all instance indices are activated equally. Technically, this is a form of an empirical hierarchical Bayesian approach where the embedding average is inherited by a new instance.

\begin{table}[htp]
 \centering
 \begin{tabular}{|c|c|l|}
 \hline 
 Sequential Index Pattern & Equivalent Triple Statements & Mode \\
 \hline 
 \hline 
 
 \textit{Sparky, Dog, Friendly, Black}
 & (\textit{Sparky, hA, \{Dog, Friendly, Black\}}) & SM:U \\
 \hline 
 \textit{Sparky, Jack, looksAt, ownedBy} & 
\textit{(Sparky, \{looksAt, ownedBy\}, Jack}) & SM:B \\
 \hline 
 \hline
 
 \textit{t, Sparky, Dog, Friendly, Black}
 & \textit{t: (Sparky, hA, \{Dog, Friendly, Black\}})
 & EM: U \\
 \hline
 \textit{t, Sparky, Jack, looksAt, ownedBy}
 & \textit{t: (Sparky, \{looksAt, ownedBy\}, Jack}) & EM: B\\
 \hline 
 \hline

 \textit{Dog, Friendly, Black} 
 & \textit{$t', s'$: ($s'$, hA, \{Dog, Friendly, Black\}})
 & P-Dir/SA: U \\ 

\textit{looksAt, ownedBy} 

 & \textit{$t', s', o'$: (s', \{looksAt, ownedBy\}, $o'$})
 & P-Dir/SA: B \\
 \hline
 
 \textit{Sparky, Dog, Friendly, Black} 

 & \textit{At $t'$: (Sparky, hA, \{Dog, Friendly, Black\}})
 & P-Samp: 
 U \\ 
 
 \textit{Sparky, Jack, looksAt, ownedBy}

 & \textit{At $t'$: (Sparky, \{looksAt, ownedBy\}, Jack})
 & P-Samp: B 

 \\ 
 
 \hline 
 \hline 

 \end{tabular}
 \caption{
 The first column shows the sequence of activated indices (``firing neurons''). The second column shows equivalent triples. The last column shows the operational model: SM (semantic memory), EM (episodic memory), P-Dir (direct perception), P-SA (semantic attention), and P-Samp (Perception with sampling). 
 We indicate that for given episodic instances, several unary labels (U) and binary labels (B) can be generated. If we extend the knowledge graph to also contain nodes for predicates and time instances, then, in sampling, exactly one node is active at a time.
}
 \label{tab:firing}
\end{table}

\section{Perception, the Representation Layer, and the Dynamic Context Layer}
\label{sec:expdisc}

In this and the following sections, we present experimental results. 
Here, we focus on perception, and in the following sections on engrams, semantic memory, reasoning, language,
episodic memory, and self-supervised learning. 
Intertwined with the experiments, we make the connection to cognition and neuroscience.

There is a long tradition in cognition and neuroscience to distinguish between anatomical brain structure, for example, the actual anatomical structure of the biological neural network, and functional architectures of the cortex \citep{friston1995characterising,van2013network,bassett2017network,sporns2018graph,leopold2019functional}. 
We discuss the functional architectures primarily but occasionally also consider structure. 

\subsection{Data Set}
\label{sec:data}

We tested our approach experimentally using an augmented version of the VRD data set \citep{lu2016visual}. In the past, this data set has been the basis for much research on visual relationship detection.
Each visual entity is labeled as belonging to one out of 100 classes.
There are 70 binary labels, with 37,993 binary statements total.
We followed other work and assigned 4000 images to the training set and 1000 images to the test set.
The training images contain, overall, 26,430 bounding boxes, thus on average, 6.60 per image.

\begin{figure}[htp]
\centering
 \includegraphics[width=\textwidth]{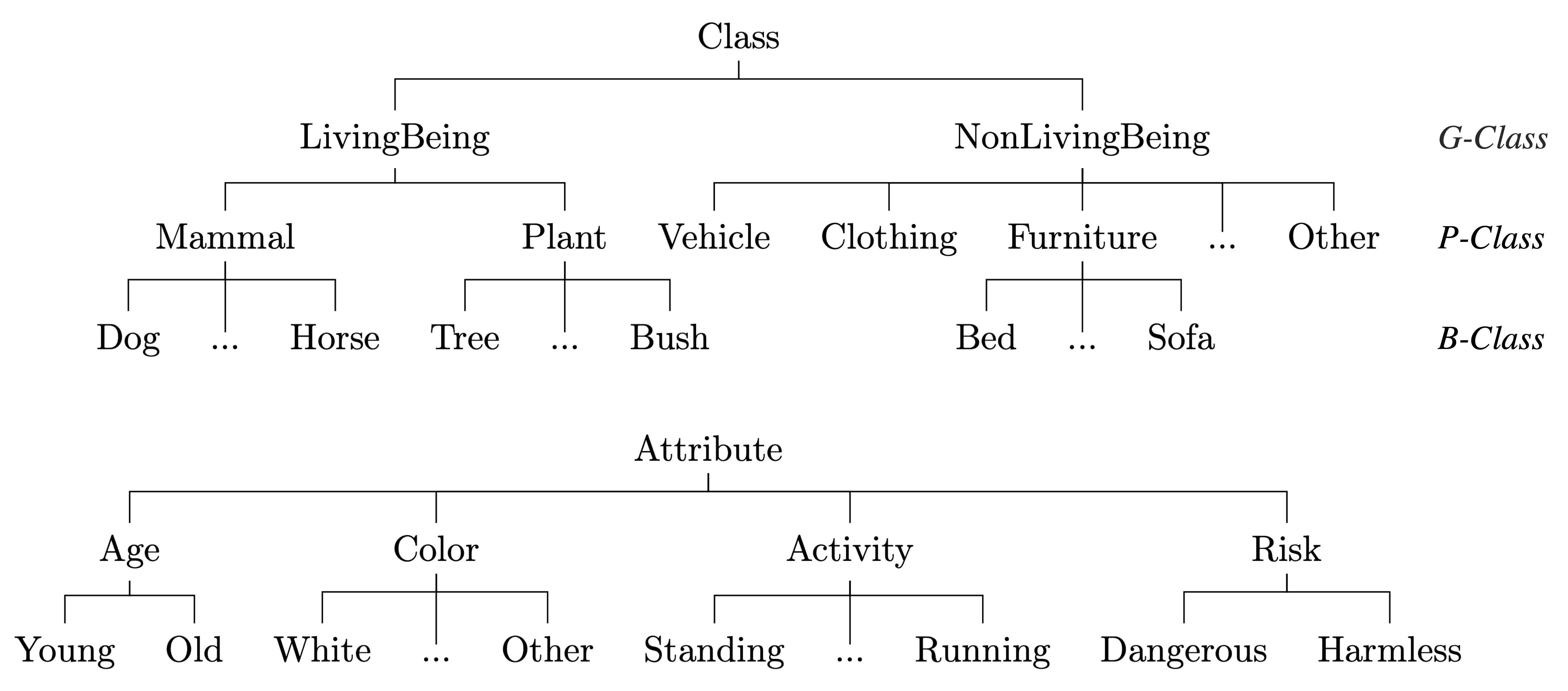}
 \caption{Ontologies. (Top) Class ontology. (Bottom) Attribute ontology.}
 \label{fig:ontology}
\end{figure}

We generated a first derived data set, VRD-E (for VRD-Entity), with additional labels for each visual entity.
First, each entity in each image obtains an individual entity index (or name).
The 26,430 bounding boxes in the training images describe as many entity indices.
Second, we used concept hierarchies from WordNet (\cite{fellbaum2010wordnet}; see Figure~\ref{fig:ontology}).
Each entity is assigned exactly one basis class (or \textit{B-Class}) from VRD (e.g., \textit{Dog}), one parent class (or \textit{P-Class}, for example, \textit{Mammal}), and one grandparent class (or \textit{G-Class}, for example, \textit{LivingBeing}).
At any level, we use the default class \textit{Other} for entities that cannot be assigned to a WordNet concept.
We perform subclass reasoning in the training data and label an entity with its entity index, its \textit{B-Class}, \textit{P-Class}, and \textit{G-Class}. 

In addition, we used pretrained attribute classifiers \citep{Anderson2017up-down,wu2019detectron2} to label visual entities using the attribute 
ontology shown in Figure~\ref{fig:ontology}. 
 Each visual entity obtains exactly one color (including the color \textit{Other}) and exactly one activity attribute, for example, a person can be standing or running. 
 We also introduce the unary labels \textit{Young} and \textit{Old}, which are randomly assigned, such that these can only be predicted for test entities that already occurred in training, not for novel entities.

Furthermore, we introduce the nonvisual, or hidden, unary label \textit{Dangerous} to all living things and \textit{Harmless} to all nonliving things. We do not use these labels in perceptual training; we use them instead to demonstrate how semantic memory can supplement nonvisual labels. 

In summary, every visual entity receives one entity index and, in addition, seven positive unary labels--- for example,
\textit{Entity=Sparky, B-Class=Dog, P-Class=Mammal, G-Class=LivingBeing, Age=Young, Color=Black, Activity = Standing, Risk=Harmless}.

\begin{figure}[htp]
 \centering
 \includegraphics[trim= 0 4.5cm 0 4.5cm, clip, width=\textwidth]{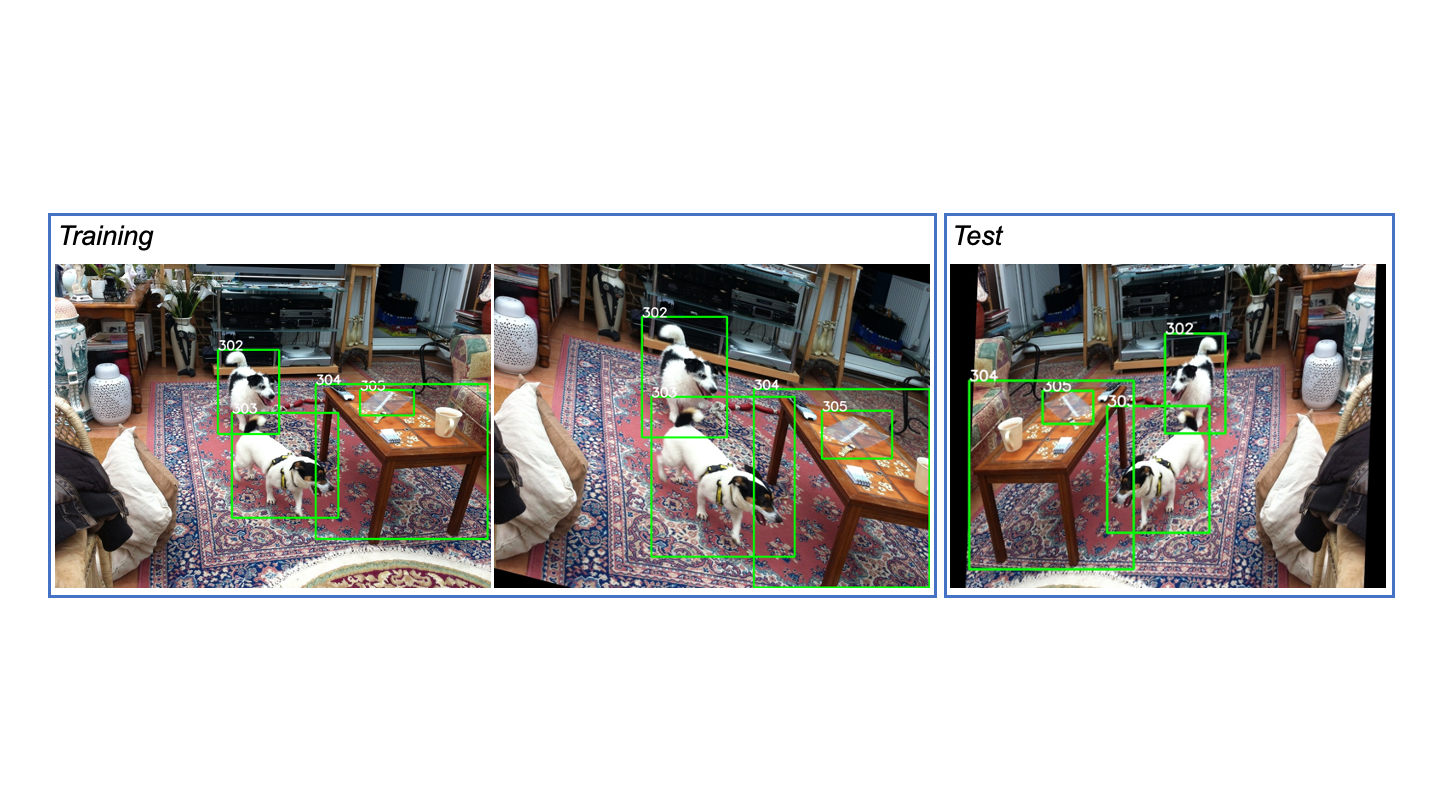}
 \caption{Generation of VRD-EX images.
 (Left) An original VRD image. (Center) A distorted VRD-EX image that is assigned to the training set. (Right)
 A distorted VRD-EX image that is assigned to the test set.}
 \label{fig:training_sample}
\end{figure}

Based on the VRD-E data set, we generate the VRD-EX data set.
Here, we distort each image in the training data set, which generates another 4000 training images\footnotemark.
We used an open library to distort the image \citep{imgaug}. 
To obtain a distorted image, we apply a sequence of transformations, including translation, rotation, shearing, and horizontal flipping of the original image \citep{bloice2017augmentor}. Each transformation is associated with a probability of actually using it. When the operation has coefficients, such as the displacement of translation, a random value within a reasonable range is generated. Thus VRD-EX has 8000 training images. Then we perform another distortion on each original image and create a set of another 4000 test images.\footnote{Due to distortion, some objects and images are discarded, resulting in a reduced number of samples.}
Note that in these new 4000 test images, every visual entity has already occurred in the training set twice
(See Figure~\ref{fig:training_sample}).

In addition to the visual concepts, we introduce nonvisual entities, which do not occur in any image.
In the experiments, we relate visual entities to those hidden entities, for example, by the predicate \textit{ownedBy} or the predicate \textit{lovedBy}.
For example, each visual dog is owned by a person who is not in any scene. Table ~\ref{tab:data} shows the overall statistics.

\begin{table}[ht]
 \centering
 \resizebox{\textwidth}{!}{
 \begin{tabular}{c c c c c c c}
 data set & Training & Test & \#BB & \#VisEnt& \#BinStat & \# Attr/Ent \\
 & Images & Images & Train & Train & Train & Train \\ \hline
 VRD \citep{lu2016visual}
 & 4000 & 1000 & 26430 & 26430 & 30355 & 1\\
 \textbf{VRD-E}ntity & 4000 & 1000 & 26430 & 26430 & 30355 & 8\\
 \textbf{VRD-EX}tended & 7737 & 3753 & 50910 & 26430 & 59095 & 8 \\
 \hline
 \end{tabular}}
 \caption{Statistics of the data sets VRD-E and VRD-EX. Attr/Ent stands for the average number of unary labels per episodic instance. Overall, 15\% of the labels are \textit{Others}. 
}
 \label{tab:data}
\end{table}

Training was performed using the Adam optimizer \citep{kingma2014adam}. For evaluation, we consider top-1 accuracy for unary labels and Hits@k for binary labels and sampling of entities. Hits@k is the fraction of correct entities that appear in the top-k-ranked entities. 

\subsection{DCNN}

As discussed, we use a deep convolutional neural network (DCNN) for the mapping $\mathbf{f}(\cdot)$ from scene and bounding box content to the representation layer. 
 \cite{kriegeskorte2018cognitive} and others have discussed how DCNNs might represent functional modules in the brain. This is not the topic of this article. Our approach is object-centric. 
 The specific DCNN architecture we use is the VGG-19 architecture \citep{simonyan2014very}, pre-trained on ImageNet \citep{russakovsky2015imagenet} and fine-tuned to our data. The VGG-19 architecture constitutes the backbone layer of the Faster R-CNN \citep{ren2015faster}, which we employ to produce a set of bounding boxes as output, where each bounding box contains an object. 
 (For more details, see the Appendix~\ref{sec:appImpl}.) In all of our experiments, we used rank $r = 4096$ as the dimension of the representation layer.

\subsection{Direct Perception}
\label{sec:semdec}

In our approach, four bounding boxes are analysed: first, a bounding box describing the complete scene; second, a bounding box describing the subject entity; third, a bounding box describing the object entity; and finally, a bounding box describing the predicate. In \textit{direct perception}, we assume that labels are independently generated by perception, purely based on bounding box content, and entirely in bottom-up mode. There are connections from the representation layer to the index layer but not in the opposite direction. Also, there are no connections between the representation layer and the dynamic context layer. 

From a perceptional view, this approach exploits neither dependencies between labels for the same bounding box (that a detected Sparky is known to be a black dog), nor dependencies between labels for the different bounding boxes. 
Direct perception requires concept indices, together with their associated embedding vectors, which become the connection weights from the representation layer to the index layer. 
Without connections from the index layer back to the representation layer, no episodic memory or semantic memory can be formed. 
In direct perception, it is unclear how the overall brain is informed about winning or likely concept labels. 
This is one of the issues addressed in BTN perception, discussed next. 

Table ~\ref{tab:simple_decoder} shows results for unary labels on direct perception (rows labeled P-Direct).
The results on VRD-E are pretty competitive.
The performance is improved in VRD-EX, likely due to some form of memorization by overfitting since different views on the same entities occur in the training set and the test set. 
The results on the prediction of binary labels (see Table~\ref{tab:complex_perception}) are not very good; as expected, and a well-known result from scene graph analysis, information on the subject and the object bounding boxes is required for binary label prediction.

\subsection{BTN Perception}

In perception with the BTN, we add connections from the index layer back to the representation layer. We also connect the representation layer with the dynamic context layer. 
This has several important benefits. 

{First, in } a top-down mode of operation, the representation layer, and thus the whole brain, is informed about which concepts are detected in the scene.
{Thus, if an entity is labeled as being \textit{Dangerous} or recognized as being \textit{Jack}, this information is communicated to the whole brain. This pattern completion process is a property typically discussed in the context of associative memories. } 
If this affects earlier processing layers, this process is referred to as embodiment. 
Some theories on embodiment assume, as the name indicates, that thought and language influence bodily states, or at least the brain regions mapping directly to bodily states~\citep{lakoff1999review}.

Second, labels would get biased: If \textit{Sparky} is detected in the scene, it will bias the unary labeling toward the unary label \textit{Black} if this is known from semantic memory to be true. 

Third, by connecting the representation layer with the dynamic context layer, we obtain a memory that can transport information from the analysis of one bounding box to another. This change of attention from one bounding box to another might be associated with saccadic eye movement, where the dynamic context layer stores information during the transition phase. The dynamic context layer is crucial for evaluating triples involving binary labels.

Finally, the bidirectional connectivity between index layer and representation layer enables the formation of episodic and semantic memory and can enrich perception with information from either, as will be discussed in the following sections. 

One can think of perception as a triple-generating language model in the context of visual inputs. Through the top-down information flow via the representation layer, the whole brain is kept up to date about scene content, about which concepts have been detected in the scene and which statements describe scene content. 

The representation layer reflects the context. 
For example, if Sparky is detected in the subject bounding box, then $\mathbf{q}_S$ is an embedding of Sparky in the context of the scene. 
If Sparky in the scene is labeled as \textit{Friendly}, then $\mathbf{\ddot q}_S$ would be the embedding of the statements \textit{(Sparky, hA, Friendly)} in the context of the scene.
If Sparky looks at Jack in the scene and this is detected, then $\mathbf{q}_P$ would be the embedding of the statement \textit{(Sparky, looksAt, Jack)} in the scene's context.

\subsection{Experiments with BTN Perception}

In all experiments, we use episodic attention (EA): a sampling of a past episodic instance $t^*$ is used only to evoke a related episodic memory, as discussed in Section~\ref{sec:cogmem-em}. 

Experiments with semantic attention are labeled P-SA where there is no commitment to a specific $s^*$ or a specific $o^*$. In semantic attention, one is primarily interested in fast labeling rather than in associating memories.

The results in Table~\ref{tab:simple_decoder} show that on the VRD-E data set with novel entities in the test set, perception with semantic attention (P-SA) shows the best results. However, direct perception (P-Direct) is quite competitive. 

On the VRD-EX data set with known entities in testing, perception with sampling, P-Samp, is best, where the algorithm could ``remember'' past encounters of the same entities. Here, P-Direct is not competitive. 
For P-Samp, we use winner-take-all sampling. 

Table~\ref{tab:complex_perception} shows results from binary label prediction. For the binary labeling with novel entities on the VRD-E data, P-SA shows the best performance. The inferior performance of P-Direct confirms that the dynamic context layer is important for achieving good performance. We see a large improvement for the VRD-EX data set.

\begin{table}[htp]
 \centering
 \begin{tabular}{c | c | c c}
 \hline
 & & \multicolumn{2}{c}{Binary labels}\\
 & Model & @10 & @1 \\ \hline
 VRD-E & P-Direct & 85.45 & 31.68 \\
 & P-Samp & 90.39 & 45.09 \\
 & P-SA & 91.33 & 46.84 \\
 \hline
 VRD-EX & P-SA & \textbf{99.58} & \textbf{80.63} \\
 \hline

 \end{tabular}
 \caption{Binary label prediction in perception. 
 As the inferior results for P-direct indicate, the dynamic context layer is essential to perform well in binary label prediction ({rows 1-3}). 
 For known entities (VRD-EX) ({fourth row}), entity indices permit some memorization and improve performance. 
 Best results in each column are in bold.
}
 \label{tab:complex_perception}
\end{table}

In Table~\ref{tab:comparison_methods}, we compare our model with other visual relationship detection methods. In the methods from the literature, only binary labels are tested, with one class label for each entity. In our framework, this would be the decoding of 
generalized statements.
The table demonstrates that generalized statements (i.e., probabilistic rules) can be generalized themselves.
In a zero-shot situation, where the triple \textit{(subject-class, predicate, object-class)} never occurred in training, our approach achieved a recall score of 81.61\%, which is much better than the result from the BFM \citep{baier2017improving} with 76.05\%.
In summary, our approach gives competitive results overall and superior results for zero-shot binary labeling.

\begin{figure}[htp]
\centering
 \includegraphics[trim= 0 3.5cm 0 2cm, clip, width=\textwidth]{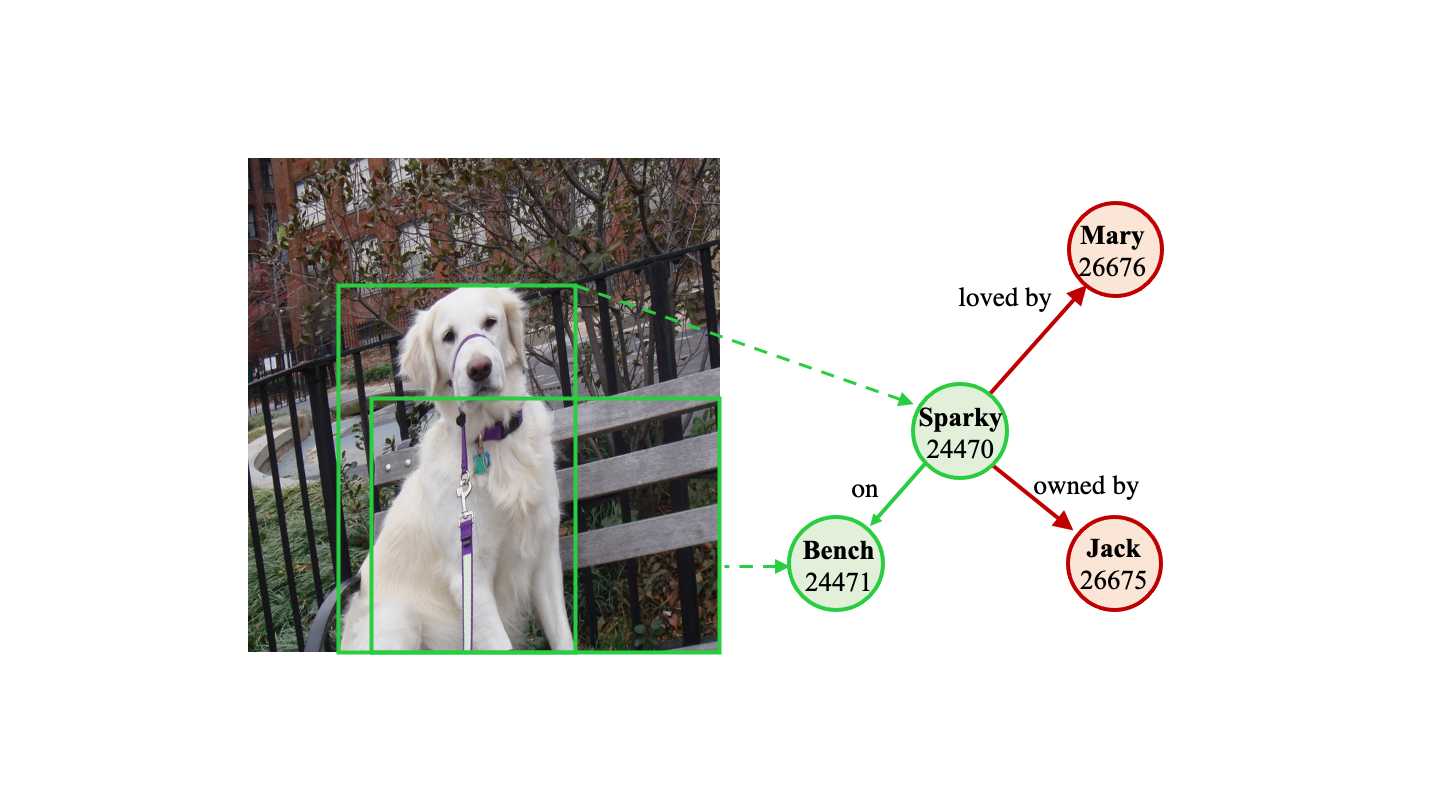}
 \caption{
 Illustration of perception with known entities ({VRD-EX}). The left bounding box is identified (sampled) as $s' = s^* = 24470 = \textit{Sparky}$.
 Highly ranked unary labels are: \textit{Dog, Mammal, LivingBeing, Young, White, OtherActivity}. 
 Highly ranked unary labels for the second bounding box ($24471$) are: 
 \textit{Bench, Furniture, NonLivingBeing, Old, OtherColor, OtherActivity}. Sampled binary statements are: 
 \textit{(Dog, sitsOn, Bench), (LivingBeing, on, Furniture), (Mammal, sitsOn, Old), (White, sitsOn, Bench)}. 
 We also indicate how semantic memory can support
 perception by adding binary statements to object entities, {not in the scene:} \textit{(Sparky, ownedBy, Jack(26675)), (Sparky, lovedBy, Mary(26676))}, where \textit{Jack} and \textit{Mary} are not on the scene, but in the agent's semantic memory.
 The semantic memory experience is further discussed in Section~\ref{sec:conceptorg}. 
}
 \label{fig:triple_example}
\end{figure}

Figure~\ref{fig:triple_example} illustrates perception with unary labels and visual and nonvisual binary statements. 

\begin{table}[ht]
 \centering
 \resizebox{1\textwidth}{!}{
 \begin{tabular}{c | c | c c c c c c c c}
 \hline
 & & \multicolumn{8}{c}{Unary labels (accuracy)} \\
 & Model & Entity & B-Class & P-Class & G-Class & Y/O & Color & Activity & Average\\ \hline

 VRD-E &P-Direct & - & 81.30 &88.44 &94.67 &\textbf{50.60} &\textbf{69.06} &\textbf{83.73} &77.97\\
% & P-noI & - & 81.68 &88.49 &94.61 &50.17 &67.02 &83.95 &77.65 \\
 &P-Samp& - & 80.91 &88.85 &95.02 &47.88 &68.02 &82.90 &77.26	\\
 &P-SA& - & \textbf{81.80} &\textbf{89.34} &\textbf{95.44} &49.46 &68.93 &83.58 &\textbf{78.09} \\

 \hline
 VRD-EX & P-Direct & 92.20 &90.87 &93.54 &96.47 &76.54 &84.49 &92.77 &89.56 \\
% &P-noI & - &93.03 &94.97 &97.21 &81.39 &87.90 &94.23 &91.94\\
 &P-Samp & \textbf{92.81} &\textbf{96.47} &\textbf{97.49} &\textbf{98.47} &\textbf{94.58} &\textbf{94.08} &\textbf{97.59} &\textbf{95.93}\\
 &P-SA & 92.65 &95.27 &97.11 &98.20 &93.10 &92.71 &96.83 &95.12\\

 \hline

 \end{tabular}}
 \caption{Unary label prediction in perception. We evaluate \textit{(s', hA, c)} for the entities $s'$ in the scene.
 The columns labeled ``B-Class, P-Class, G-Class'' are class labels, and the columns labeled ``Y/O, Color, Activity'' are unary labels. 
 ``Average'' refers to the average over all columns.
In P-Direct, there are no links from $n$ to $q$, and $q$ and $h$ are independent. 
 P-Samp stands for the sampling approach using the index $s^*$ with maximum activation. 
 P-SA uses semantic attention. 
 On the VRD-E data set ({rows 1-3}), where each entity is novel, P-SA shows the best results. Not surprisingly, P-Direct is also quite competitive. 
 On the VRD-EX data set ({rows 4-6}), where each entity is known, P-Samp shows the best results since it can recognize specific entities. In particular, the unary label Y/O can only be learned for already known entities.
 Not surprisingly, P-Direct is significantly worse since it cannot benefit from memory, although overfitting leads to better results than with the VRD-E data. 
 The column labeled ``Entity'' evaluates if the right entity is recognized. Thus we evaluate \textit{($s'$, sameAs, s)} for the entities $s'$ in the scene. 
 Best results in each block are in bold.
}
 \label{tab:simple_decoder}
\end{table}

\begin{table}[htp]
 \centering
 \begin{tabular}{c | c | c | c | c}
 \hline
 Model & ph & z-s-ph & rl & z-s-rl\\ \hline

 P-SA & 24.50 & 8.55 & \textbf{93.99} & \textbf{81.61} \\
 P-Direct & 13.54 & 5.73 & 84.64 & 68.35\\
 BFM \citep{baier2017improving} & \textbf{25.11} & 7.96 & 93.81 & 76.05\\
 Approach in \citep{tresp2017tensor} & 23.45 & \textbf{10.95} & 93.32 & 78.79 \\
 \hline

 \end{tabular}
 \caption{Binary label prediction in perception. We compare binary labeling with methods published in the literature using the original VRD data set. 
 \textit{Phrase (ph}) shows the recall of binary labels, where also the extracted bounding box contents are evaluated. The
 \textit{binary (relationship) label (rl)} shows the recall of the binary label given the ground truth class of subject and object.
 \textit{z-s-ph} and \textit{z-s-rl} denote zero-shot performance for triples that did not occur in the training set. Our proposed model, P-SA ({first row}), is superior in the last task. 
 The much better performance of P-SA compared to P-Direct ({second row}) demonstrates the importance of the dynamic context layer. The {third and fourth rows} show results from approaches published in the literature.
 Best results in each block are in bold.
}
 \label{tab:comparison_methods}
\end{table}

\subsection{Representation Layer}
\label{sec:repl}

In our model, $\mathbf{q}$, that is, the activation of the representation layer, represents the cognitive brain state.
In cognitive neuroscience, the representation layer is referred to as ``mental canvas'' or the ``global workspace,'' enabling a global information exchange. It is also known as ``theater of the brain,'' ``communication platform,'' ``communication bus,'' or ``blackboard.'' It is assumed to have a distributed representation involving large parts of the brain.  \citep{binder2011neurobiology} states: ``The neural systems specialized for storage and retrieval of
semantic knowledge are widespread and occupy a large proportion of the cortex in the human brain.''
During perception or memory recall, the representation layer integrates information and then makes this information available to the brain as a whole. In particular, the embedding vector of the episodic experience, that is, $\mathbf{a}_t$, represents a holistic, integrative view of the cognitive state at that instance. 

Since the world of an agent mostly changes smoothly, one can develop models to forecast embedding vectors of future instances \citep{tresp2015learning,han2020graph}. The brain's cognitive state
is, to a large degree, determined by perception, which might explain a personal feeling of personal instability and sensitivity to external influences, whereas other individuals, represented by their embedding vectors, appear stable and slowly changing.

\subsection{Dynamic Context Layer}
\label{sec:wom2}

To really capture the content of a scene, it is essential to understand the relationships between the scene entities; this requires an additional storage facility since, following our one-brain hypothesis, the brain possesses only exactly one global representation layer (see the discussion in Section~\ref{sec:obhyp}). 
As illustrated in Figures~\ref{fig-basis-arch} and~\ref{fig-justsem}, we propose that this functionality is represented by the dynamic context layer ${h}$, which represents the state of some of the brain modules not explicitly discussed in this article. It receives input from the representation layer, processes that information as a recurrent neural network, and then returns the processed information to the representation layer, which starts processing visual input from a subsequent bounding box. Thus, the dynamic context layer stores and processes information between saccadic eye movements. 
The dynamic context layer might also involve higher brain regions, like working memory.

\begin{figure}[htp!]
\centering
 \includegraphics[trim= 0 0cm 0 0cm, clip, width=0.8\textwidth]{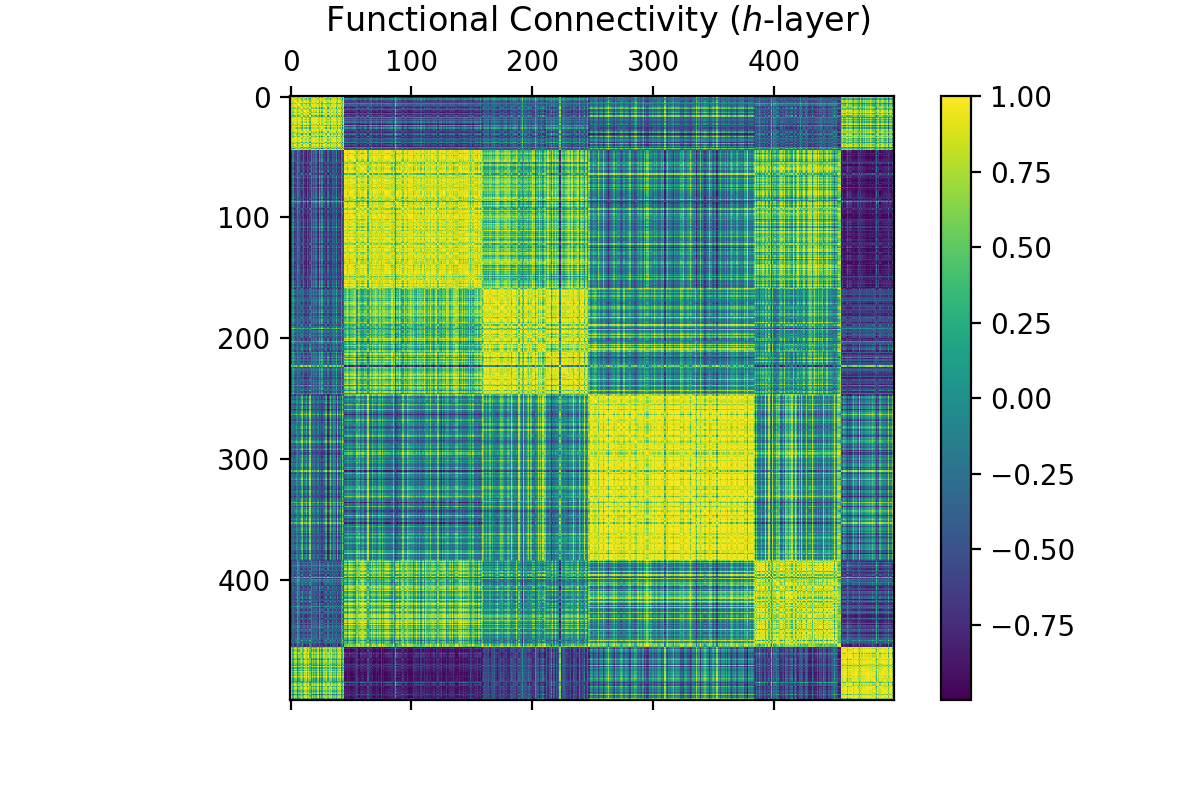}
 \caption{Pearson correlation between units in the dynamic context layer, followed by spectral clustering to order columns and rows. 
}
 \label{fig:h-layer}
\end{figure}

Figure~\ref{fig:h-layer} shows the Pearson correlation between activities of units in the dynamic context layer. It shows a certain clustering structure, as exhibited in recent studies on functional and structural brain connectivity (e.g., Figure~2 in \cite{pope2021modular}). In the brain, such a block structure is a signature of a small-world architecture. 
Since it is not clear what the dynamic context layer in our model actually represents in the brain, we refrain from a deeper discussion of this apparent similarity. 

The dynamic context layer also represents internal mental states that might not be driven directly by perception and memory.
In neuroscience, the default network is assumed to be active when a person is not focused on the outside world; instead, the brain is at wakeful rest, such as during daydreaming or mind-wandering \citep{gazzaniga2004cognitive}. 
 Perception, memory recall, the initiation of activity, and other causes can induce the brain to leave the default state.
In the default state of our model, the brain's representation layer interacts with the dynamic context layer, without being driven by perception. 

\section{Semantic Memory Engrams, Semantic Decoding, and Multimodality}
\label{sec:conceptorg}

\subsection{Background on Semantic Memory}

According to \citep{tulving1985elements}, the semantic memory experience is independent of a particular episodic experience. 
It developed out of perception as an emerging property where semantic enrichment became independent of perceptual input. Thus, in the transition from episodic memory to semantic memory, provenance is lost. 
It is the longest-lasting and most durable memory since it models the stable statistics in the world. It aggregates information and is a dictionary view of the agent's life experience. 
Our approach explains the remarkable similarity between episodic and semantic memory: Semantic memory models the episodic memory of a future instance. Whereas an episodic memory experience requires the activation of 
the embedding vector of the corresponding instance in the representation layer, a semantic memory experience requires the activation of
the semantic memory representation $\mathbf{\bar a}$.
Semantic memory can also be realized by the activation of all instance indices in the episodic attention approximation (see Section~\ref{sec:aas}).  Another alternative is represented in \citep{tresp2017embedding} where semantic memory is generated by averaging in embedding space. But that approach is restricted to   multilinear  models.

\subsection{Neuroscience Perspective: Semantic Memory Engrams}

Engrams are memory traces in the brain \citep{ralph2017neural}. 
{In our approach, the combination of an index with its embedding vector would be an engram for that concept, although we will typically only refer to the latter part as the engram. 
An index is a symbol for a concept, whereas embeddings are part of an implicit concept memory and provide symbol grounding~\citep{harnad1990symbol,barsalou2008grounded}.}

Here, we are in agreement with several theories on semantic memory engrams from neuroscience. For example, \cite{binder2011neurobiology} state that semantic memory engrams consist of both modal and amodal representations, supported by the ``gradual convergence of information throughout large
regions of temporal and inferior parietal association cortex.'' 
The amodal representations are described as a high-level convergence zone.
In our approach, the embeddings would be the modal distributed representations and the indices the amodal, local, and symbolic representations. 

The relationship between the index layer and the representation layer reminds one of the hub-and-spoke model \citep{ralph2017neural}.
The hub is supposed to be located in the anterior temporal lobes (ATLs), which might be where concept indices are consolidated. 
The hub is connected to several different areas (e.g., visual cortex, auditory cortex, orbitofrontal cortex), which might be part of the biological realization of the representation layer. 
Other hubs might be located in the parietal and the temporal lobe \citep{binder2011neurobiology} and maybe in the frontal lobe \citep{tomasello2017brain}.

Some works propose that the anatomical distinction between the representation layer and the index layer might be blurred in the brain.
{One should instead assume an ``{interactive continuum of hierarchically ordered neural ensembles}, supporting progressively more combinatorial and idealized representations'' \citep{binder2011neurobiology}.}

\subsection{Neuroscience Perspective: Indices}

Indices are explicit concept representations, that is, an index is a symbol for a concept and could be implemented functionally by a single unit. In the sampling mode of operation, a symbolic unit in the index layer competes with other units in the index layer. 
This is in contrast to the representation layer, where the activations of the ensemble of units contribute to the processing. 
 
We are purposely imprecise about how exactly an index is represented anatomically (i.e., structurally), in brainware.
In the one extreme, an index might be a single neuron, realizing localist codes, where neurons respond highly selectively to single entities ({``grandmother cells''}). The other extreme are densely distributed codes where items are encoded through the activity of many neurons \citep{kumaran2016learning}.
Most researchers favor a sparse population of cells, realizing a population code. A distributed representation might facilitate the consolidation of new information (see Section~\ref{sec:coglearn}).

The debate about localized representations in the brain is ongoing.
Specific concept cells have been found in the medial lobe (MTL) region of the brain. MTL includes the hippocampus, along with the surrounding hippocampal region consisting of the perirhinal (``what'' path), parahippocampal (``where'' path), and entorhinal neocortical regions. 
Researchers have reported on a remarkable subset of MTL neurons that are selectively activated by strikingly different pictures of given individuals, landmarks, or objects and, in some cases, even by letter strings with their names \citep{quiroga2012concept,quiroga2005invariant}.
Naturally, locality of representation is probably discovered only in well-designed experiments.
In our model, an activated concept index activates many units in the representation layer, and a unit in the representation layer in turn activates many indices. Since index activations might change rapidly, the general appearance might be that of a globally activated system, hiding the locality of representation.
An index is a focal point of activity, but it is never active in isolation.

\begin{figure}[htp!]
\centering
 \includegraphics[trim= 0 0cm 0 0cm, clip, width=0.9\textwidth]{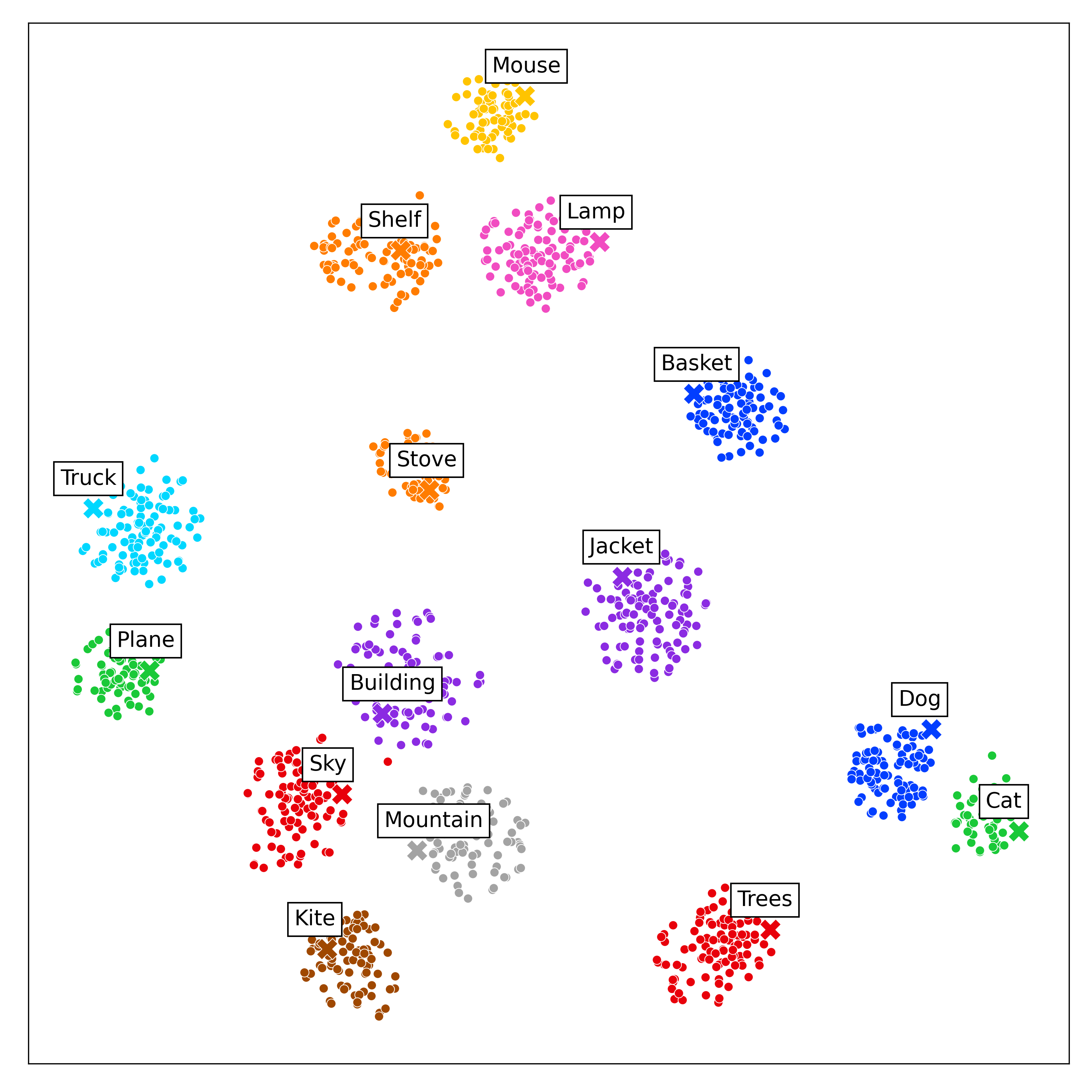}
 \caption{
t-SNE visualization of embeddings of 15 classes and randomly selected entities from these classes. 
Features for the analysis are respective embedding vectors. A dot stands for an entity, for example, a specific dog, \textit{Sparky}. The color of a dot marks the basis class of that entity. 
A cross stands for a class, for example, \textit{Dog}, which is labeled by the same color as entities belonging to the class.
The embedding vector of a class concept lies within the cluster of the classes' entities. Recall that we are not indicating cluster centers but the embeddings that happened to be learned for the class concepts in learning.
In cognitive neuroscience, the semantic embedding space is sometimes referred to as conceptual space, where points denote objects, and regions denote concepts \citep{gardenfors2016geometry}. 
}
 \label{fig:tsne_visualization}
\end{figure}

\subsection{Neuroscience Perspective on Concept Grounding}

Our discussion can be related to recent results from neuroscience where, in different brain regions, maps have been discovered that code, for example, for visual appearance, sound, and function.
For example, if the concept \textit{Hammer} is activated in the index layer, it might excite brain areas indicating a typical hammer appearance, the sound of hammering, and the required motor movement of hammering, all represented in the biological equivalence of the representation layer \citep{rueschemeyer2010function}.
Another example is the concept \textit{Cat}, which includes the information that a cat has four legs, is furry, meows, can move, or can be petted \citep{kiefer2012conceptual}.
In terms of \citep{binder2011neurobiology}, the embeddings are modality specific (here: visual), and the indices represent
convergence zones. As \citep{dehaene2014consciousness} puts it: ``Every cortical site holds its own specialized piece of knowledge.''

{Evidence for distributed semantic activation} has also been described by \citep{Huth2016,de2017hierarchical}.
Both papers developed a detailed atlas of semantic categories and their topographic organization through extensive fMRI studies, showing the involvement of the lateral temporal cortex, the ventral temporal cortex, the lateral parietal cortex, the medial parietal cortex, the medial prefrontal cortex, the superior prefrontal cortex, and the inferior prefrontal cortex \citep{Huth2016}.

Recently, it has been proposed that embeddings are context-dependent \citep{popp2019processing}. Our model suggests that concept embedding is rather stable but that the induced activation of a concept in the representation layer is superimposed with context information, as discussed throughout the article. The representation layer is activated by sensory input and activated concept indices, so the model is informed about a concept in context, even with the connection weights between index and representation layer being fixed.

\subsection{Semantic Memory Engram in the BTN}
\label{sec:prior}

In perception, the embedding vectors determine which indices are activated by the representation layer in a bottom-up operation. 
However, the operation is bidirectional: An index can activate the representation layer by its embedding, which might even activate earlier perceptual layers. We call this top-down mode the grounding or the embodiment of that concept.
Since an embedding vector is optimized for a concept's role in perception and memory, it implicitly reflects all background that is known about it. 

Consider Equation~\ref{eq_aip}, which labels entity $s$ with unary label $c$. 
With no visual inputs, $\mathbf{{a}}_{c}^\top \textrm{sig} (\mathbf{{a}}_{s})$ is the contribution from semantic memory; it acts as a prior for the unary statements.
The visual input contributes the term $\mathbf{{a}}_{c}^\top \textrm{sig} (\mathbf{f}(\textit{BB}_{\textit{sub}}))$, which is the inner product of the embedding vector with high-level features generated from visual inputs. 
Thus, whereas semantic memory is dependent on the embedding vectors and does not care about the meaning of a dimension, this meaning is provided by perception--- the inner product with high-level features. We call this the grounding of embedding vectors. 
 
The embedding vector is a prototypical vector for that concept in a high-dimensional representation space; the assumption is that, in this space, distances are meaningful, and the complex mapping performed by the DCNN performs a normalization, such that, for example, all kinds of dogs with different shapes and appearances form a connected subspace in the semantic embedding space.
In cognitive neuroscience, the semantic embedding space is sometimes referred to as conceptual space (\cite{gardenfors2016geometry} p. 21ff), 
where points denote objects, and regions denote concepts.

Figure ~\ref{fig:tsne_visualization} shows an analysis of the entity embeddings, based on the t-SNE visualization \citep{van2008visualizing}. The plot clearly shows a schema-like organization of concept embeddings in this two-dimensional projection, also known as a cognitive map \citep{tolman1948cognitive}.

\begin{table}[ht]
 \centering
 \resizebox{\textwidth}{!}{
 \begin{tabular}{c l l l}
 \hline
 $s^*$ (ID) & Unary attribute/ & Unary entity labels & Binary statements \\ 
 & Class labels & ($p$: \textit{sameAs}) & \textit{($s^*$, p, $o^*$)} \\ \hline
 10830[Person] & Person, 0.96 & 10830[Person] & (10830, wears, Shirt)\\
 &Mammal, 0.97 &&(10830, wears, 3495[Glasses])\\
 & LivingBeing, 0.98\\
 & Old, 0.98 \\
 & Other, 0.98 &\\
 & Walking, 0.96\\\hline
 Dog & Dog, 0.99 & 3537[Dog] & (Dog, on, Grass)\\
 & Mammal, 1.0 & 602[Dog] &(Dog, behind, Person) \\
 &LivingBeing, 1.0& 5976[Dog] \\
 & Young, 0.52 \\
 & Brown 0.35 \\
 & Other 0.99\\\hline
 Mammal & Dog, 0.38 & 3901[Cat] & (Mammal, on, Street)\\
 & Mammal 1.0& 9100[Horse] \\
 &LivingBeing, 1.0\\
 &Young 0.6 \\
 &Brown 0.31 \\
 & Other 0.96\\\hline
 Black & Person 0.24 & 9812[Bag] & (Black, on, Person)\\
 & Other 0.43 & 3634[Keyboard] & (Black, under, Sky)\\
 &NonLivingBeing 0.95\\
 & Old 0.59 \\
 &Black 0.99 \\
 & Other 0.99\\
 \hline
 \end{tabular}
}
 \caption{Semantic memory experience with generalized statements on VRD-EX data. The first column shows the queried $s^*$ (an entity, a class, or an attribute), the input to the algorithm. The second column shows highly rated attribute and class labels describing $s^*$. The third column shows highly-ranked (sampled) entities for the \textit{sameAs} predicate.
 The fourth column shows binary statements. 
 We see that person [10830] is a mammal and often wears shirts and glasses. However, we also see that the model ``explains'' what the class \textit{Dog} stands for, what the class \textit{Mammal} stands for, and what the attribute \textit{Black} stands for: We see that a dog is a mammal with 100\% probability, is brown with 35\% probability, and is often on the grass or behind a person. We see that if something is black, it is often a person (24\%), and black entities are often ``on persons'' and ``under the sky.'' See our discussion on generalized statements in Section~\ref{sec:gensta}.
}
 \label{tab:semantic_samples}
\end{table}

\begin{table}[ht]
 \centering
 \resizebox{1\textwidth}{!}{
 \begin{tabular}{c | c c | c c c c c c}
 \hline
 & \multicolumn{2}{c|}{Binary label} & \multicolumn{6}{c}{Unary labels: attributes/classes} \\
 Model & @10 & @1 & B-Class & P-Class & G-Class & Y/O & Color & Act. \\ \hline
 SM-givenClass & \textbf{92.67} & \textbf{49.74} \\
 RESCAL-givenClass & 89.95 & 26.06 \\ \hline
 SM-givenEntity & \textbf{100.0} & \textbf{90.42} & 100.0 & 100.0 & 100.0 & 100.0 & 100.0 & 100.0\\
 RESCAL-givenEntity & \textbf{100.0} & 90.12 \\\hline

 \end{tabular}}
 \caption{Semantic memory experience for models trained on the VRD-E data set. 
 In the \underline{first two rows}, we set the class labels of the subject and the object, e.g, \textit{Dog} or \textit{Person}, and predict binary labels. 
 Our results are better than a state-of-the-art model (RESCAL). 
 \citep{ruffinelli2019you} showed in a recent study that RESCAL is competitive with state-of-the-art approaches.
In the \underline{third and fourth rows}, we set the entity indices (e.g., \textit{Sparky} or \textit{Jack}). In this memorization, the performance is boosted by the entity representations. 
In these experiments, we are testing memory retrieval and not generalization, which explains the high scores. Best results in each block are in bold.
} 
 \label{tab:semantic_memory}
\end{table}

\begin{table}[ht]
 \centering
 \begin{tabular}{c | c c c c c c}
 \hline
 & Classification Accuracy \\
 Model & \textit{Dangerous}\\ \hline
 P-SA & 52.01\\
 SM & \textbf{100.0} \\
 P-enriched & 98.24 \\ 
 \hline

 \end{tabular}
 \caption{
 Semantic memory experience integrated with perception. The task is to predict the unary label \textit{Dangerous} with or without semantic memory. P-SA is the perceptual system, where the label \textit{Dangerous} was not provided in training. It can only be predicted by chance if an entity is dangerous.
 We trained the label \textit{Dangerous} as part of semantic memory. 
 In SM, the semantic memory experience is activated, which supplements the information from semantic memory if a visual entity is dangerous. P-enriched shows that, when the semantic memory is trained with the \textit{Dangerous} label, this information is also automatically integrated with perception, without an extra activation of a semantic memory experience.
 Enrichment works well with not-perceptual information, like social network background, which is, in a way, orthogonal to the visual scene input and is an indication that statements become dependent in training by the shared embeddings. The best result is in bold. 
}
 \label{tab:perception_semantic}
\end{table}

\subsection{Structure of the Embedding Vector}

All memories are implemented as embedding vectors forming the matrix $A$. 
In accordance with the standard neurocognitive view, the entries in $A$ correspond to synaptic coupling strengths~\citep{gazzaniga2004cognitive}.

 In our approach, we implemented symmetric connections between the units in the index layer and the representation layer. 
Thus, we have connection matrices $A$ and $A^{\top}$ in Figures~\ref{fig-basis-arch} and~\ref{fig-justsem}. 
 Although backward connections are common in the brain, they are typically not symmetrical. We did extensive experiments where we removed that constraint. The result was that the performance dropped by about $1\%$ in basically all experiments, so we stayed with symmetric connections in our work. 

 The representation layer is high-dimensional, although embeddings might be sparse, and a given index only activates a small number of components of the representation layer. 
 Naturally, the index that represents the color \textit{Red} is likely to mostly connect to components of the representation layer that are excited by red images. 
 Another advantage of sparse distributed representations \citep{rolls2016cerebral} is that this might lead to increased memory capacity \citep{ma2018holistic}.
Sparsity in the embedding vectors can be achieved in technical models, for example, by using appropriate regularization terms, like Lasso \citep{tibshirani1996regression}.
In our experiments, when we applied Lasso on all parameters, we obtained $70\%$ sparsity.

\subsection{Semantic Memory Experience: Semantic Decoding and Embodiment}
\label{sec:smr}

The engram of a concept is identical to its embedding. 
However, its treasure is only discovered by the semantic decoding. 
A complete semantic memory experience is not only the activation of a concept index and the activation of the representation layer with its embedding (i.e., its grounding) but also the subsequent decoding of the embedding into triples. For example, after $s = \textit{Sparky}$ activates $\mathbf{q} \leftarrow \mathbf{a}_{s}$, the agent can classify, for example, if $s$ is $c = \textit{Friendly}$, producing the triple statement \textit{(Sparky, hA, Friendly)}.
This exactly is going on in semantic memory decoding when we generate the sequential index pattern \textit{Sparky, Friendly}. 
The decoding produces triples from the  pre-observation model
which reflects the prior probabilities for triple statements.

In the dual subsymbolic view, the semantic decoding results in an embedding of the decoded sentences: Activating the 
index for Sparky, for example, embodies Sparky's embedding vector in $\mathbf{q}_S$; 
activating the index for \textit{Friendly} then leads to a superposition of both embeddings 
in $\mathbf{\ddot q}_S$
and represents Sparky but with an emphasis that Sparky is friendly. 

In a semantic memory recall, a second entity $o$ can be activated, that is, \textit{$o$ = Jack}, and then the binary label \textit{ownedBy} obtains high activation, producing the triple \textit{(Sparky, ownedBy, Jack)}. The subsymbolic representation of that statement is activated in $\mathbf{q}_P$. 
 
Both grounding and semantic decoding are local to the entity and might integrate information on that entity that was collected at different episodic instances and in different modalities. Table~\ref{tab:semantic_memory} shows that in our model, semantic memory can realize a very precise memory recall.

Table~\ref{tab:semantic_samples} illustrates the semantic memory experience, which is triggered by an entity, a class, or an attribute.
 The latter two correspond to a recall of generalized statements (see Section~\ref{sec:gensta}). 
 For example, it is shown that semantic memory can recall general information on \textit{Dogs} (that they are mammals with $100\%$), \textit{Mammals} (that they are dogs with $38\%$ and living beings with 100\%), and the color \textit{Black} (which, with 24\% probability, is a property of a person).

\subsection{Multimodality and Social Networks}
\label{sec:snets}

An agent can obtain data from different modalities.
This could concern subsymbolic sensory data besides vision
or symbolic information from conversations, books, movies, and other media. 
Multimodal data become part of episodic memory and semantic memory and enriches perception in different ways.
First, a multimodal episode is represented by an episodic index, and by activating that index, the data from that modality in that episode can be retrieved. 
Second, semantic memory serves as a site of multimodal integration simply because semantic memory is trained on triples from all modalities. So multimodal data will become part of semantic memory. 
Third, consider perception. We can now distinguish between visual labels ---labels, which were used in the training images like the color \textit{Black}--- and multimodal labels, which are the remaining labels, like \textit{Dangerous} or \textit{Rich}. 
Table~\ref{tab:perception_semantic} shows how multimodality directly enters in perception. We call this enriched perception, P-enriched. The unary label \textit{Dangerous} was not trained in perception but just in semantic memory. The table shows that information from the semantic memory is integrated into perception and episodic memory, even without activating an explicit semantic memory experience.
{Activating an entity index in perception or episodic memory subsequently activates the unary labels of that entity in context, but also recalls background of that entity, for example, relating to other modalities. Here it relies on semantic memory (see Section~\ref{sec:erg}).
}

To further examine multimodality, we derived a social network involving all persons in the data set by linking entities representing persons with the predicate \textit{knows}. Each person is linked with five other persons, whom we also denote as friends. 
 In addition, we define a social network's episodic instance as an instance at which the agent learns about the social contacts of one person.
 Our social network data set has 4987 person entities (along with their unary labels), 24953 \textit{knows} statements, and 4987 social episodes. We refer to this data set as VRD-S. More details on the generation of the social network data can be found in the Appendix~\ref{sec:appSN}. 
 
Table~\ref{tab:numerical_socialnetwork} shows numerical results on VRD-S. 
Given a person of interest, the semantic memory can recover friends (object $o^*$) and predict unary labels. 
Table~\ref{tab:clustering_snn} illustrates episodic and semantic memory, including social network data recall.

\begin{table}[ht]
 \centering
 \resizebox{\textwidth}{!}{
 \begin{tabular}{c | c c | c c | c c c c c c}
 \hline
 Model & \multicolumn{2}{c|}{$s^*$} & \multicolumn{2}{c|}{$o^*$} & \multicolumn{5}{c}{Unary labels of $s^*$}\\
 & @10 &@1 & @10 &@1 & B-Class & P-Class & G-Class & Y/O & Color & Activity\\\hline
 Episodic Memory & 100 & 51.47 & 99.97 & 65.80 & 100.0 & 100.0 &100.0 &100.0 &100.0 &100.0 \\
 Semantic Memory & - & -& 97.39 & 18.25 & 100.0& 100.0& 100.0 &100.0 &100.0 &100.0 \\
 RESCAL & - & - & 95.76 & 17.96 \\
 \hline
 \end{tabular}}
 \caption{Episodic and semantic memory experience for the social network data VRD-S. 
 The {first row} shows the performance of an episodic recall, with $t^*$ given. Columns 2 and 3 evaluate the prediction of a subject entity. 
 For columns 3 to 10, the subject $s^*$ is also given.
 The second row shows the semantic memory experience where the subject $s^*$ is given. 
 The unary labels have perfect recall. 
 Columns 3 and 4 show the performance of predicting a friend. Since each person has five friends, the recall@1 is upper-bounded by 20\%. 
 As a comparison, we show the performance of RESCAL in the third row, which shows slightly worse performance.
}
 \label{tab:numerical_socialnetwork}
\end{table}

\begin{table}[htp!]
 \centering
 \begin{tabular}{| c | c | c | c | c|}
 \multicolumn{5}{c}{\includegraphics[width=0.4\textwidth]{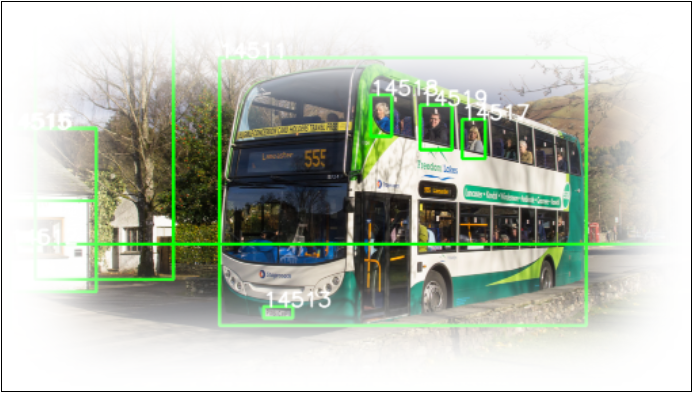}}\\
 \hline
 t* & s* & Unary labels & o* & Binary labels \\ \hline \hline
 \multirow{6}{*}{2177} & 14518 & Person, Mammal, LivB & 14511 & on \\
 &[Person] & Young, Other, Sitting & [Bus] & \\
 \cline{2-5}
 & 14515 & Building, Other, NonLivB & 14516& has \\
 &[Building] & Old, White, Other & [Roof] & \\
 \cline{2-5}
 & 14511 & Bus, Vehicle, NonLivB& 14512 & 
 nextTo (-) \\
 & [Bus] & Old, Other, Other & [Road] & \\
 \hline \hline
 \multirow{6}{*}{6662} & 14518 & Person, Mammal, LivB & 10669 & knows \\
 & & Young, Other, Sitting & &\\
 \cline{2-5}
 & 18010 & Person, Mammal, LivB & 14518 & knows\\
 & & Young, Other, Other & & \\
 \cline{2-5}
 & 8318 & Person, Mammal, LivB & 14518 & knows\\
 & & Old, Other, Other & & \\
 \hline \hline
 \multirow{3}{*} & \multirow{3}{*}{14518} & Person, Mammal, LivB & 10669 & knows\\
 \cline{4-5}
 & & Young, Other, Sitting & 25066 & knows \\
 \cline{4-5}
 & & & 12825 (-) & {knows} (-) \\
 \hline
 \end{tabular}
 \caption{Episodic and semantic memory experience that includes multimodal data (from a social network). The figure shows the original visual scene, which is unavailable during episodic recall. 
 The scene index of the image is $t^* = 2177$. The {top segment} shows sampled subject entities ($s^*$), highly ranked unary labels ($c^*$), a sampled object $o^*$, and the top predicted binary label ($p^*$). All labels are correct except for one binary label (indicated by (-)). 
 The {second segment} shows an episodic recall of a \textit{social episode} with index ($t^* = 6662$). At that episodic instance, social network information (binary labels \textit{knows}) was provided, which is recovered in the episodic recall. Shown are three correct triples that were recovered in the sampling of the episodic memory. 
 The {bottom segment} shows \textit{knows} statements recovered from semantic memory for $s^* = 14518$ (thus without a recall of a special episodic memory). 
 Two binary statements are correct, and one is incorrect (indicated by (-)). 
}
 \label{tab:clustering_snn} 
\end{table}

 \section{Reasoning and Language}
 \label{sec:reas}

 \subsection{Embedded Reasoning and Generalization}
 \label{sec:erg}

When the states of new triples are observed, this will be reflected in changes in the 
${\underline {\underline T}}$ tensor, which affects
$\mathbb{P} (A | {\underline {\underline T}})$ and the embedding matrix $A$ (Equation~\ref{Eq:delta}). Thus, although triple statements are independent, given all entries in $A$, entries in $A$ change with new data. 
For example, $\mathbf{a}_{s}$ changes when new truth values of statements involving $s$ become available. 
 Thus, if it becomes known that Sparky is a ``Dog'', 
 $\mathbf{a}_{\textit{Sparky}}$ will be adapted and the unary statement for the label ``Mammal'' will exhibit a high probability. 

 Episodic memory, that is, the observation model, generates samples for observed entities in the modality of the instance. Although episodic memory purely models the observations, there is also some degree of generalization by exploiting correlations in observations. Semantic memory, that is, the pre-observation model, can generate samples for all entities and modalities and fills in the unobserved. Semantic memory can show a significant degree of generalization. 
If a particular triple is often sampled in episodic memory but not in semantic memory, this would indicate an anomaly. As an example, consider that Sparky is a friendly dog (semantic memory), but in a particular episode, he was agitated and barking (episodic memory), then that is remarkable. 

Perception is based on episodic attention (EA). After perceptual decoding, an instance embedding $\mathbf{a}_{t}$ is realized, which is fitted to model the extracted triples. 
 Subsequently, $\mathbf{a}_{t}$ generalizes perception and becomes the basis for episodic memory. See also the discussion in Section~\ref{sec:coglearn}.

The mind can distinguish between what is currently being observed, what the semantic memory fills in (by activating $\bar t$ and $\mathbf{\bar a}$), and what an episodic memory could contribute (by activating a past $t$ and $\mathbf{a}_t$).

\subsection{Symbolic Reasoning}
\label{sec:symreas}

In contrast to embedded reasoning, symbolic reasoning solely depends on indices and their relationships, without involving the representation layer.
It would be completely amodal, in contrast to the modal embedded reasoning. 
In particular, it does not involve the adaptation of the embedding matrix $A$. 
Examples would be the definitional structure of a bachelor (a male person who is not and has never been married) and reasoning with subclass relationships (any dog is also a mammal). Definitional examples, like the definition of a bachelor, enter the mind of the agent via language and are crisp. They describe how the world should be. 
{Symbolic reasoning can also link singular events with state changes. For example, one might specify that if someone gets married (a singular event), the state of this person changes from single to married.} Another example is the triangle rule in social networks: if person~A and person~B are known to be friends and person~A is known to be a friend of person~C, then person~B is also likely a friend of person~ C. This example shows that symbolic reasoning is not necessarily deterministic. Descriptional dependencies, like the triangle rule, are learned by means of observations, are often probabilistic, and 
 describe how the world actually is. \citep{halford2014categorizing} has explored how analogical reasoning can be performed in the data tensor, loaded in working memory. The work assigns symbolic reasoning to System-2. One might call it ``inference by contemplation.'' \citep{nickelreducing2014} is an example where both embodied and symbolic reasoning are combined in an additive model.

\subsection{Embedded Symbolic Reasoning}
\label{sec:embsymreas}

 Triple sentences used in symbolic reasoning have an embedding and lead to another form of embedded reasoning. But in contrast to standard embedded reasoning, the embedding matrix $A$ is not modified in embedded symbolic reasoning. A simple symbolic rule would be that a dog is a mammal, that is, $\forall s: \textit{(s, hA, Mammal)} \leftarrow \textit{(s, hA, Dog)}$. The corresponding triple statement is $Y_{\textit{Dog, hA, Mammal}}$ is embedded and $\mathbb{E}(Y_{\textit{Dog, hA, Mammal}})$ can be learned from data. Section~\ref{sec:chaining} describes how the BTN's embedding approach can produce the sequential index pattern \textit{Sparky, Dog, Mammal}, which is interpreted as 
\textit{Sparky} being a \textit{Dog} and a
\textit{Mammal}. We call this embedded symbolic reasoning. It builds on the same BTN architecture as perception and memory. 
Embedded symbolic reasoning is integrated in sampling.
Assuming that, based on scene input, the unary label \textit{Mammal} is active with $60\%$ for a given visual entity, so the agent is uncertain about this label. Then, if sampling ``decodes'' that the visual entity is a \textit{Dog}, embedded symbolic reasoning lifts the probability for the label \textit{Mammal} to $100\%$ (due to the learned rule that every dog is a mammal) and the corresponding unary label is sampled as well.
Embedded symbolic reasoning can capture complex dependencies and thus can go beyond embedded reasoning. Embedded symbolic reasoning can be executed fast since it does not depend on a tuning of embeddings for entities or instances.

Table~\ref{tab:semantic_samples} shows generalized statements, which are the basis for embedded symbolic reasoning. We see that \textit{(Dog, hA, Mammal)} is true with $100\%$ (every dog is a mammal) and \textit{(Mammal, hA, Dog)} is true with $38\%$ (not every mammal is a dog). Also, 
Table~\ref{tab:complex_perception} shows results for generalized binary statements.

\subsection{A Foundation for Consciousness?}

The dynamic context layer might play a role in working memory. 
In general, working memory is associated with decision-making and cognitive control \citep{baddeley1992working} and is necessary for keeping task-relevant information active, as well as manipulating that information to accomplish behavioral tasks.
A modern view is that working memory is distributed across the cortex \citep{buschman202031}.
There is an emerging consensus that most working memory tasks recruit a network of the prefrontal cortex (PFC) (front-of-the-brain hypothesis) and/or parietal areas in the prefrontal parietal network (PPN) (back-of-the-brain hypothesis)~\citep{seth2022theories}. 

PPN activity is consistently reported in both attention and consciousness studies \citep{bor2012consciousness}. Their publication proposes that the PPN can be viewed as a ``core correlate'' of consciousness. \citep{dehaene2014consciousness} defines consciousness as ``global information sharing'' where information has entered into a specific storage area that makes it available to the rest of the brain. 
\cite{koch2016neural} argue that the posterior hot zone (PHZ) is the minimal neural substrate essential for conscious perception.
The PHZ includes cortical sensory areas in the parietal, temporal, and occipital lobes. 
 
The interactions between the index layer, the representation layer, and the dynamic context layer (which might involve working memory), might be a foundation from which evolution eventually generated human consciousness. The representation layer, associated with the parietal area of the brain, plays a major role in our approach, as well as its interaction with the dynamic context layer, which might include the working memory. The activation of the representation layer encodes the cognitive brain state. It integrates information and then makes this information available to the brain as a whole. In particular, the activation of the representation layer represents a holistic, integrative view of the agent's cognitive state at that instance. See also our discussion in the remaining part of this section. The role of the memory system in consciousness is discussed in~\citep{budson2022consciousness}.

\subsection{Serialization of Parallel Computing and the Central Bottleneck}
\label{sec:cbo}

Here are some examples of compositionality and serial processing in the BTN, and possibly in the mind.
First, perception analyses a scene that evolves in time, which is a serial process. Second, an episode is composed of several events. Third, an event consists of several bounding boxes, which are analysed serially. 
Fourth, many triples are generated serially to decode a scene semantically. Finally, a triple itself is represented as a sequence of indices (see Table~\ref{tab:firing}).

On the other hand, any computation involving the representation layer ---in particular, the mappings involving the DCNN--- and the mappings between the index layer and the representation layer are highly parallel.

It is this mixture of parallel and sequential processing exhibited in our model that might also make up the brain's operation.
Consider, that in a first step, we map $\textit{scene}_{t'}$ to 
$\mathbf{f}(\textit{scene}_{t'})$, and this representation can be analysed rapidly and might trigger action with a limited understanding of scene content. 
Only then is bounding box content analysed in detail, which might be supported by saccadic eye movements, and this is a slower serial process, permitting a transition to language, 
as discussed later in this section.
Maybe not surprisingly, language makes thoughts explicit but at the same time might slow down thinking.

Sequential processing is also a core concept in the theory of a global workspace \citep{baars1997theater, bor2012consciousness, dehaene2014consciousness,goyal2021coordination}:
\citep{koch2014keep} discusses that Dehaene's workspace has extremely limited capacity (``the central bottleneck'') and that the mind can be conscious of and pay attention to only one or a few items or events, although these might be quickly varying.
In cognition neuroscience, the general understanding is that parallel multitasking of cognitive tasks likely is an illusion. Even working memory is assumed to be able to store only three to four items at a time \citep{awh2020online}. 
Sequential processing would also contribute to a potential solution of the binding problem \citep{singer2001consciousness} since the decoding focuses on concepts in a serial fashion, and associations between activities in the representation layer and the index layer are well defined. The importance of sampling is also recognized in \citep{dehaene2014consciousness} in the context of conscious perception.
For example, the author states that ``consciousness is a slow sampler.''

Another interesting point is that both \citep{dehaene2014consciousness} and \citep{koch2016neural} assume mental states, well delineated from all the other states. Switching between different interpretations is also a property of our sampling approach if one interprets a sample as a temporary decision or an interpretation:
”It's a bird, or a plane, or it's Superman, but not all of them at the same time” \citep{dehaene2014consciousness}.
Dehaene discusses a similar process of ``collapsing all unconscious probabilities into a single conscious sample.'' 
{His model assumes a ``winning neural coalition''; in our approach, the embedding of an episodic instance integrates all information available at that instance, and the winning neural coalition would be the samples and triple sentences associated with that embedding. The optimization of the embedding of an episode, that is, $\mathbf{a}_t$, is a step that occurs after the decoding of a scene, and thus the conscious experience associated with it might be slightly delayed (see also Section~\ref{sec:fslearn}). A current hypothesis in cognitive neuroscience is that conscious awareness often follows after an observation or a decision is made and serves to explain and justify but not to trigger an action ~\citep{gazzaniga2004cognitive,budson2022consciousness}.
Only some important decisions are actually made with the help of conscious slow thinking. 
}
In the semantic decoding of our model, the representation layer is periodically activated, which might be reflected in neural signals and could be related to some of the neural oscillations found in the brain.
A candidate is the beta rhythm (13-35 Hz), considered to be related to consciousness, perception, and motor behavior. Also of relevance might be the gamma wave (25-140 Hz), which is correlated with large-scale brain network activity and cognitive phenomena such as working memory, attention, and perceptual grouping.

\subsection{One-brain Hypothesis}
\label{sec:obhyp}

We discuss the one-brain hypothesis in the context of two aspects. 
First, the representation layer can contain a superposition of embedding vectors and activation vectors, but it cannot represent a concatenation of two vectors. This would, in a way, require two separate representation layers. 
We propose that this is biologically not plausible, and we adhere to Dehaene's concept of a central bottleneck. 
Applied to perception, this means that the mind can only process one bounding box content at a time. 

Second, perception, episodic memory, semantic memory, and even embedded symbolic reasoning are realized by the different operational modes of a \textit{single architecture}; thus, we do not assume different modules for these different functions. Instead, the brain repurposes the same architecture for different functions.

Of course, the brain does many things concurrently, and many of these activities are hidden from conscious awareness. This does not contradict our one-brain hypothesis, which is only concerned with the proposed operations of perception and memory, and in particular, when they involve the representation layer, that is, the cognitive state of the brain.

\subsection{Cognitive Linguistics}
\label{sec:lang}

Humans differ from other animals in their ability to express themselves through natural language. Human language is the basis for communication but also a means to argue and reason. 
Thus, an agent can tell another agent not to leave the hideout since a bear is lurking outside, even when the bear is not visible.
Perception, episodic memory, and semantic memory are all declarative (i.e., explicit), and humans can verbally report on either.
We propose that the generated triples in our approach are a basis from which the rich human language might have evolved.

Leading approaches in cognition and linguistics are, first, the formal approach, second, the connectionist approach, 
and third, the embodied approach \citep{evans2012cognitive}. Our work relates to all three. 

Let's consider first the formal approach. The \textit{language of thought hypothesis} assumes that mental representation has a linguistic structure, as well: thoughts are sentences in the mind. 
\cite{fodor1975language} describes the nature of thought as possessing ``language-like'' or compositional structure (sometimes referred to as mentalese). 
In this view, simple concepts combine systematically (akin to the rules of grammar in language) to build thoughts. Also, in our approach,
the brain talks to itself by producing triple sentences and their embeddings. 
We also agree that language offers a window into the operation of the brain. However, in our approach, we do not follow Fodor's formal logic-based view: Our brain is more of a chatterbox. 

Second, we follow a \textit{connectionist approach}~\citep{mcclelland2020placing}, since the BTN includes a DCNN and neural processing. 
However, our notions of discrete symbolic indices, embodiment, and compositionality go beyond a more standard connectionist approach. 

Finally, we can relate to the concept of an embodied language.
There is general agreement that in a form of bottom-up processing, perception, the state of mind, and the body influence language.
The idea of an embodied language is that in a form of top-down processing, language can influence the perceptual path and the body \citep{evans2012cognitive}. As discussed, we see our approach as being embodied, at least as far it concerns earlier processing in the brain. 
One might even consider that language takes over the mind, particularly in memory, with no immediate visual input to be decoded. From a brain, talking to itself, of course, there is a small step to a brain that speaks to others. Most generated triple sentences are not transferred into language; what is actually spoken is obviously more complex, nuanced, and sophisticated, and is modulated, for example, by intent, social context, and cultural background. Thus an internal triple-oriented fast speech is transferred into an external, sophisticated slow speech.
When individuals learn a new language, this might involve the mappings between fast speech and slow speech, in addition to the direct mapping from one language to another language. The inner fast speech might already exist in some animals.

{If the agent is the recipient of language, 
say, in a conversation, by reading, or by consuming media, the acquired information could be stored as an episodic memory instance, that is, as an episodic index with its associated embedding vector. This knowledge then can become part of semantic memory. In our simplistic triple-oriented conversation, the agent can learn about facts, for example, \textit{(Sparky, hA, Dog)} and generalized statements, such as Hearst patterns \citep{hearst1992automatic} like \textit{(Dog, hA, Mammal)}.
}

Language compresses information. 
Significantly different scenes might generate very similar descriptions, demonstrating great invariance in triple descriptions and language in general.

\section{Episodic Memory}
\label{sec:cogmem-em}

\subsection{Background on Episodic Memory}

Episodic memory documents the life of an agent. 
\cite{tulving1985elements} describes episodic memory as a memory that, in contrast to semantic memory, requires a recollection of a prior experience. It is considered to be the result of rapid associative learning in that a single episode and its context become associated and bound together and can be retrieved from memory after a single episode.
Episodic memory stores information of general and personal events \citep{tulving1972episodic,tulving1985elements,tulving2002episodic,gazzaniga2004cognitive} and concerns information we ``remember,'' including the spatiotemporal context of events \citep{gluck2013learning}.

Some theories emphasize the sequential nature of episodic memory and the memory process.
\cite{moscovitch2016episodic} considers an episodic memory experience to be an active process that involves details of the event and its location. Sometimes the reconstruction is regarded as a Bayesian process of reconstructing the past as accurately as possible based on available engram information \citep{hemmer2009bayesian}.

\begin{figure}[htp!]
\centering
 \includegraphics[width=\textwidth, height = \textwidth]{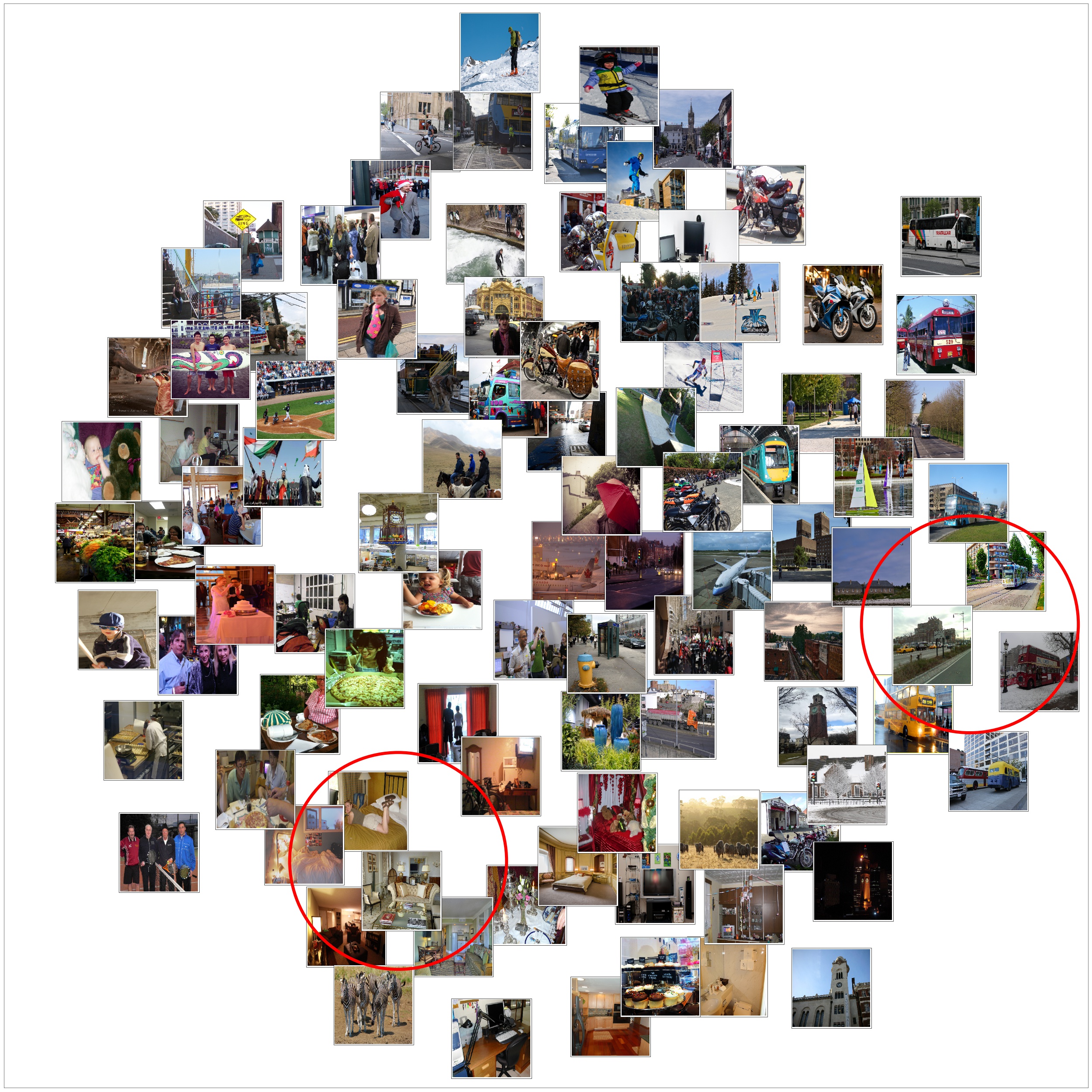}
 \caption{t-SNE visualization of episodic instances based on their embeddings $\{\mathbf{a}_t\}_{t=1}^{N_T}$. One can see that similar scenes are often in proximity. For example, the circled areas show images of indoor scenes (bottom) and of buses on streets (right).
} 
 \label{fig:clustering_time}
\end{figure}

\subsection{Event Memory and Episodic Memory}

In this article, we use the terms event memory and episodic memory almost synonymously. 
In the narrower sense, we consider event memory to be related to a single instance in time and episodic memory to a sequence of events that is, a story.
This is in agreement with~\cite{mannila1997discovery}, who define an episode as a collection of events that occur relatively close to each other in a given partial order.
We propose that the distinction between event memory and episodic memory is blurred. 
First, we often remember rather static images of past episodes.
Second, even an analysis of a static scene is a sequential process, where each bounding box is recovered in sequence. 
Third, during a single saccade, the scene might have already changed: 
Consider, for example, an agent riding a bicycle where the scenery is constantly changing. 
So the recovery of a single scene might already describe a dynamic event.

\subsection{Recall of Episodic Memory Engrams}

An activated past episodic instance $t$ restores its embedding $\mathbf{a}_{t}$, that is, its engram, in the representation layer. 
Figure~\ref{fig:clustering_time} shows that the embedding vectors of episodic memories form meaningful maps and also are organized as a conceptual space.

\begin{table}[ht]
 \centering
 \resizebox{1\textwidth}{!}{
 \begin{tabular}{c | c c | c c c c c c |c c}
 \hline
 & \multicolumn{2}{c}{$s^*$} & \multicolumn{6}{|c|}{Unary labels} & \multicolumn{2}{c}{Binary labels} \\
 Model & @10 & @1 & B-Class & P-Class & G-Class & Y/O & Color & Act. & @10 & @1 \\ \hline
 EM & 99.84 & 37.13 & 100.0& 100.0& 100.0 &100.0 &100.0 &100.0 & 100.0 & 90.40\\
 P-noI & 0.0 & 0.0 & 41.19 & 20.83 & 51.76 & 49.03 & 12.97 & 71.05 &39.34 &7.79\\
 \hline

 \end{tabular}}
 \caption{Top row: episodic memory experience using {VRD-E} data. For randomly selected past episodic instances $t^*$ as input, we determine the highest-ranked entity indices (first two columns). 
 For the other columns, we set the correct entity indices ($s^*$ and $o^*$) and predict unary labels and binary labels. The performance is, in many cases, 100\%. Thus, an agent might recall seeing a black dog, but it might be less confident that it was \textit{Sparky}. Considering the large number of entities in the data set, the performance on entity prediction (first two columns) is still impressive as well.
Bottom row: In P-noI we removed all entity indices, using only class and unary labels.
The bad performance demonstrates the relevance of representing entities for memory recall. 
}
 \label{tab:episodic_numeric}
\end{table}

\begin{figure}[htp!]
\centering
 \includegraphics[trim= 0 5cm 0 3.5cm, clip, width=\textwidth]{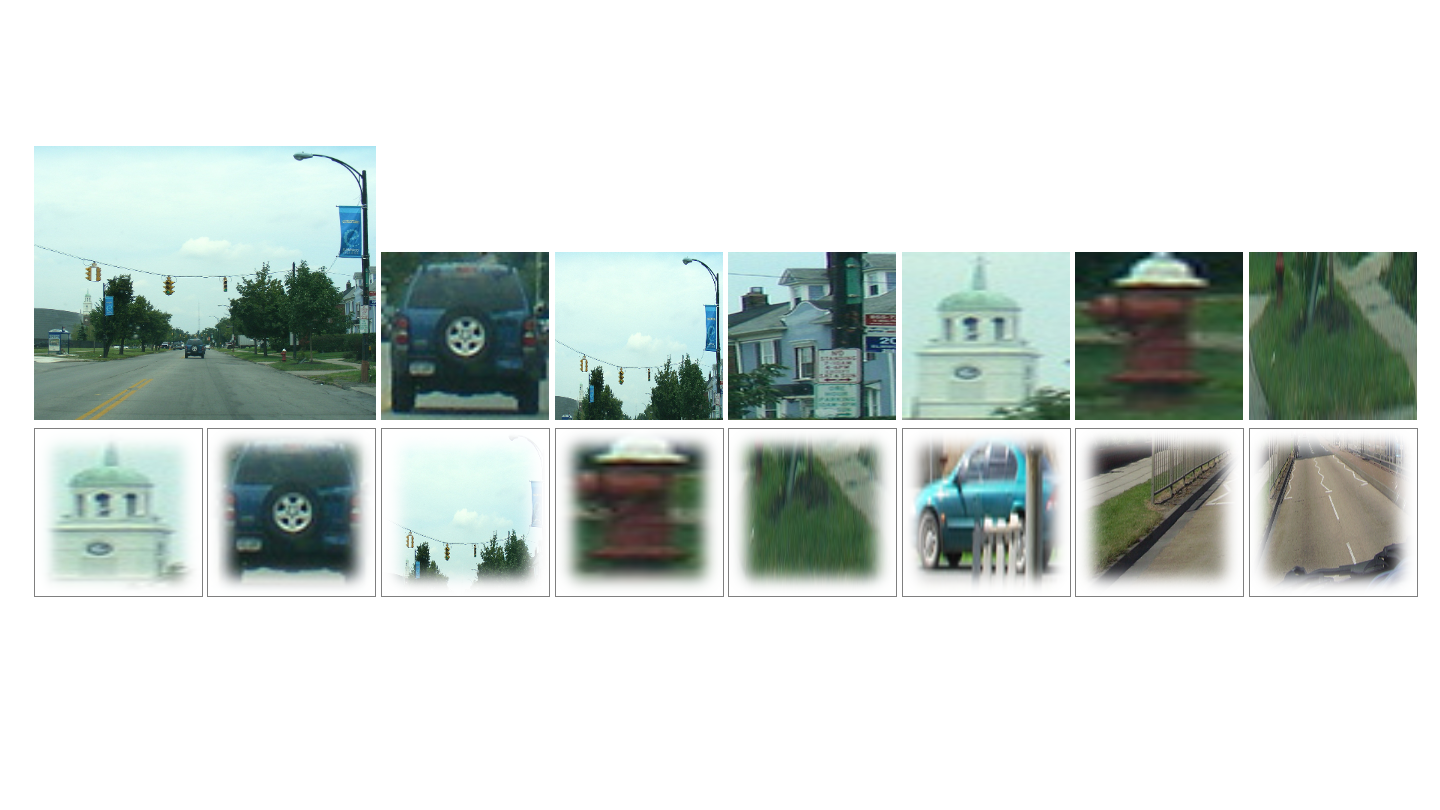}
 \caption{Episodic memory experience using {VRD-E} data.
 The first row shows the visual input to perception and then extracted bounding boxes for entities in the scene. 
 In the second row, we show an episodic recall of this scene ($t^*=t'$), which is now a memory. We show the bounding boxes of the scene entities, which were recalled. 
 We see that, in sampling, episodic recall recovers entities with correct bounding box content (e.g., fire hydrant, sky, car). Columns~6 to 8 show incorrect recalls. 
 Note that these incorrect recalls would still be entirely plausible in the scene's context. 
}
 \label{fig:episodic_recall}
\end{figure}

{
\subsection{Episodic Memory Experience: Semantic Decoding and Embodiment}
}
\label{sec:emr}

The embedding of an episodic index is all there is, but the treasures of an episodic memory are only deciphered by the semantic decoding of that embedding, that is, the symbolic reconstruction of triples involving that episode (see Table~\ref{tab:firing}).
The recall of the episodic embedding, but especially the semantic decoding into triples, has the character of a simulation: 
Episodic memory is a reactivation of a possibly multimodal memory experience \citep{evans2012cognitive} and has been described as a reliving of past experience \citep{shapiro2010embodied}.
Figure~\ref{fig:episodic_recall} illustrates an episodic memory experience. 
Table~\ref{tab:episodic_numeric} provides numerical results.

\subsection{Episodic and Semantic Memory in Perception}

We propose that in perception, an agent first uses both the episodic and semantic attention approximation, which can be executed fast and in parallel. 
Only in a second step are specific entities sampled. This permits the integration of specific multimodal background (see Section~\ref{sec:snets}), and permits associations with past episodic instances and known entities.
Thus if perception poses the hypothesis that an entity in the image is identical to Sparky, then the semantic memory experience would supplement specific background information on Sparky, which cannot be derived from the visual input. 
Semantic memory reconstructs what is known about the concept (i.e., the prior information).
Similarly, episodic memory would recall relevant past episodes.

\subsection{Recent Episodic Memories for Context}

\begin{figure}[htp!]
\centering
 \includegraphics[trim= 0 5cm 0 5cm, clip, width=\textwidth]{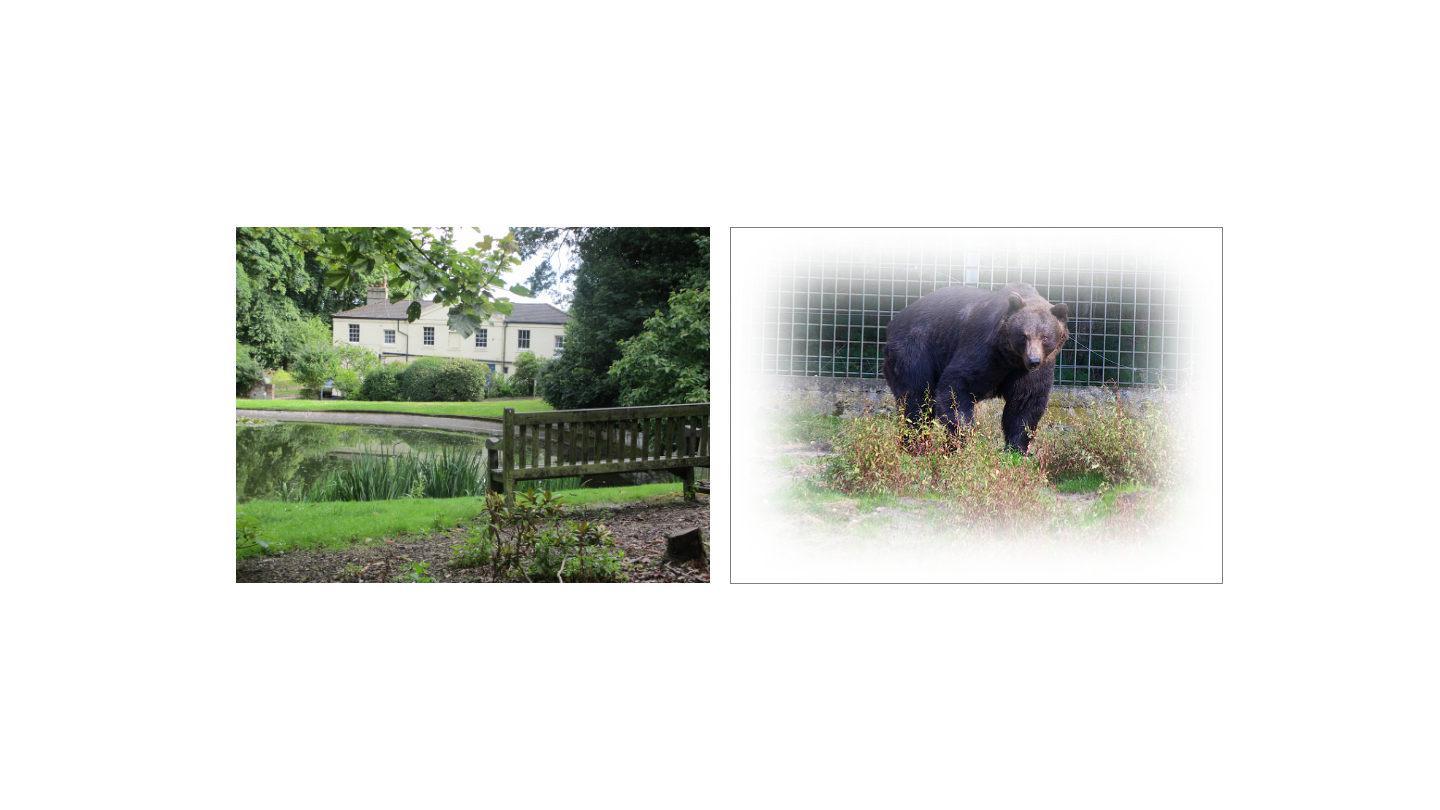}
 \caption{{Recent episodic memory experience: An illustration of the effect of a recent episodic memory experience using VRD-E data.
 The left image shows a harmless garden scene, but due to a recall of a recent episodic memory $t^*=684$, the agent is aware of the lurking bear close by (right scene). Labels for visual entity recovered in the episodic recall (right scene) are \textit{Bear, Mammal, LivingBeing, Old, Black, OtherActivity, Dangerous}. Note that episodic recall is not triggered by closeness in a scene but by recency and relevance.}
}
 \label{fig:recent_episodic}
\end{figure}

\begin{figure}[htp!]
\centering
 \includegraphics[trim= 0 5cm 0 5.5cm, clip, width=\textwidth]{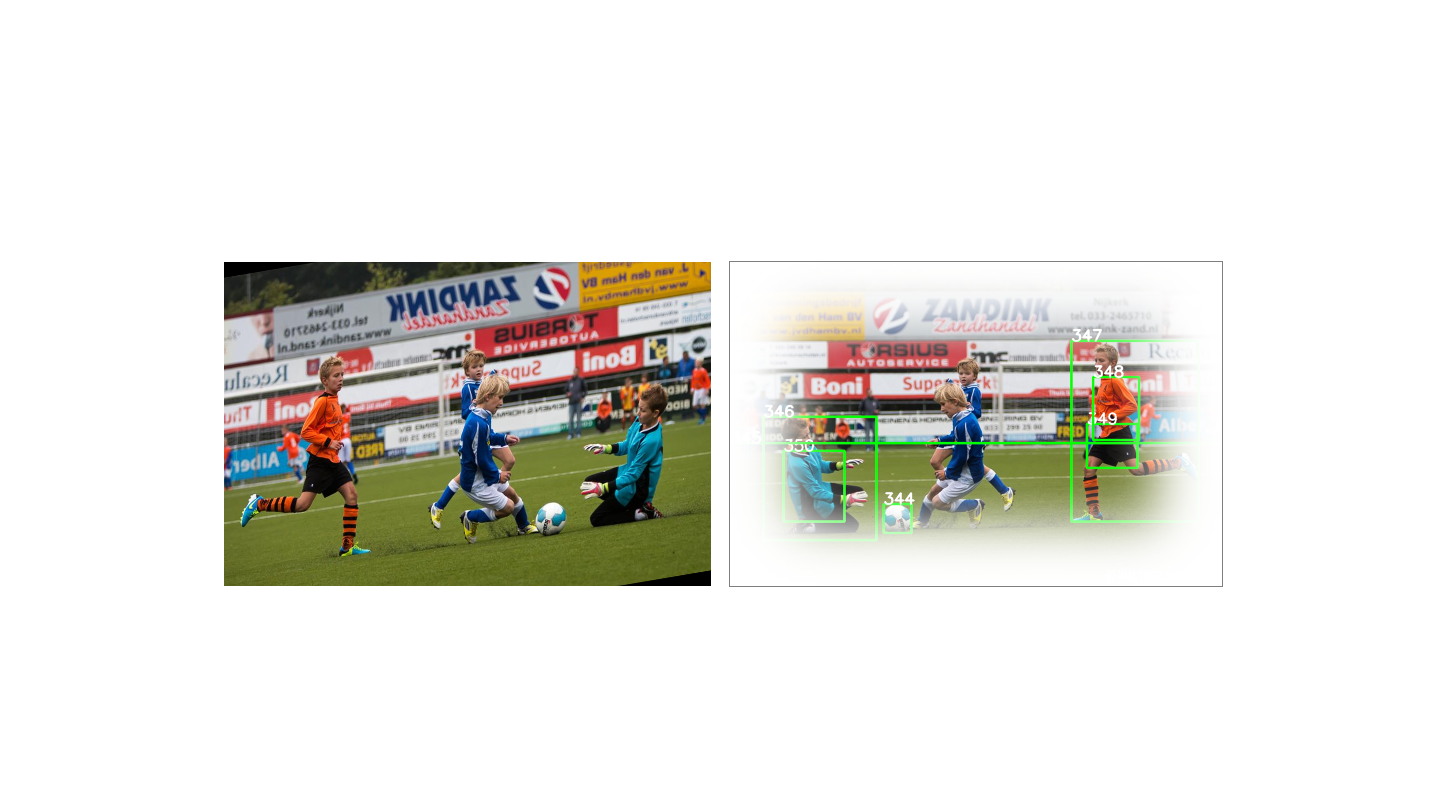}
 \begin{tabular}{|c | c | c | c|}
 \hline
 s* & Unary labels & o* & Binary label \\ \hline
 346 &
 Person, 1.00, Mammal, 1.00, LivingBeing, 1.00 & 345 & on \\
 & Young, 1.00, Other, 1.00, Other 1.00 & & \\
 \hline
 348 &
 Shirt, 1.00, Clothing, 1.00, NonLivingBeing, 1.00 & 347 & on \\
 & Young, 1.00, Orange, 1.00, Other, 1.00 & & \\
 \hline
 347 &
 Person, 1.00, Mammal, 1.00, LivingBeing, 1.00 & 348 & wear \\
 & Young, 1.00, Other, 1.00, Playing, 1.00 & & \\
 \hline
 345 &
 Grass, 1.00, Plant, 1.00 LivingBeing, 1.00 & 346 & under \\
 & Old, 1.00, Green, 1.00, Other, 1.00 & & \\
 \hline
 \end{tabular}
 \caption{Recent episodic memory experience. 
 (Top) The left image shows visual input to perception 
from VRD-EX data. Then a recent $t^*$ is sampled.
 The right image shows the image belonging to $t^*$. (Bottom) The table shows results from this episodic memory experience without the actual image for $t^*$
 being available, only based on memory recall. 
 The first column shows samples for $s^*$.
 The second column shows unary labels.
 The third column shows sampled objects $o^*$.
 The fourth column shows the most likely binary label.
}
 \label{fig:similar_episodic}
\end{figure}

Recent episodic memory can provide the agent with information on recent perceptional experiences. A recall is triggered by the recency of the episodic instance and relevance, and it contributes to an agent's sense of the world state. A recent episodic memory experience is an episodic memory that is almost treated as a current observation. An agent needs to know about the state of the world, even for parts that are not currently being perceived.
 An example is the ``lurking-bear'' situation (see Figure~\ref{fig:recent_episodic}): ``There was a bear strolling around outside the hideout, $\ldots$. Remember: it might still be there, although it might not be visible from the hideout.'' 
Figure~\ref{fig:similar_episodic} shows another example of a recent episodic memory recall. 
This emphasis on episodic closeness can be implemented as ``time encoding'' \citep{ma2018embedding}, in a similar way as ``position encoding'' is used in the attention literature \citep{vaswani2017attention}.
Patients who are unable to form new episodic memory show great deficits in 
personal orientation and context understanding. 
These deficits are often associated 
with severe bilateral damage to MTL \citep{gluck2013learning}.

\subsection{Remote Episodic Memory for Decision Support}

Remote episodic memory can provide the agent with information on remote perceptional experiences similar to the current perceptual experience, and this can contribute to decision-making. 
An episodic recall is triggered by $t=t^*$ (a sample in Equation~\ref{eq:sstime})
which reflects the similarity between the current scene representation with the episodic embedding.
It makes sense that the current event should trigger the same action as in the retrieved episodic memory ---if it led to a good outcome--- or an alternative action if not.
 For example, if the agent finds the current situation very similar to a previous one, where it had walked toward a bear and almost got attacked, it would very likely not do this again!
Remote episodic memory provides information about possible future scenarios and aids the agent in decision-making.

Thus memory guides behavior. \cite{duncan2016memory} describe this process as integration across relational events by imaging possible rewards in the future. The value associated with a memory (e.g., reward, threat) might be an integral aspect of episodic memory. The article also states that there is now extensive empirical data supporting the prevalent use of episodic memory across various decision-making tasks in humans. 

\begin{figure}[htp!]
\centering
 \includegraphics[trim= 0 0cm 0 0cm, clip, width=\textwidth]{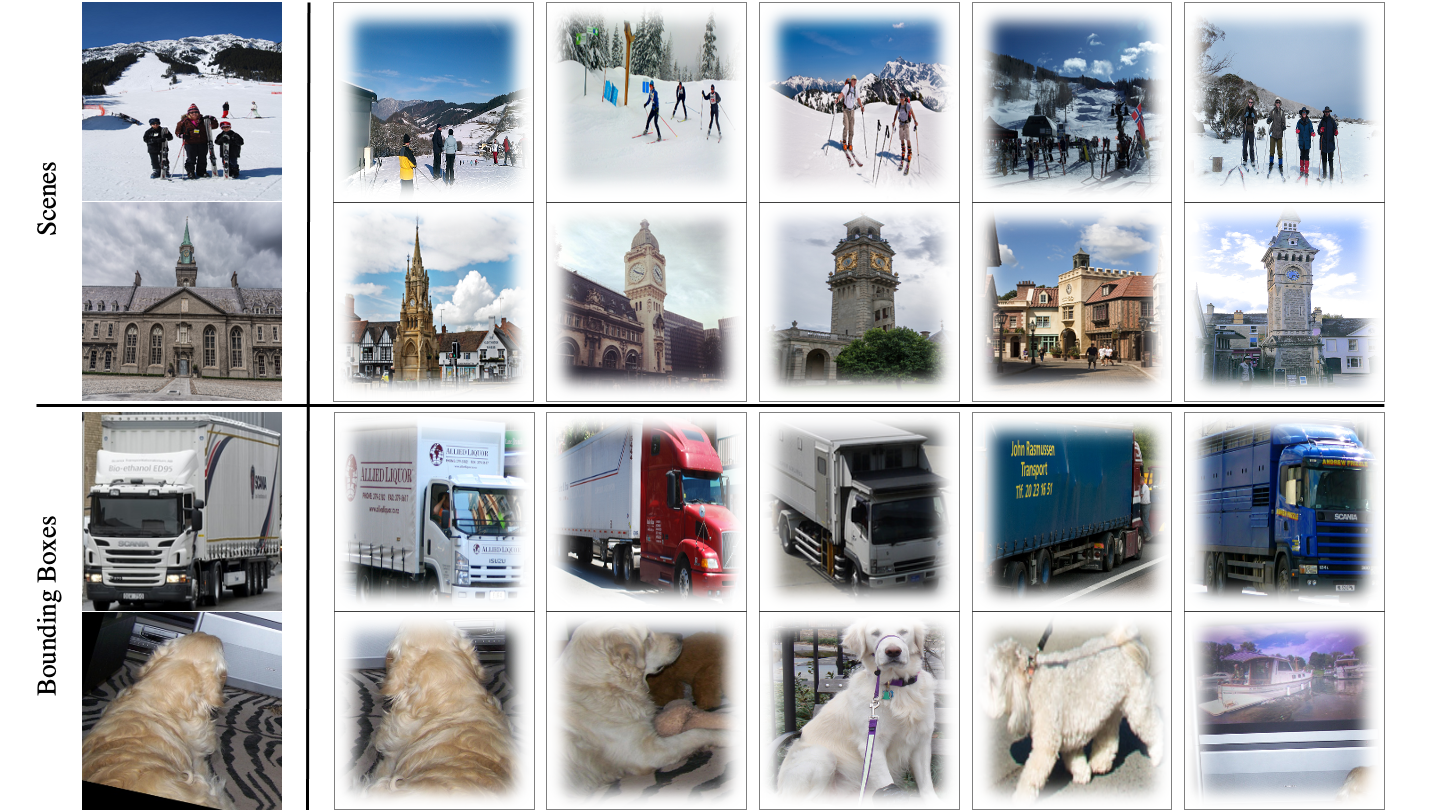}
 \caption{(Top) Remote episodic memory experience using VRD-E data. The two images in the first column show the visual input to perception. 
 Then we sample $t^*$, representing past episodic memories. The images of the scenes associated with the $t^*$ show that, indeed, recalled past episodic memories are related.
 (Bottom) The bounding boxes in the first column represent visual entities $s'$ in perception. The right bounding boxes display retrieved entities with high activation.
}
 \label{fig:time_samples}
\end{figure}

Figure~\ref{fig:time_samples} shows that a perceptual scene indeed can activate memories of similar scenes in remote episodic memory.

\subsection{Forecasting and Future Episodic Memories}
\label{sec:fem}

To forecast the state at the next instance $t'$, the agent might consider recent episodic memories to be current, that is, the instance index for those episodic memories is virtually changed to $t'$.
The post-observation model then combines these memories with current observations and  with the semantic memory to estimate the world state. 

A future episodic memory is a forecasted event in the future, which at some point in time is predicted to become a regular episodic memory. 
This helps plan future actions. 
Events might be predicted using, for example, temporal knowledge graph models \citep{han2020explainable}, assuming temporal smoothness and predictability of the embedding of time instances. 
Here we consider forecasts made entirely on a symbolic level, say, by verbal communication or reading. For example, f the agent has learned by verbal, symbolic communication that there is a football match in the stadium this evening and the weather will be bad, embedded reasoning, based on this assumed future information, predicts that there will be a traffic jam in the city and that driving will be difficult.
Technically, the BTN for future episodic memories is identical to regular episodic memories, only that the episodic index embedding is calculated based on assumed events in the future, either without or with imagined and simulated future sensory input.

\subsection{Memory Supports the Agent in the Present and the Future}
\label{sec:stem}

It is of great interest for an agent to estimate the state of the world at the current instance $t'$, and to predict how it will evolve.
We propose that the mind estimates the world's state using perception and memory: The memory systems provide information that makes the agent act right but is not communicated by current perceptual experience. 

\begin{figure}[t]
\vspace{0cm}
\begin{center}
\includegraphics[width=0.9\linewidth]{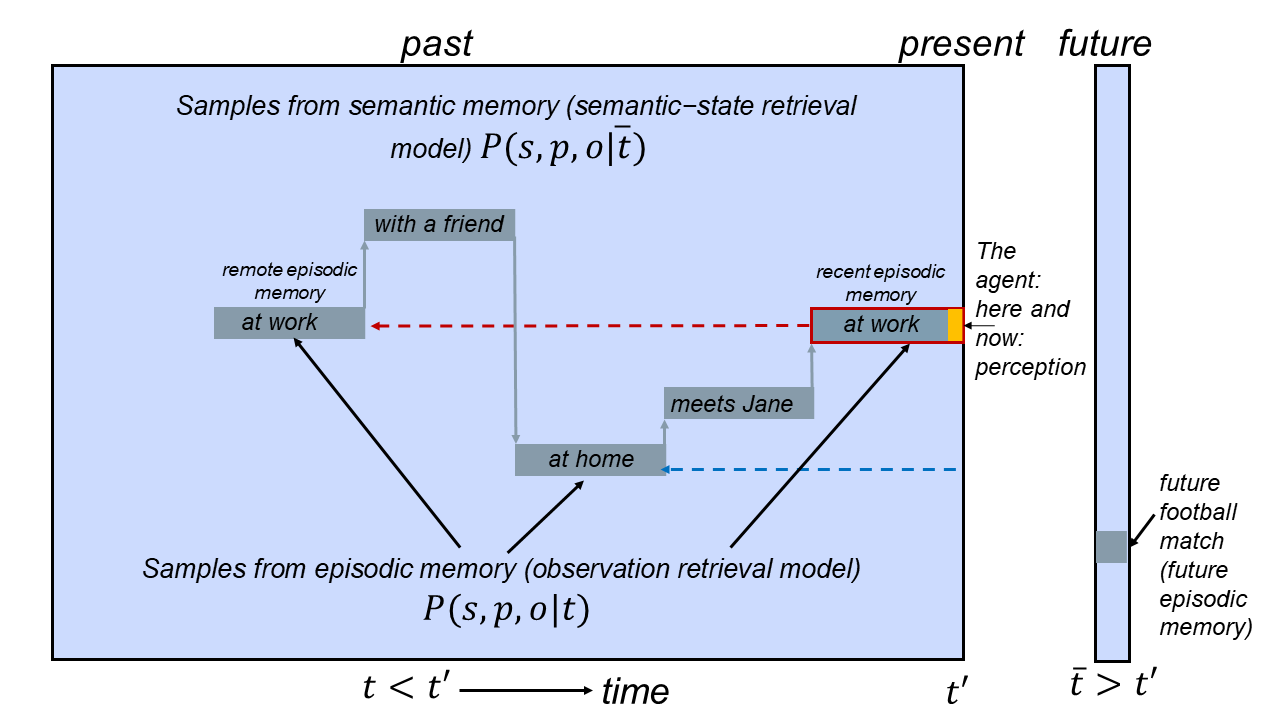}
\end{center}
\vspace{-0.5cm}
\caption{The horizontal axis stands for time and the vertical for some abstract triple-dimension.  We show past, present, and future. 
The current instance is $t'$ on the right. 
The light blue background box stands for the predictions of the pre-observation model of semantic memory. 
The  observation model of the episodic memory is represented by the horizontal gray bars; they indicate where, in the past, information was acquired. 
 At episodic instance $t'$, the agent learns about some statements by perception (orange), and these become part of the episodic memory (recent episodic memory).
 There might be an association with another portion of the episodic memory, that is, the remote episodic memory (red dotted arrow).
If the agent contemplates information unrelated to current perception, it can query on remote episodic memory (blue dotted lines). 
{The right vertical bar indicates a future episodic memory.}
}
\label{fig-allmemories}
\end{figure}

For example, to illustrate the relevance of memory in daily life, consider a typical day at the office of agent Mary (see Figure~\ref{fig-allmemories}). When Mary arrives at the office in the morning, she expects that everything is as usual, as modelled by \textit{semantic memory}. The pre-observation model (semantic memory) will produce statements describing the state of normality. This sets the stage. 

\textit{Perception} will produce triples describing the actual situation, for example, the status of the coffee machine, which is broken. 
Since the coffee machine has mostly been working, semantic memory by itself would have predicted (incorrectly) that the coffee machine most likely is working.
She immediately informs Jack in an office nearby: 
``Jack, can you believe this?! The coffee machine is broken, again!'' (\textit{from triples to language}). \textit{Remote Episodic memory} reminds her that this is not the first time that the coffee machine was broken. 
 \textit{Recent episodic memory} will remind Mary all day that the coffee machine
is broken, even when she is not in the same room as the coffee machine
and thus does not have immediate perceptual information on its status. As discussed, recent episodic memory is an episodic memory that is almost treated as a current observation. 
The state change of the coffee machine is only slowly integrated into semantic memory.  Being a long-term average, semantic memory is a sluggish state estimator and relies on recent episodic memory to emphasize recent state changes. 

Jack might ask Mary if the coffee machine had been working last Tuesday. If she does not have an episodic recall from last Tuesday about the status of the coffee machine, she might consult semantic memory,  reminding her that the coffee machine mostly has been working. But if she has an episodic memory that 
last Monday, the coffee machine was broken, she might infer that it most likely was also broken on Tuesday. Here, she consults a memory that is recent to the past time instance of interest.

Another \textit{recent episodic memory} might remind Mary that she had met Jane on her way to work and that she had said that she would drop by at the office (\textit{simulation of a recent episodic memory}). Jane is a good friend (\textit{semantic memory}).

Mary recognizes that Sparky is in the office (\textit{perception}), and \textit{semantic memory} adds triples describing background on Sparky, for example, that Sparky is Jack's dog. Mary might also recall that a while ago, Sparky was at the office and behaved well (\textit{remote episodic memory}), although dogs in general can be quite a nuisance 
{(as concluded by a \textit{generalized statement} and by employing embedded symbolic reasoning).}

When Mary relaxes, she suddenly recalls that people are repairing the heater today (\textit{episodic memory, not triggered by perception}) and that she should call her roommate to check if they arrived in time. 

Then Mary remembers that the football match will be in the stadium this evening and that the weather has been predicted to be bad (\textit{future episodic event}): there will likely be a traffic jam (\textit{generalization of future episodic events}), and, due to expected bad weather conditions, she should better drive carefully (\textit{an association stored in semantic memory).} 
Future-state prediction uses both episodic memory and generalized future episodic memory. 

 The story illustrates that background is provided by semantic memory.
Semantic memory is mostly relevant as it concerns triple statements, whose states are static or change rather slowly. 
Recent episodic memory provides information that is not yet or will not be absorbed into semantic memory  but contributes to the current state estimate.
 Episodic memory is mostly relevant in as much as it concerns state changes (broken coffee machine), singular events (meeting Jane on her way to work) or information, not yet known. 
As our story indicates, narratives often tell causal stories and might also reflect the events and facts the brain pays attention to for decision-making.
Future episodic memory supports planning. Remote episodic memory guides behavior: since the last time Sparky behaved well, he might behave well again this time. 

The whole story is on a symbolic level. But each index that fires is embedded and leads to embodiment. We would propose that the storyline involves System-1 type associative reasoning, supported possibly by some background knowledge of how the world works. As our story suggests: to deal with daily life, agents might need little effortful System-2 reasoning.

\section{To Perceive is to Learn}
\label{sec:coglearn}

\vspace{0.25cm}
{\raggedleft\textit{There is nothing in the mind that was not first in the senses.} ---John Locke}
\vspace{0.25cm}

\subsection{Overview}

In normal operations, a new perceptual event that becomes an episodic memory requires the establishment of a new episodic index and its associated embedding vector. 
This roughly corresponds to the MTL-based fast \textit{nonparametric learning system} in the complementary learning systems (CLS) theory \citep{mcclelland1995there,kumaran2016learning}.
A new perceptual event might contain novel entities not yet known to the agent. For important ones, the agent needs to establish new indices and their embedding vectors as well. More rarely, new indices for 
attributes, classes, and predicates need to be established. Being quite stable, those might be represented in neocortex. 

The second component of the CLS-theory is the \textit{parametric learning system}, where the neocortex is trained by replay in a slow process from data from the nonparametric learning system. 
Slow training serves as the basis for the gradual acquisition of structured knowledge about the environment and its transfer to neocortex \citep{kumaran2016learning}. An agent might get new information on existing entities, and this information might need to be integrated into their embeddings. In some applications involving online adaptation, interference of new knowledge with old ones can be a problem
and might lead to catastrophic forgetting \citep{kumaran2016learning}.
Catastrophic forgetting does not show up in our preliminary experiments. The large dimensionality and modularity of the embedding vector lead to robustness and stability; for example, learning first about Jack's size and hair color might not interfere with the learning of his social network later. And when he later dyes his hair, the changes in his embedding vector are local, that is, it affects only a few dimensions. 
Learning happens as an online process, so more recent episodes will have a greater effect than more distant ones.

\subsection{Neuroscience Perspective on Establishing Episodic Memory Engrams}

Episodic memories are first formed in the hippocampus, which is part of medial temporal lobe (MTL). 
The idea that episodic memory is index based is by now one of several accepted theories \citep{tonegawa2018role}.
It goes back to the hippocampal memory indexing theory \citep{teyler1986hippocampal,teyler2007hippocampal}, which was long controversial.
The event indices have a relational memory function in that they bind together different pieces of experience.
{Recent research found evidence for the existence of time cells in the hippocampus (CA1) \citep{eichenbaum2012towards,eichenbaum2014time,kitamura2015entorhinal,kitamura2015entorhinal2}.}

There is some evidence that neurogenesis might be involved in forming new episodic memories. 
Neurogenesis by special stem cells has been discovered in the dentate gyrus (part of the hippocampal formation) and is active throughout adult life; these new neurons may be preferentially recruited in the formation of memories.
In fact, it has been observed that the adult macaque monkey forms a few thousand new neurons daily \citep{gluck2013learning,gould1999neurogenesis}, possibly to encode new information \citep{becker2005computational}.

The establishment of new indices, together with their embedding vectors, are some of the most demanding learning tasks in the brain.
Functionally, our model assumes that a new episodic memory engram is quickly stored by establishing a new index and its connections to the representation layer, copying the episodic memory trace.
Although there exist several theories, little is known about how exactly new time indices are formed in the brain anatomically and how they quickly set up the bidirectional connection patterns with the representation layer, forming a hippocampal–cortical network \citep{frankland2005organization}. Here, the brain might employ already existing networks, which might also explain how a single index can influence the distributed representation layer with potentially numerous neurons.
A structural intra-hub connectivity might facilitate this process.

\subsection{Experiments with Self-supervised Learning}
\label{sec:fslearn}

Self-supervised learning (SSL) concerns learning without teacher-provided explicit training labels. It is biologically highly relevant. During most of life, an individual has to learn without explicit supervision (lifelong learning). 
In our approach, self-supervised learning works exactly like supervised training; the difference is that the agent's predicted 
winner-take-all unary and binary labels become the training labels. 
This form of self-supervised learning is a type of bootstrap learning; see the bootstrap Widrow-Hoff rule \citep{hinton1990bootstrap} and learning with pseudo-labels \citep{lee2013pseudo}. The generated data is then used to train our model using cross-entropy cost-function terms derived from perception, episodic, and semantic memory (cf. Appendix~\ref{sec:appImpl}).

In our SSL experiments (see Table~\ref{tab:self_supervised}), we establish new indices and their embeddings for perceptional episodes (episodic indices) and for new entities in those episodes (concept indices); this would be part of the fast nonparametric learning system. The self-supervised training for a novel episodic instance and a novel entity\footnote{The agent might decide that an entity is novel if the activation of all elements in $\mathrm{sig} (\mathbf{n}_S)$ are below a threshold for all known entities.} is modular and fast, both technically and most likely also biologically.
 Thus, we do not assume domain closure and consider that new perceptual experiences might require representations for new entities. The embedding vector of a new episodic instance $\mathbf{a}_{t'}$ is adapted to model the decoded triple statements. Decoded triple statements, which are more certain, will be decoded more often, will be more influential in memory formation, and will then be retrieved more likely in an episodic memory recall. Row 3 in the table shows that the semantic recall on entities is reasonable, but not as good as on entities trained with annotator-labeled data (see row 1). This is understandable since predicted labels are noisier than human-annotator-provided training labels. 

Now consider the slow parametric learning system.
Row~2 shows that the embeddings of already established entities are not negatively affected by SSL. Rows~4 and ~5 show perception performance on unseen entities. We see 
better performance after SSL has been applied. This shows that embeddings for classes and attributes are improved by SSL.

\subsection{Replay for Episodic Memory Consolidation}

As self-supervised learning adds information, the required memory capacity grows. In our approach, for each salient episode, a new episodic index with its embedding vector is established. 
Similarly, for each new entity, a new concept index with its embedding vector is added. 
The current thinking in neuroscience is that this capacity problem can be solved by a consolidation from MTL, with a limited capacity, to the neocortex, with an essentially unlimited capacity. This is called systems consolidation of memory (SCM).

We now consider the leading two theories about the consolidation of episodic memory. 
The \textit{standard theory} assumes that at some point, episodic memory becomes independent of the hippocampus and MTL over a period of weeks to years \citep{squire1995retrograde,frankland2005organization}.
In contrast, the \textit{multiple trace theory} assumes that both the hippocampus and MTL remain involved \citep{nadel1997memory,jonides2008mind,greenberg2010interdependence}; MTL remains the manager of complex spatial and relational memories \citep{whittington2020tolman}. 
In general, it is assumed that consolidation involves both the medial prefrontal cortex (mPFC) and the MTL. The temporal lobe might be where the transferred indices are established \citep{frankland2005organization, tonegawa2018role}.
After consolidation, episodic memories might be organized in temporal order or according to a similarity in representation (see Figure~\ref{fig:clustering_time}). 
It is assumed that consolation might be a process executed entirely or partially during sleep \citep{stickgold2005sleep}.

We propose consolidation by replay, which, in our model, can be executed as follows. An episodic instance $t$ in MTL is activated, which activates the representation layer with vector $\mathbf{a}_t$; this activation is then learned in the connection weights of a newly formed index in the neocortex, for example, by a form of Hebbian learning.
Thus, these index duplicates in the neocortex would inherit the connection weights. If the index in the neocortex becomes more distributed, this will lead to greater robustness of memories after consolidation. For a while, both representations exist in parallel, and this facilitates the learning and consolidation of new memories; gradually, the index representation in the neocortex might become dominant. 
Consolidation by replay has the advantage that there is no need for direct interactions of indices in both storage sites, only indirect interactions by a shared activation of the representation layer. Replay might also be one way of how the brain implements large-scale structural changes in the brain, in general, for example, as a consequence of brain damage or as a consequence of a changing world with new statistics.
Following the principle ``use it or lose it,'' replay might also be essential such that relevant information in consolidated memory is not forgotten.
From our model's perspective, the location of the index is irrelevant. It could be in MTL, in neocortex, or both at the same time. In either case, decoding relies on the same machinery (see Figure~\ref{fig-justsem}).
Cognitive maps might be more pronounced after consolidation in the neocortex \citep{binder2011neurobiology}.  From a cognitive perspective, it makes sense that the brain mainly consolidates unusual episodic instances involving semantic state changes or singular events. It is tempting to assume that recent episodic memory is represented in MTL, whereas remote episodic memory is stored in neocortex.

\subsection{Replay for Semantic Memory Consolidation}

Training by episodic memory replay might also be important for the gradual transition from episodic to semantic memory, in which episodic memory reduces its sensitivity and association to particular episodes so that the information can be generalized as semantic memory. 
 It might enable the semantic memory to adapt more quickly to state changes, for example, to a friend's status change 
from being single to being married.
Although this forgetting of the ``single'' state might require time, the label ``married'' might quickly show higher activity than ``single''; the mind might consult episodic memory to resolve the conflict, discovering that the label ``married'' was assigned to recent instances.
Semantic memory has a built-in forgetting mechanism since Equation~\ref{eq:multpr} shows that if an observation of a triple is not renewed, dividing by $N_{\mathrm{total}}$ implies that it will get less significant over time.
 Some theories speculate that episodic memory may be the gateway to semantic memory \citep{baddeley1974working,squire1987memory,baddeley1988cognitive,steyvers2004word,socher2009bayesian,mcclelland1995there,yee2014cognitive,kumar2015ask}.

\begin{table}[ht]
 \centering
 \resizebox{1\textwidth}{!}{
 \begin{tabular}{c c | c c c c c c c}
 \hline
 Training & Test & \multicolumn{7}{c}{Unary labels (accuracy)} \\
 Regime & Set & B-Class & P-Class & G-Class & Y/O & Color & Activity & Average\\ \hline
 SL & SL-D	& 100.0 & 100.0 & 100.0 & 100.0 & 100.0 & 100.0 & 100.0\\
 +SSL & SL-D	& 100.0 & 100.0 & 100.0 & 100.0 & 100.0 & 99.99 & 100.0\\ 
 +SSL & SSL-D & 85.51 & 88.36 & 93.16 & 48.32 & 59.42 & 79.78 & 75.76\\\hline
 SL & G	& 77.43 & 85.79 & 92.63 & \textbf{49.08} &62.35 &81.39 & 74.78\\
 +SSL & G	& \textbf{77.54} & \textbf{86.15} & \textbf{92.83} & 48.58 &\textbf{62.62} & \textbf{82.78} & \textbf{75.08}\\ \hline
 \end{tabular}
}
 \caption{
 Self-supervised learning on the VRD-E data. SL stands for supervised learning, {SSL} for self-supervised learning.
 We first trained the model ({SL}) in a supervised way with human-provided labels ({SL-D}) with only 50\% of the training images (2000 images). 
 Columns 3 to 9 show perfect semantic memory performance for unary labels ({first row}). 
 Then we continued to train the model with only the other 50\% of the training images
 (SSL-D) in the SSL-mode (+SSL). 
 The {second row} shows that self-supervised learning does not lead to a deterioration of prediction performance on the entities in SL-D. 
 The {third row} shows that the performance on SSL-D is quite good, but of course, not as good as on SL-D since predicted labels on this data are noisier than the human-provided training labels. {Rows 4 and 5} show performance in perception on new entities (generalization, G). The performance of +SSL is better than SL, which shows that self-supervised learning also improves the detection of classes and attributes, helping to fine-tune their embeddings.
}
 \label{tab:self_supervised}
\end{table}

\subsection{Forgetting and Modularity}
\label{sec:forgetting}

Our hypothesis is that for a new instance or a new entity, 
an index is established in MTL. Let's focus the discussion on a past episodic instance. 
If an episodic index is never consolidated into neocortex, the instance might eventually be forgotten since the resources in MTL are limited,  and the brain might need to reuse them.
On the other hand, if an index
is consolidated into neocortex, it might be there for the lifetime of the agent. 
In neuroscience, it is an open question if consolidated memories really ever are forgotten, or if only the access to those memories is lost. 
The proposed index representation is essential since a particular functional synapse between an index and a dimension in the representation layer only serves this one purpose of connecting the index with the corresponding dimension in the representation layer. It is not reused for any other purpose. Assuming that an index and its embedding vector are stable, there might still be an ``aging'' of a memory. One reason is that the meaning of a dimension in the representation layer might change over the lifetime of an agent. In other words, the grounding or embedding might later in life be different from the time the memory trace was established. A second issue is that the representations of other indices change over the lifetime of an agent. Thus, if Jack was part of a 10-year-old episode, the embedding vector for the index ``Jack'' might have changed since then. Reactivation of a past memory in combination with self-supervised learning might stabilize a past episodic memory, associated with the danger of changing that memory. It is well known that the reactivation of a past memory does not just strengthen that memory but also might change it. Note that this stabilization mechanism also only works in the context of index representations.

\section{Summary, Conclusions and Future Work}
\label{sec:concl}

We have shown how perception, memory, reasoning, and foundations for language and consciousness
can all be realized by different functional and operational modes of the oscillating interactions between an index layer and a representation layer in the BTN. Whereas the representation layer, as a global workspace, is prominent in the current discussion on consciousness, 
the introduced index layer is an original contribution. We have proposed that an index representation increases modularity and counteracts forgetting.
We have emphasized the role of episodic and semantic memory in perception. We have proposed an associative memory where recency is key to the recall of recent episodic memory and 
 similarity of episodic representations with the present scene representation 
 for remote episodic memory. 
  We argue that episodic memory models the observed data, and semantic memory provides background knowledge. Our approach explains the great similarity between episodic and semantic memory: 
 semantic memory is the expected episodic memory of a future instance.

 We have proposed that perception and memories first produce subsymbolic representations, which are subsequently decoded semantically to produce symbolic triple sentences.
Our paper contributed to the discussion (a recent example being \citep{lecun2022LLM}) of how language, thought, and subsymbolic processing interact.

As part of future work, we will address more systematically subsumption hierarchies and
how parts bind to form the whole, which is another form of compositionality addressed, for example, in capsule networks \citep{sabour2017dynamic}.
The agent itself is also an important concept: 
it is part of each event.
It might be quite relevant if the angry bear looks at a deer or at the agent itself. Other issues of the ``I'' are mood, state of mind, current mission, social contexts, and spatial location. 

In this article, we did not focus on spatial representations. It is well known that MTL is not only instrumental for forming novel episodic memories but also contains spatial representations, for example, in the form of grid cells and place cells.
It is generally assumed that MTL is central for reconstructing spatial memories, permitting complex spatial reasoning. We plan to investigate if instance indices, their embeddings, and their labels might also be the basis for spatial navigation and reasoning.

The brain does many different things simultaneously and addresses a particular problem with several strategies. 
{Our work emphasizes the role of perception and memory, and it might provide some insights into part of the amazing faculties of human intelligence.}

\appendix

\section{Cost Function Terms}
\label{sec:diffpom}

 $\mathbb{P}(s, p, o, t)$ has four categorical variables. 
 The number of states is $N_C^2 N_P N_T$. So there is one state for each Boolean variable.

\subsubsection*{Observation Model for Episodic Memory}

If $i_{s, p, ,o, t} = 1$, then the contribution to the cross-entropy cost function for binary labels is 
\[
- (
\log \mathbb{P}(p | s, o, t)
+ \log \mathbb{P}(o | s, t)
+ \log \mathbb{P}(s | t)
) 
\]
and for unary labels $- (\log \mathbb{P}(c | s, t)
+ \log \mathbb{P}(s | t)
)$.

\subsubsection*{Pre-observation Model for Semantic Memory}

If $i_{s, p, ,o, t} = 1$, then the contribution to the cross-entropy cost function for binary labels is
\[
- 
(
\log \mathbb{P}(p | s, o, \bar t)
+ \log \mathbb{P}(o | s, \bar t)) 
\]
and for unary labels $- \log \mathbb{P}(c | s, \bar t)$ .

\subsubsection*{Perception}

In perception (see Section~\ref{sec:ppp})
 \begin{equation*} \label{eq_aipp6}
 \mathbb{P}(s'=s, p'=p, o'=o | t'=t,
 \textit{BB}_{\textit{sub}}, \textit{BB}_{\textit{obj}}, \textit{BB}_{\textit{pred}}, \textit{scene}_{t'}) .
\end{equation*}
If $i_{s, p, ,o, t} = 1$, then the contribution to the cross-entropy cost function is
\[
- \log \mathbb{P}(s'=s, p'=p, o'=o | t'=t, 
 \textit{BB}_{\textit{sub}}, \textit{BB}_{\textit{obj}}, \textit{BB}_{\textit{pred}}, \textit{scene}_{t'}) . 
\]

\section{Boolean versus Categorical Models}
\label{sec:bm}

{As discussed, the interface between the multinomial and the two-state Bernoulli model is via sampling. If the multinomial model generates an $(s, p, o)$ or $(s, c)$ sample, then the 
random variable $Y_{s, p, o, t}$ is assumed true. Here, we show a close mathematical relationship between the probabilistic scores.} The key correspondence is that 
for entities that were observed in the episodic experience at $t$, 
\[
\mathbb{P}(s, p, o |t) \approx \mathbb{E}(Y_{s, p, o, t} ) / {N_{\textit{t}}} .
\]
From this, one can derive
\begin{equation}
\label{eq_ncond}
\mathbb{P}(p | s, o, t) \approx \mathbb{E}(Y_{s, p, o, t}) / {\sum_{p'} \mathbb{E}(Y_{s, p', o, t})}
\end{equation} 
\begin{equation}
\label{eq_ncond2}
\mathbb{P}(c|s, t) \approx
{\mathbb{E}(Y_{s, \textit{hA}, c, t})} / {\sum_{c'} 
\mathbb{E}(Y_{s, \textit{hA}, c', t})} 
\end{equation}

 For semantic memory, we have
 \[\mathbb{P}(s, p, o | \bar t) 
 \approx N_{s, p, o, \mathrm{known}} \mathbb{E}(Y_{s, p, o, \bar t}) / {N_{\textit{total}}}.
\]
Thus, the samples
from the pre-observation model
reflect the expected-state model weighted by the number of measurements on the triple statement. Sometimes it is possible to apply knowledge on mutually exclusiveness. For example, 
if we define the complementary of $c$ as $\bar c$, then
 \[
 \frac{\mathbb{P}(s, \textit{hA}, c, \bar t)}
 {\mathbb{P}(s, p, \textit{hA}, c, \bar t)
 + \mathbb{P}(s, \textit{hA}, \bar c, \bar t)}
 = 
\mathbb{E}(Y_{s, \textit{hA}, c, \bar t})
 .
\]
As an example, assume Sparky is observed to be barking in 10 episodic instances and not-barking in 1000 episodic instances. Then, 
$\mathbb{E}(Y_{\textit{Sparky}, \textit{hA}, \textit{Barking}, \bar t}) \approx 1\%$, so Sparky is rarely barking. But since he is sometimes barking, this would be reflected in a firing of ``Barking'', following 
$\mathbb{P}(\textit{Sparky}, \textit{hA}, \textit{Barking} | \bar t)$. The agent would have problems to understand if Sparky is often barking or not. 
In contrast, if the agent observes both the firing of 
``Barking'' and ``NotBarking'', the fraction of firing rates (see last equation) would tell the agent, that Sparky is mostly quiet.

Furthermore, 
\begin{equation} \label{eq_ncond3}
\mathbb{P}(p | s, o, \bar t) \approx \mathbb{E}(Y_{s, p, o, \bar t}) / \sum_{p'}
 \mathbb{E}(Y_{s, p', o, \bar t})
\end{equation}
\begin{equation} \label{eq_ncond4}
\mathbb{P}(c | s, \bar t) \approx
\mathbb{E}(Y_{s, \textit{hA}, c, \bar t})/ \sum_{c'} 
 \mathbb{E}(Y_{s, \textit{hA}, c', \bar t}) .
\end{equation}
In the last two equations, we have exploited that due to the LCWA, 
$N_{s, p, o, \mathrm{known}} = N_{s, p', o, \mathrm{known}}$ and
$N_{s, c, \mathrm{known}} = N_{s, c', \mathrm{known}} $.

\section{Visualisation}
\label{sec:visual}

\begin{figure}[htp]
\begin{center}
\includegraphics[width=1\linewidth]{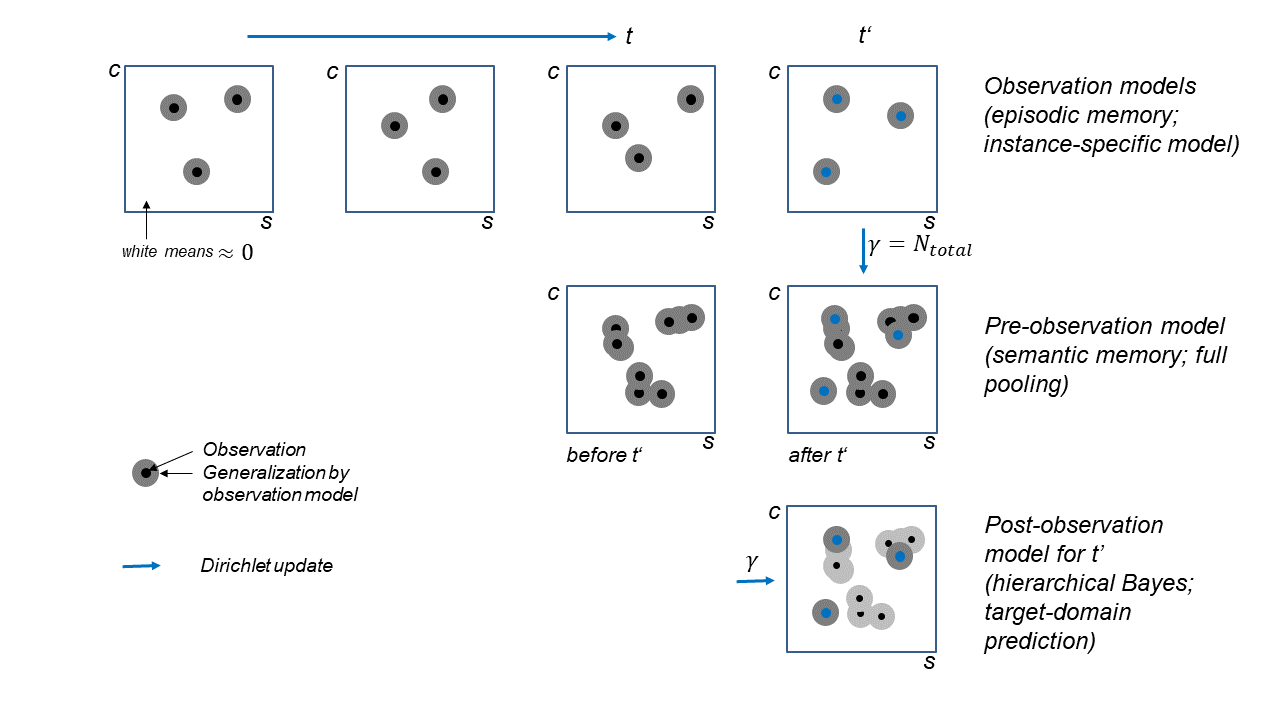}
\end{center}
\vspace{-0.5cm}
\caption{
Visualization for categorical variables. Here, we show the situation for unary statements. For binary statements, replace $c$ with $(p, o)$.}
\label{fig-Visual2}
\end{figure}

\begin{figure}[htp]
\begin{center}
\includegraphics[width=1\linewidth]{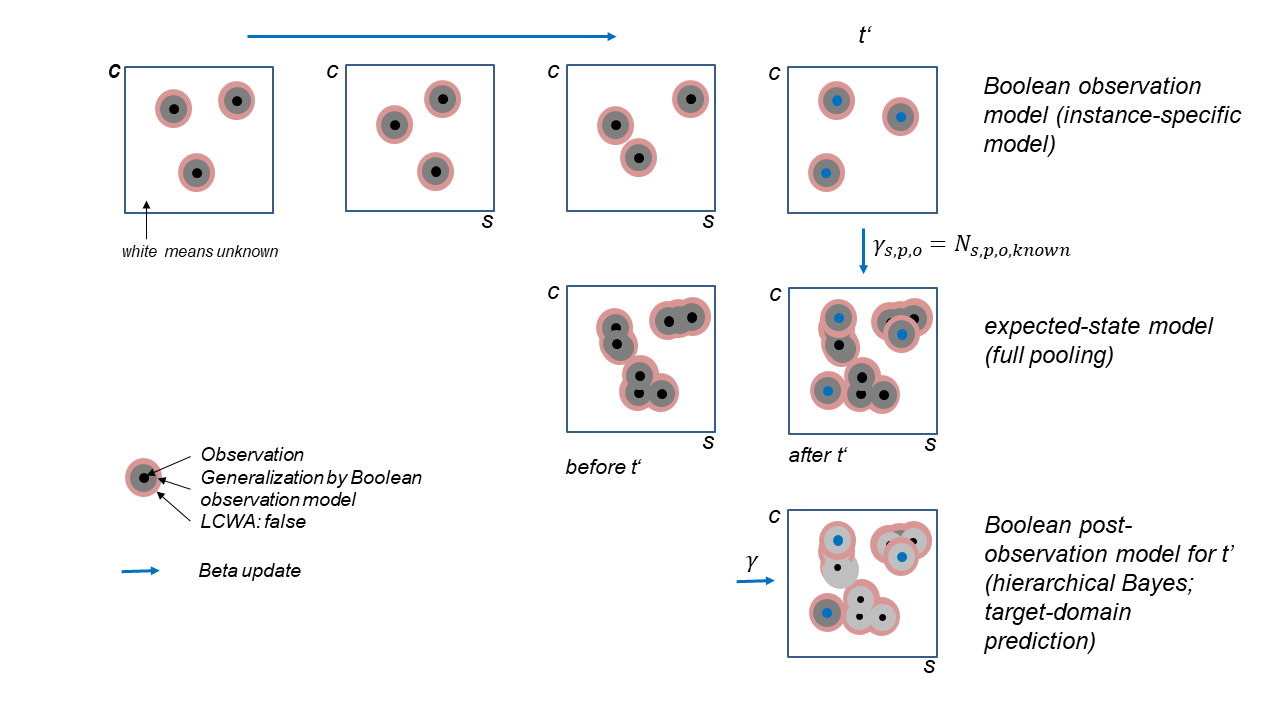}
\end{center}
\vspace{-0.5cm}
\caption{Same, but for Boolean variables. 
}
\label{fig-Visual}
\end{figure}

Figure~\ref{fig-Visual2} illustrates the situation for categorical variables. The update for the pre-observation and post-observation models are almost identical. The difference is that 
the observed data have a higher weight in the 
post-observation model. 
Figure~\ref{fig-Visual} shows the situation for Boolean variables.

% is with a known triple (true or false) at $t'$
% \begin{equation} \label{eq:dirfus}
% \mathbb{E}(Y_{s, p, o, \bar t}) \leftarrow
% \frac{1}{N_{s, p, o, \textit{known}} + 1} \left(N_{s, p, o, \textit{known}} \; \mathbb{E}(Y_{s, p, o, \bar t})
% + \mathbb{E}(Y_{s, p, o, t'})\right) .
% \end{equation}

% Figure~\ref{fig-Visual2} illustrates the situation for categorical variables. The update for the expected-state model is 
% \begin{equation} \label{eq:dirfus22}
% \mathbb{P}(s, p, o | \bar t) \leftarrow
% \frac{1}{N_{\textit{total}} + N_{t'} } \left(N_{\textit{total}} \; \mathbb{P}(s, p, o | \bar t)
% + N_{t'} \; \mathbb{P}(s, p, o | t'))\right) .
% \end{equation}
% In the actual algorithm we do not use these update equations explicitly; we use stochastic gradient decent updates. 
% The 

% The posteriors and the prior models use a $\gamma$ that reflects the belief in the prior models, whereas for the updates of the prior models one uses
% $N$

\section{Social Network}
\label{sec:appSN}

We consider all 4987 persons in the data set and link a person $s$ to persons $s'$, if the score $\mathbf{a}_s^{\top} \mathbf{a}_{s'}$ is in the top 5 of all scores related to $s$. Thus links exist between persons with similar embeddings, simulating homophily. We then determine the link direction. 
Considering two entities $s$ and $s'$, $\exp \beta \|\mathbf{a}_{s}\| / (\|\mathbf{a}_{s} + \mathbf{a}_{s'}\|)$ is proportional to the probability that we determine that \textit{(s, knows, s')}, otherwise, \textit{(s', knows, s)}.
At a social network episodic time step $t$, all links to one person $s$ are added. This defines
4987 episodes for the tKG. 
The pKG then aggregates the tKG. Overall, we have 24953 \textit{knows} statements. 
In summary, our social network data set has 4987 person entities (along with their unary labels), 24953 friendship statements, and 4987 episodes of social events.

\section{Implementation Details}
\label{sec:appImpl}

In this section, we focus on the implementation aspects and provide some details about the network architecture and training hyperparameters. Our program is written in Python and utilizes PyTorch.

\subsection{Network Architecture}

Our model mainly consists of two layers. The representation layer $\mathbf{q}$ has $r=4096$ neurons. The index layer $\mathbf{n}$ has $N_T + N_C + N_P$ neurons. We use VGG-19 backbone as the deep neural network $\mathbf{f}(\cdot)$ which takes as input a scene or a bounding box and outputs a 4096-dimensional feature vector. The VGG-19 network consists of a sequence of convolutional blocks followed by two fully connected hidden layers. Each convolutional block is a sequence of 2 convolution layers with 3x3 filters, a max-pooling layer, and another two convolution layers with the same parameters. We use the activations before the nonlinear transformation from the last hidden layer and copy them over to $\mathbf{q}$. Thus the layer $\mathbf{q}$ stores pre-activation values in the range of $\mathbb{R}$. Next, we apply a nonlinear function (LeakyReLU) to the values in $\mathbf{q}$ and feed them to the index layer via connection weights $\mathbf{A}^{\top}$. At different decoding steps, $\mathbf{n}$ covers different sets of indices, namely $\mathbf{n}_T$ has $N_T$ units for episodic instances, $\mathbf{n}_S$, $\mathbf{n}_C$, $\mathbf{n}_O$, all have $N_C$ concepts units (entities, classes, attributes), and $\mathbf{n}_P$ has $N_P$ units for predicates. The index layer in turn activates the representation layer via the same weights $\mathbf{A}$. To calculate the enhanced representation $\mathbf{q}_T$, $\mathbf{q}_S$, and $\mathbf{q}_O$, we add the pre-activation of layer $\mathbf{q}$ and corresponding activations from $\mathbf{n}$. To obtain $\mathbf{n}_S$ and $\mathbf{n}_O$, we apply softmax on the output with an inverse temperature $\beta=1$. For $\mathbf{n}_C$ we split the concepts into eight sets and apply softmax on each set. Alternatively, we could also use the sigmoid function as stated in our algorithm. However, softmax fits here as our labels are mutually exclusive, and in practice, softmax leads to faster convergence and slightly better performance. The dynamic context layer $\mathbf{h}$ contains 500 neurons with a self-connection via weight matrix $\mathbf{B}$. The input to the dynamic context layer is post-activation of $\mathbf{q}$, that is, values after a nonlinear transformation. There is a direct path for the hidden layer between different decoding steps (the dotted line between $\mathbf{h}$ blocks in figure ~\ref{fig-justsem}), which stores the state of the working memory and leads to a slight improvement (1\%) in relationship prediction. For $\bar{\mathbf{a}}$, we use a learnable embedding vector of length 4096.

\subsection{Training Scheme}

We train a global set of parameters for our BTN with an objective of minimizing the multi-task loss for perception, episodic, and semantic memory experience. We set the batch size to 128, the learning rate to 0.0001, and dropout $p$=0.5. The BTN is optimized using an Adam optimizer for 60 epochs. During training, we freeze the VGG-19 layers except for its last fully connected layer so that the knowledge in the pre-trained model will not be destroyed. For the last layer of VGG-19, we use a smaller learning rate of 0.00001 to allow it to adapt to the new task. Except for the VGG-19 backbone, we initialize our network using Kaiming uniform initialization proposed in \citep{he2015delving}. We minimize the summed cross-entropy loss on $\mathbf{n}_T$, $\mathbf{n}_S$, $\mathbf{n}_C$, $\mathbf{n}_O$, and $\mathbf{n}_P$ for the perception and memory experience. For $\mathbf{n}_S$ and $\mathbf{n}_O$, we randomly activate an index from the set of entity indices and class/attribute labels for learning generalized statements (cf. \ref{sec:gensta}). For $\mathbf{n}_C$, we apply cross-entropy loss on each subset. For the social network, we train the model for attribute prediction and relationship prediction on the social network data set. Concretely, we minimize the cross-entropy loss for $\mathbf{n}_S$, $\mathbf{n}_{C}$, $\mathbf{n}_O$, and $\mathbf{n}_P$. For SSL, we use a batch size of 128, a learning rate of 1e-5, and total training epochs of 20. Except for the embeddings of time and entities, all weights are frozen. Regarding the ablation studies, we train the RESCAL model using the implementation of PyKeen \citep{ali2021pykeen}. The rank of entity embeddings and predicate embeddings are both set to 1000, which gives a comparable number of learnable parameters as our model. All experiments are conducted on an Nvidia GTX 1089 Ti GPU with a 4 core CPU of 16G memory.

\section*{Code}
The source code related to this article is available at the following link: \url{https://github.com/hangligit/BTN}

\section*{Acknowledgements}
We thank the anonymous reviewers for their helpful comments on the manuscript.

\bibliography{TensorBrainSem}

\end{document}